\documentclass[journal]{IEEEtai}

\usepackage[colorlinks,urlcolor=blue,linkcolor=blue,citecolor=blue]{hyperref}

\usepackage{color,array}
\usepackage{cite}
\usepackage{amsmath,amssymb,amsfonts}
\usepackage{algorithmic}
\usepackage[linesnumbered,ruled,vlined]{algorithm2e}
\usepackage{graphicx}
\usepackage{textcomp}

\usepackage{booktabs}
\usepackage{tabularx}
    \newcolumntype{L}{>{\raggedright\arraybackslash}X}
\usepackage{multirow}

\DeclareMathOperator*{\argmin}{argmin}

\begin{document}

\title{Wireless Ad Hoc Federated Learning: \\A Fully Distributed Cooperative Machine Learning}

\author{\textbf{Preprint version}\\ Hideya Ochiai, Yuwei Sun, Qingzhe Jin, Nattanon Wongwiwatchai, and Hiroshi Esaki, \IEEEmembership{Member, IEEE}
\thanks{This work was supported by JSPS KAKENHI Grant Number JP 22H03572.}
\thanks{Hideya Ochiai, Yuwei Sun, Qingzhe Jin, Nattanon Wongwiwatchai, and Hiroshi Esaki are with the Graduate School of Information Science and Technology, University of Tokyo, Tokyo, 113-8656, Japan (e-mail: ochiai@elab.ic.i.u-tokyo.ac.jp).}

\thanks{This paragraph will include the Associate Editor who handled your paper.}}

\markboth{IEEE Transactions on Neural Networks and Learning Systems, Vol. 00, No. 0, Month 2020}
{H. Ochiai \MakeLowercase{\textit{et al.}}: Wireless Ad Hoc Federated Learning: A Fully Distributed Cooperative Machine Learning}

\maketitle

\begin{abstract}
Privacy-sensitive data is stored in autonomous vehicles, smart devices, or sensor nodes that can move around with making opportunistic contact with each other. Federation among such nodes was mainly discussed in the context of federated learning with a centralized mechanism in many works. However, because of multi-vendor issues, those nodes do not want to rely on a specific server operated by a third party for this purpose. In this paper, we propose a wireless ad hoc federated learning (WAFL) -- a fully distributed cooperative machine learning organized by the nodes physically nearby. WAFL can develop generalized models from Non-IID datasets stored in distributed nodes locally by exchanging and aggregating them with each other over opportunistic node-to-node contacts. In our benchmark-based evaluation with various opportunistic networks, WAFL has achieved higher accuracy of 94.8-96.3\% than the self-training case of 84.7\%. All our evaluation results show that WAFL can train and converge the model parameters from highly-partitioned Non-IID datasets over opportunistic networks without any centralized mechanisms.
\end{abstract}



\section{Introduction}
\label{sec:introduction}

\IEEEPARstart{C}{loud} computing has allowed massive data collection from users, forming Big Data for machine learning in 2010s \cite{hashem2015rise, l2017machine}. However, because of the privacy concerns, such as (1) utilization of them other than the original intent or (2) leakage of them by cyberattacks, privacy-related regulations are getting severe these days in order to protect especially those who are not information literate including children against unintended use of personal data\cite{hessel2020regulation, voigt2017eu, xiao2012security, pearson2013privacy}.

Privacy-sensitive data is stored in autonomous vehicles, smart devices, or sensor nodes that can move around and can make opportunistic contact with each other. Geographical locations, private photos, healthcare signals, and power consumption of homes are examples. Federation among such nodes was mainly discussed in the context of federated learning (FL) in many works with a centralized parameter server\cite{konevcny2016federated, li2020federated, tan2022towards}. However, because of multi-vendor issues, they do not want to rely on a specific server operated by a third party for this purpose but want to directly interact with each other in an ad hoc manner only when they have in contact just as cooperative intelligent transport systems do\cite{festag2014cooperative}.


We propose a wireless ad hoc federated learning (WAFL) -- a fully distributed cooperative machine learning organized by the nodes physically nearby. Here, each node has a wireless interface and can communicate with each other when they are within the radio range. The nodes are expected to move with people, vehicles, or robots, producing opportunistic contacts with each other.

\begin{figure}
\centering
\includegraphics[width=0.45\textwidth]{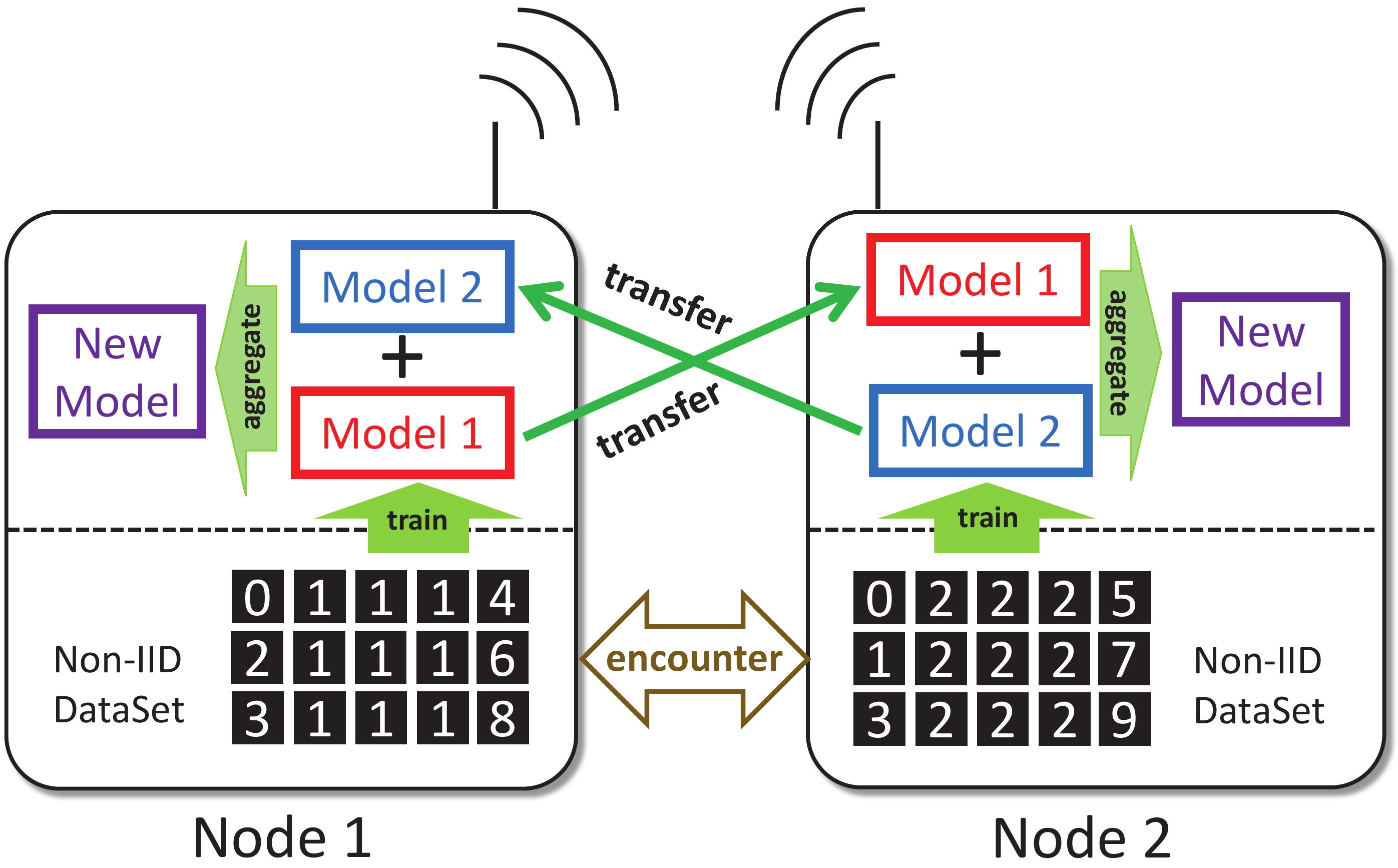}
\caption{The most basic wireless ad hoc federated learning. They exchange their models with each other when they encounter, and then aggregate them into new models. The aggregated models are expected to be more general, compared to the locally trained models with their own Non-IID dataset.}
\label{fig:WAFL_overview}
\end{figure}

In WAFL, we can start the discussion from the most basic peer-to-peer case as shown in Fig. \ref{fig:WAFL_overview}. In this scenario, each node trains a model individually with the local data it has. Here, node 1 can be a Toyota car, and node 2 can be a Mitsubishi car. When a node encounters another node, they exchange their local models with each other through the ad hoc communication channel. Then, the node aggregates the models into a new model, which is expected to be more general compared to the locally trained models. With an adjustment process of the new model with the local training data, they repeat this process during they are in contact. Please note that there is no third-party server operated for the federation among multi-vendor devices.

As WAFL does not collect data from users, the distributions of the data on individual nodes are not the same; e.g., a user has a larger portion of photos of hamburgers, but another user has a larger portion of dumplings based on their preferences or circumstances. This is a well-known problem of conventional federated learning as ``user data is not independent and identically distributed (Non-IID)''\cite{zhao2018federated}. The challenge is to develop a general model which does not over-fit into specific user data on the fully distributed, or partially-connected environment. 

In this paper, we introduce several node mobility patterns for evaluation from static cases to dynamic cases as benchmarks. In static cases, nodes do not move and organize static networks. We consider the topologies of line, tree, ringstar, and full-mesh in this study. In dynamic cases, nodes move around and sometimes make opportunistic contacts with each other. We consider several movement patterns based on random waypoint (RWP) mobility\cite{bettstetter2004stochastic}, and community structured environment (CSE) mobility\cite{ochiai2008mobility}.

We use Non-IID datasets for training and an IID dataset for testing for evaluating the generalization of the developed models. We create Non-IID datasets from the MNIST dataset based on the study of \cite{wang2020optimizing}. We use a fully-connected neural network for the classification of data samples intended for understanding the basic nature of the WAFL learning scheme including the behavior of model parameters.

We admit that a fully-connected neural network with MNIST dataset is a rather simple configuration if we consider today's evolution of machine learning research. Here, we emphasize that our focus or main contribution is the proposal of wireless ad hoc federated learning, which can be further applied into convolutional neural networks (CNN)\cite{lecun1995convolutional}, graph neural networks (GNN)\cite{kipf2016semi}, long short-term memory (LSTM)\cite{hochreiter1997long}, transfer learning schemes, Autoencoder\cite{hinton2006reducing}, generative adversarial network\cite{goodfellow2014generative}, word to vector\cite{mikolov2013efficient}, reinforcement learning\cite{mnih2016asynchronous}, and so on. We focus on the analysis of model aggregations with simple layer settings with a benchmark dataset on varieties of benchmark mobility scenarios. Further applications are open problems and beyond the scope of this paper. 

The motivations and contributions of this paper are summarized as follows:

\begin{itemize}
\item We assume federated learning in multi-vendor scenarios for autonomous vehicles, smart devices, or sensor nodes that can move around and have opportunistic contact with each other.
\item We propose WAFL for gaining generalized models from highly-partitioned Non-IID data just using opportunistic contacts without relying on centralized services i.e., without relying on third-party systems in multi-vendor scenarios. 
\item We also provide a theoretical analysis and available interfaces for model exchange.
\item This paper focuses on benchmark-based evaluation with MNIST and several node mobility patterns for understanding the characteristics of WAFL in learning performances and model convergence.  
\end{itemize}

This paper is organized as follows. Section II describes the related work. In section III, we propose wireless ad hoc federated learning. In section IV, we provide performance evaluation and model analysis. Section V provides the discussion on open research directions, and we finally conclude this paper in section VI.

\section{Related Work}

\begin{figure*}
\centering
\includegraphics[width=0.92\textwidth]{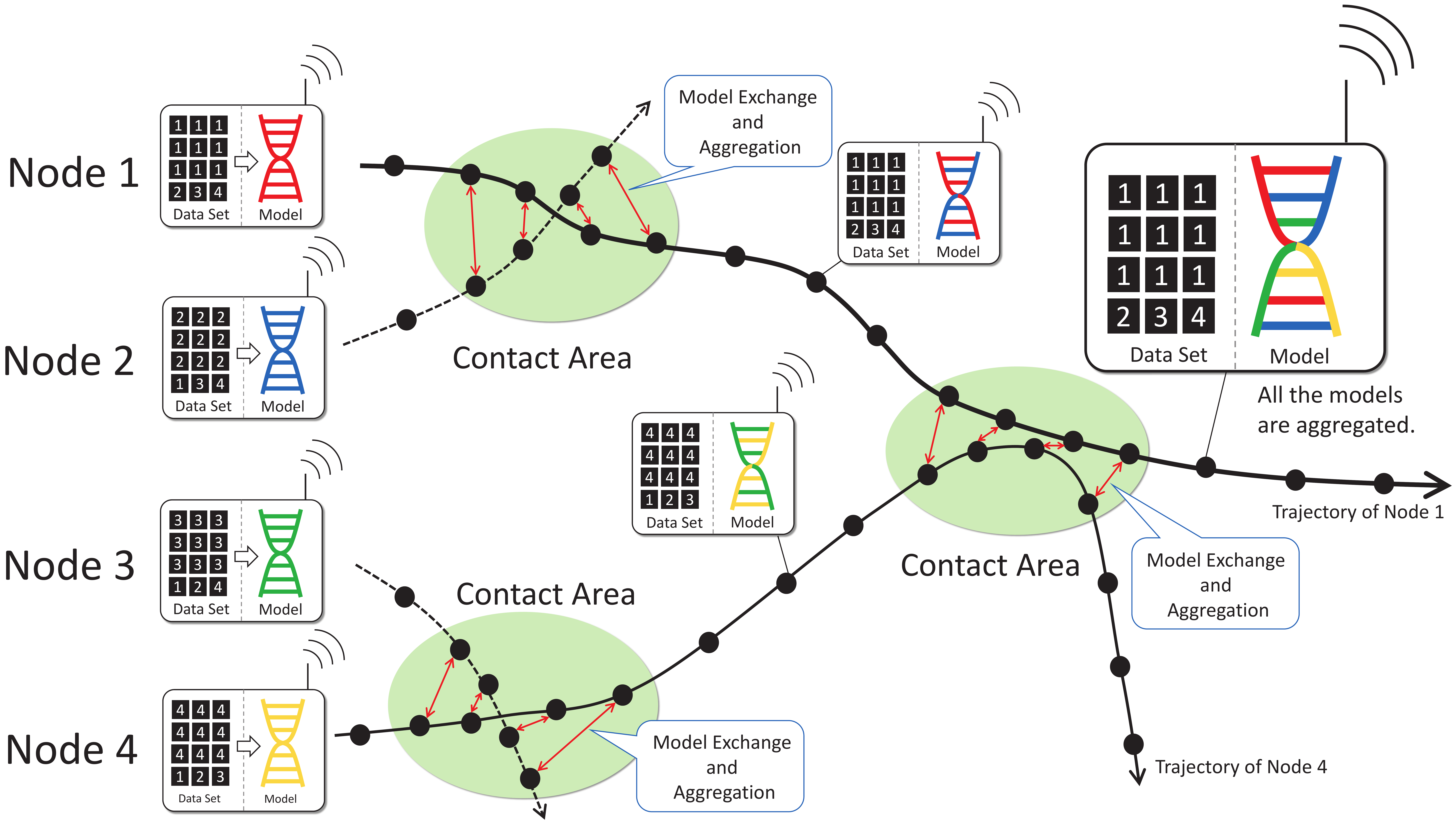}
\caption{Model exchange and aggregation with encountered smart devices in wireless ad hoc federated learning (WAFL). The nodes exchange and aggregate their models among the nodes encountered in an ad hoc manner. The initial models are trained too specific to their local Non-IID data, but in the long run, many contacts allow the mixture of locally trained models, making them more generalized.}
\label{fig:WAFL}
\end{figure*}

Wireless ad hoc federated learning (WAFL) is associated with wireless ad hoc network (WANET)\cite{frodigh2000wireless}, which is often discussed as a family of peer-to-peer (P2P)\cite{schollmeier2001definition} network in that it is server-less. However, P2P is usually discussed as an overlay network of the Internet infrastructure, which is constructed by the nodes joined on the Internet, finding each other, and constructing another network over the Internet. WANET is infrastructureless. Even without infrastructure, it can construct a communication network based on opportunistic physical contacts. Mobile ad hoc network (MANET)\cite{abolhasan2004review}, vehicular ad hoc network (VANET)\cite{hartenstein2008tutorial}, delay/disruption tolerant network (DTN)\cite{fall2003delay} are the family.

According to a comprehensive survey of federated learning\cite{nguyen2021federated}, our approach could be categorized into decentralized federated learning (DFL). BrainTorrent \cite{roy2019braintorrent} was proposed as a P2P decentralized federated learning where all the nodes on their P2P network can communicate with each other, and a node selects a target node for sharing and aggregating their models. In the case of WAFL, the nodes do not rely on the existing communication infrastructure and thus they cannot select such a target node. In the context of P2P, there were several attempts of applying Blockchain to add P2P characteristics to federated learning. However, in spite of the P2P network, blockchain-based federated learning basically has an idea of a ``global model'' for downloading, uploading, and aggregating \cite{li2020blockchain, bao2019flchain, pokhrel2020federated, ur2020towards}. Swarm learning \cite{warnat2021swarm} has been also proposed on a blockchain network for the application of machine learning to medical data. These systems are built on a peer-to-peer system, but the service is not basically decentralized. The fully decentralized federated learning such as WAFL should not have the idea of the global model.

Gossip-based federated learning (GFL) \cite{hu2019decentralized} and IPLS \cite{pappas2021ipls} could be said as another form of P2P federated learning. In GFL, a node specifies the target node from the available nodes and sends or pulls the part of the models. This is inspired by Gossip-based protocol\cite{haas2006gossip} -- another form of a distributed network similar to flooding-based or epidemic routing\cite{vahdat2000epidemic}. The key idea of these works is the segmentation or partition of model parameters with responsibility management. In our observation, such partitioning and responsibility management could work in the case of IID scenarios, which is the case that they evaluate their frameworks mainly, but cannot work in Non-IID scenarios like our case.

Federated learning is now also combined with unmanned aerial vehicles (UAV)\cite{brik2020federated}, where the nodes have to communicate over wireless channels. UAV's network is sometimes discussed with flying ad hoc network (FANET)\cite{bekmezci2013flying}. In this application, they currently assume the server-role UAV, which is the form of centralized federated learning \cite{liu2020federated, wang2020learning}. Other studies focus on the efficiency of power consumption \cite{pham2021uav, shiri2020communication} rather than the architecture of federated learning.

WAFL, our proposal, is fully autonomous, distributed, and tolerant for network partitioning or opportunistic contacts. WAFL does not have the idea of the centralized global model but still allows the development of a generalized model at each node by exchanging trained models from local Non-IID datasets through encounter-based ad hoc communication channels.

\section{Wireless Ad Hoc Federated Learning}

\begin{figure*}
\centering
\includegraphics[width=0.95\textwidth]{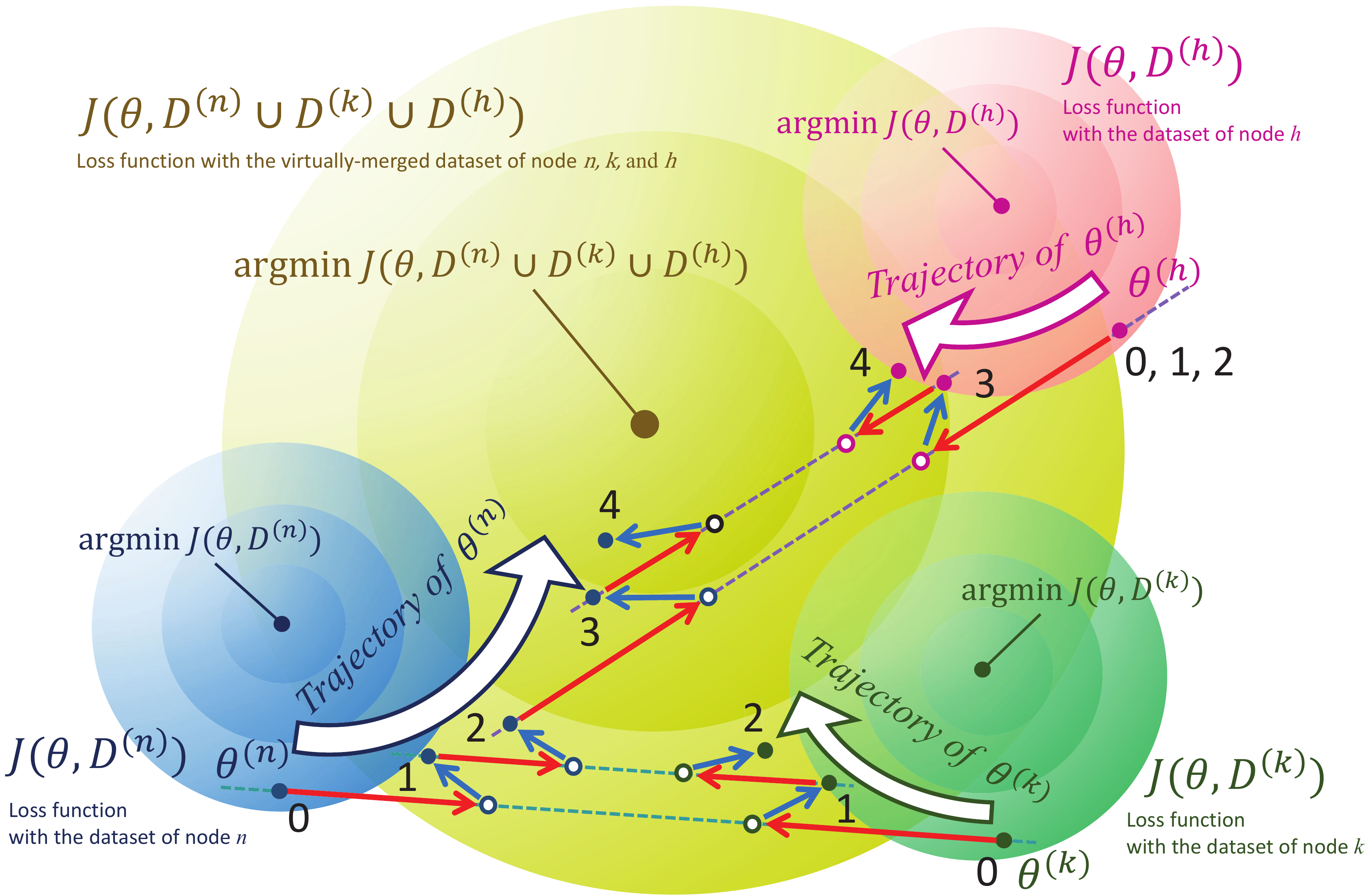}
\caption{Trajectories of model parameters $\theta$ on the loss fields in the parameter space. In this example, Node $n$ and $k$ encounter each other at epochs 0 and 1, and then leave at epoch 2. Instead, node $n$ encounters with node $h$ at epochs 2 and 3. The numbers shown in the figure indicate the epochs described in this statement. After the pre-training process, the initial (i.e., epoch=0) model parameters $\theta^{(n)}$, $\theta^{(k)}$, and $\theta^{(h)}$ are located near the argmins of $J(\theta,D^{(n)})$, $J(\theta,D^{(k)})$, and $J(\theta,D^{(h)})$ respectively. Here, because of the Non-IID characteristics, those argmins locate at different positions. After the process of model aggregation, i.e., Eq. (3), shown by red arrows, with local mini-batches, i.e., Eq. (4), shown by blue arrows, the model parameters shift toward the argmin of $J(\theta, D^{(n)} \cup D^{(k)} \cup D^{(h)})$ -- the parameters that minimize the loss of virtually merged data of node $n$, $k$, and $h$. }
\label{fig:WAFLLossField}
\end{figure*}

In wireless ad hoc federated learning (WAFL), each node has a wireless communication interface and can interact with the nodes when they are within the radio range. A node has its own data for training, but the data will not be exchanged with others because of privacy issues. Instead, they can exchange the parameters of the trained models with others, which can be aggregated to make the models more general.

Here, we assume that the data owned by a node is not independent and identically distributed (Non-IID). For example, Node 1 in Fig. \ref{fig:WAFL_overview} and \ref{fig:WAFL} has a larger portion of label 1 data, whereas Node 2 has a larger portion of label 2 data. If they train their model parameters simply based on their own local dataset, it will not become a generalized model, giving a poor performance on the accuracy to a general (i.e., an IID) testing dataset. Thus, in WAFL, we allow each node to exchange the trained model with the encountered nodes and aggregate the models. The developed models in this way will provide better performance because they originate in both larger portions of label 1 and 2 data samples. However, it is still oriented for 1 and 2 data.
In the WAFL scenario, as Fig. \ref{fig:WAFL} shows, the nodes move around and make opportunistic contacts among them. In the long run, with enough contacts based on node mobility, we expect that the developed models will provide further better performance to an IID test dataset.

\subsection{Definitions of Terms}

Before going into the WAFL proposal, let us describe the preliminaries to clarify our terms. Let $(X^{(n)},Y^{(n)}) \in D^{(n)}$ be a dataset owned by node $n \in \{1, 2, \ldots, N\}$, and let $\theta^{(n)}$  be the model parameters of node $n$. For $x \in X^{(n)}$, we present $f(x;\theta^{(n)})$ gives the prediction $\hat{y}$ : i.e., $\hat{y}=f(x;\theta^{(n)})$.

For each mini-batch, denoted by $D^{(n)}_B \subset D^{(n)}$, we calculate the loss by 
\begin{equation}
J(\theta, D^{(n)}_B)= \frac{1}{\vert D^{(n)}_B \vert}\sum_{(x,y) \in D^{(n)}_B} \ell(y, f(x; \theta)).
\label{eq:supervised}
\end{equation}

Here, $\vert D \vert$ denotes the number of elements in the set of D, and $\ell$ denotes the loss function. Conventionally, we calculate the next model parameters by
\begin{equation}
\theta^{(n)}_{i+1}=\theta^{(n)}_i-\eta\nabla J(\theta^{(n)}_i, D^{(n)}_B).
\label{eq:update_by_minibatch}
\end{equation}

Here, we denote $i$ by total mini-batch iterations. If $D^{(n)}$ is divided into $m$ mini-batches in an epoch, with given epoch $e$ and $j^{th}$ mini-batch, $i=m \times e+j$. Please note that $\eta$ is the learning rate. 

We use the above expressions in the following discussions. 

\subsection{Model Exchange and Aggregation with Encounter Nodes}
In WAFL, when a node encounters others, they exchange their own models with each other and aggregate them by themselves. In other words, a node sends its own model to the neighbors, and at the same time, it receives other models from the neighbors. We describe the details of the model exchange later in Section III.E. After the exchange, they aggregate into their own model. They repeat this process every time they are in contact.

Let $nbr(n) \subset \{1, 2, \ldots, N\} \setminus \{n\}$ be a set of neighbors of node $n \in \{1, 2, \ldots, N\}$. Please note that $nbr(n)$ does not include itself $n \notin nbr(n)$. In this study, we assume that the communication link is symmetric; $\forall k,n$:  $k \in nbr(n) \Leftrightarrow n \in nbr(k)$.

We define model aggregation in WAFL as follows.

\begin{equation}
\theta'^{(n)}_{me} = \theta^{(n)}_{me} + \lambda \frac{\sum_{k \in nbr(n)}{(\theta^{(k)} - \theta^{(n)}_{me})}}{\vert nbr(n) \vert +1}.
\label{eq:model_aggregation}
\end{equation}

Here, $\lambda$ is a coefficient. $\theta^{(n)}_{me}$, i.e., the model parameters of node $n$ at epoch $e$, is aggregated with the received model parameters from neighbors $k \in nbr(n)$ with the coefficient $\lambda$. Here, the epoch number is not explicitly provided for neighbor model parameters $\theta^{(k)}$, because it is not necessary to be the same epoch.

Then, the node adjust the model parameters with the local dataset, i.e., for each $\{0,1,\ldots,m-1\}$ mini-batch of $D^{(n)}$:

\begin{eqnarray}
\small
\theta'^{(n)}_{me+1}&=&\theta'^{(n)}_{me+0}-\eta\nabla J(\theta'^{(n)}_{me+0}, D^{(n)}_{B_0}), \nonumber \\ 
\theta'^{(n)}_{me+2}&=&\theta'^{(n)}_{me+1}-\eta\nabla J(\theta'^{(n)}_{me+1}, D^{(n)}_{B_1}), \nonumber \\
&\vdots& \nonumber \\
\theta'^{(n)}_{me+m}&=&\theta'^{(n)}_{me+m-1}-\eta\nabla J(\theta'^{(n)}_{me+m-1}, D^{(n)}_{B_{m-1}}).
\label{eq:minibatch_in_model_aggregation}
\end{eqnarray}

Each equation corresponds to a single mini-batch training, and these equations mean that each node repeats these mini-batches until it covers the whole local training dataset. 
This adjustment process should be carried out only if $\vert nbr(n) \vert > 0$, where the mixture of model parameters by Eq. (\ref{eq:model_aggregation}) is effective. If $\vert nbr(n) \vert=0$, this minibatch-based adjustment process should be skipped because it causes over-fitting to the local dataset. 

Finally, we get the model parameters of the next epoch $e+1$ by :
\begin{equation}
\theta^{(n)}_{m(e+1)} = \theta'^{(n)}_{me+m}.
\label{eq:minibatch_in_model_aggregation2}
\end{equation}

\subsection{Pre-Training for WAFL Environment}

In our study, we found that before actually starting the model exchange, each node should train its model parameters by itself with its own local data. This initial self-training process allows the rapid increase of accuracy in WAFL's model aggregation phase.

This self-training can perform the following calculations independently at each node $n$,

\begin{equation}
\theta^{(n)}_0 \approx \argmin_{\theta} J(\theta,D^{(n)}).
\label{eq:self_training}
\end{equation}

These model parameters are optimized for local training data. Thus, for the virtually merged dataset, we will observe:

\begin{equation}
J(\theta^{(n)}_0,\bigcup_{u=1}^{N}{D^{(u)}})
\gg \min_{\theta}{J(\theta,\bigcup_{u=1}^{N}{D^{(u)}})}.
\end{equation}

This means that only the self-training causes an overfit to the local Non-IID dataset. The goal of WAFL is to find better $\theta^{(n)}$ that minimizes the loss on the virtually merged dataset.

Please note that this self-training should not be carried out after starting the model exchange phase. It loses learned parameters, i.e., over-fits to the local data, especially when it runs many self-training epochs.

\begin{table*}
\centering
\caption{The configuration of Non-IID MNIST data samples. 90\% of node $n$'s data is composed of label $n$'s samples, and the other 10\% is uniformly composed of the other label's samples without any overlaps among the nodes. \label{tab:noniid_distribution}}
\begin{tabular}{c|cccccccccc|c}
\hline
Node&L0&L1&L2&L3&L4&L5&L6&L7&L8&L9&Summary \\ \hline
0&5341&76&64&76&61&57&57&61&74&50&5917 \\
1&79&6078&67&52&57&59&58&79&59&68&6656 \\
2&58&67&5374&80&64&68&87&73&57&61&5989 \\
3&68&74&73&5537&51&56&72&65&67&77&6140 \\
4&73&67&80&67&5301&59&53&69&70&64&5903 \\
5&60&66&57&74&68&4896&61&59&66&69&5476 \\
6&52&78&53&56&58&66&5312&56&65&65&5861 \\
7&67&90&66&74&55&61&65&5683&63&77&6301 \\
8&59&80&59&54&63&48&88&67&5268&57&5843 \\
9&66&66&65&61&64&51&65&53&62&5361&5914 \\ \hline
Summary&5923&6742&5958&6131&5842&5421&5918&6265&5851&5949&60000 \\ \hline
\end{tabular}
\end{table*}

\subsection{Insight into Model Parameters}

If $\vert nbr(n) \vert=1$, Eq. (\ref{eq:model_aggregation}) can be simplified into the following formula with the sole neighbor $k$:
\begin{equation}
\theta'^{(n)}_{me} = \big(1-\frac{\lambda}{2}\big)\theta^{(n)}_{me} + \frac{\lambda }{2}\theta^{(k)}.
\label{eq:simple_agg}
\end{equation}

This means that the model parameters $\theta^{(n)}_{me}$ is shifted toward $\theta^{(k)}$. If $\lambda=2$, it will be replaced with $\theta^{(k)}$. If $\lambda=1$, it is shifted at the center position between $\theta^{(n)}_{me}$ and $\theta^{(k)}$. Practically, $\lambda$ will be set to 1 or less: i.e., $0 < \lambda \leq 1$.

Fig. \ref{fig:WAFLLossField} explains why this model aggregation is effective for finding optimal model parameters over a virtually merged dataset.

As described, the aggregation of the model parameters with Eq. (\ref{eq:model_aggregation}) or Eq. (\ref{eq:simple_agg}) makes $\theta'^{(n)}$ shifted toward decreasing $J(\theta^{(n)},D^{(k)})$ (denoted by red arrow in the figure). With the mini-batches, i.e., Eq. (\ref{eq:minibatch_in_model_aggregation}), it shifts toward decreasing  $J(\theta^{(n)},D^{(n)})$ (denoted by blue arrow in the figure). After several epochs, it will find better parameters for reducing $J(\theta^{(n)},D^{(n)} \cup D^{(k)})$, without actually exchanging the datasets: i.e., $D^{(n)}$ and $D^{(k)}$.

In WAFL, after finding the better parameters in this way, these two nodes will be separated, and in the next phase, encounter other nodes.  Let's assume that $n$ encounters $h$, here. Then, the parameters will be shifted to another direction -- toward decreasing $J(\theta^{(n)},D^{(h)})$, but this does not mean that it forgets the learned parameters for $J(\theta^{(n)},D^{(n)} \cup D^{(k)})$, especially when
\begin{equation}
\nabla J(\theta^{(n)},D^{(h)}) \perp  \nabla J(\theta^{(n)},D^{(n)} \cup D^{(k)}).
\end{equation}
This is because the parameter space is multi-dimensional.

After the long run with many encounters with various nodes, the model parameters will be tuned for the virtually merged data,
\begin{equation}
\hat{\theta^{(n)}} \approx \argmin_{\theta^{(n)}} J(\theta^{(n)},\bigcup_{u=1}^{N}{D^{(u)}}).
\end{equation}

\subsection{Interfaces for Ad Hoc Model Exchange}

There are several choices of wireless interfaces for device-to-device communications, e.g., (1) Wi-Fi, (2) Bluetooth, and (3) sensor network protocols. In our daily life, most of Wi-Fi interfaces are working in station mode, where the node connects to an access point to join the network. However, in WAFL, we use Wi-Fi in ad-hoc mode, which is equipped with almost all Wi-Fi devices available in the market. 

If ad hoc mode is selected, the devices can automatically exchange Wi-Fi frames with the neighbors located within the radio range. Of course, the other devices also need to work in ad hoc mode and they should be associated with the same service set identifier (SSID). 

Practically, in wireless ad hoc networks, they implement a TCP/IP protocol stack on the Wi-Fi interface. To communicate with TCP/IP-based protocols with each other, the operating system automatically assigns a random link-local address in the range of 169.254.0.0/16 (IPv4) and in the range of fe80::/64 (IPv6).

Then, the devices can communicate with neighbors with TCP/IP protocols. Here, specifically, we can use UDP multicast for sending the model to the neighbors. In the case of multicast, all the nodes listen to the specific multicast address and receive the message if sent by neighbors\cite{ochiai2011case}. Please note that each node does not route or relay the message to others.

The model for exchange should be compressed but would usually exceed the limit of UDP payload size. Thus, it should be segmented into multiple UDP packets in practice as a function of the application-layer protocol before the transmission. The size of the model we use in our evaluation is around 400kbyte, which can be easily transmitted over Wi-Fi.

We can also implement such model exchange functions over Bluetooth as also shown in \cite{ochiai2011case}. Please remind that COVID-19 contact tracing\cite{ahmed2020survey} uses Bluetooth of smartphones for beacon exchange among the encountered devices.

\subsection{Static and Dynamic Network Applications}

Because of the above nature of ad hoc networks, the nodes can autonomously self-organize static or dynamic networks without any mechanisms including centralized controls. 

We define a static network as a communication network that does not change its topology, i.e., the nodes and the links of the network graph do not change over time. Such a network can be organized especially in the case of sensor networks where wireless sensors are physically deployed. For example, electric meters for personal use installed for learning the features of their home appliances one by one will organize a static network. 

We define a dynamic network as a communication network that changes its topology over time driven by the physical movement of the nodes. For example, in a shopping mall or in an amusement park, smartphone devices move along with people and encounter other devices over dynamically changing wireless links. In such a scenario, for example, they can collaboratively make a smart shopping mall by exchanging and aggregating the models developed from user-annotated maps of the mall. 

Depending on the application cases, the characteristics of a dynamic network can be changed. For example, if we consider social communities such as a university, a student belongs to a laboratory, attends a classroom, and has lunch at a canteen. He or she meets others at these rendezvous points. Another dynamic network will be developed in the case of vehicles. They drive in open spaces where they can potentially contact a much larger number of vehicles.


\section{Evaluation}

\begin{figure*}
\centering

 \begin{minipage}[b]{0.22\textwidth}
  \raggedright
  \includegraphics[keepaspectratio, scale=0.25]{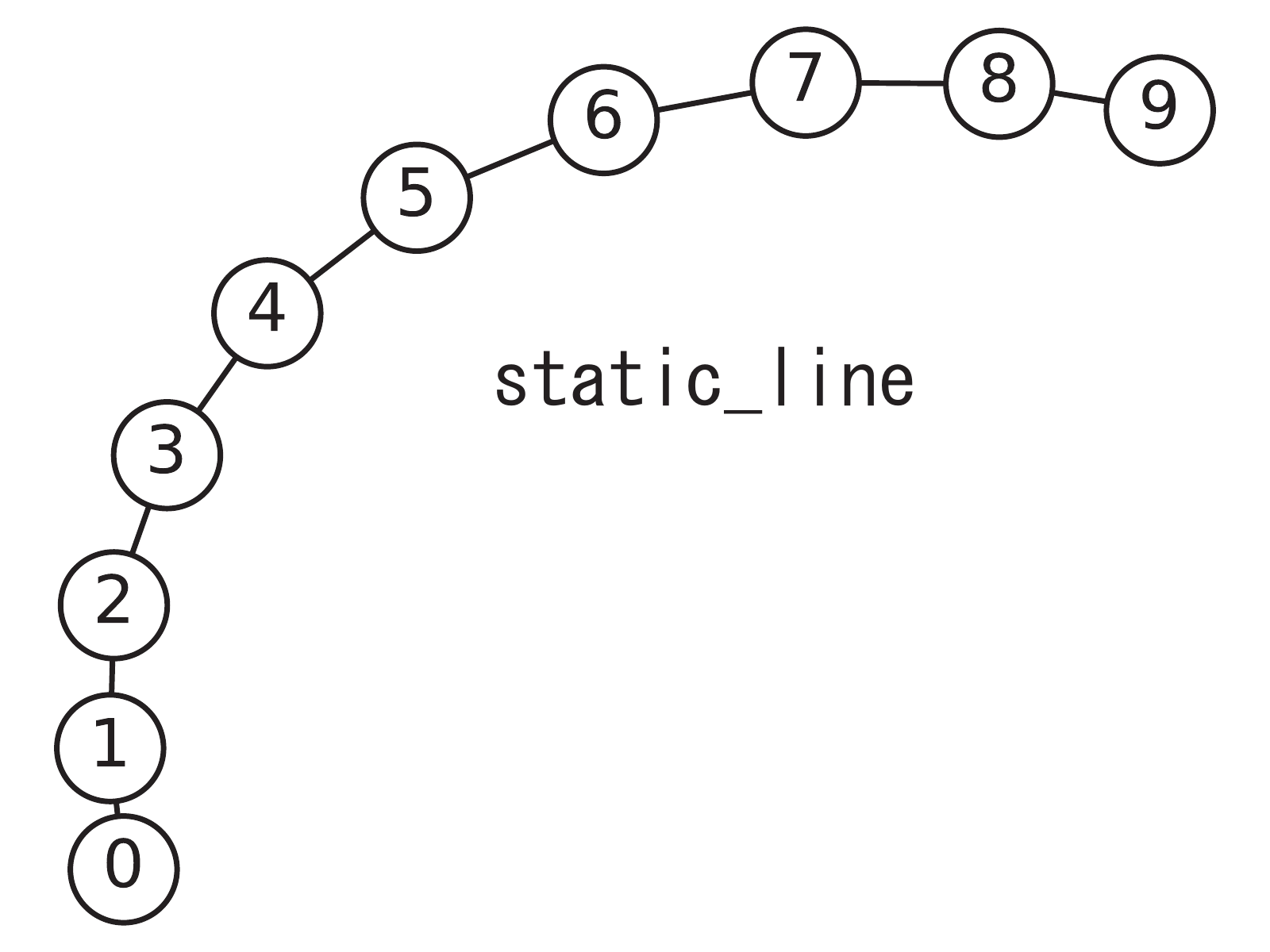}\\
  \end{minipage}
  \begin{minipage}[b]{0.22\textwidth}
  \raggedright
  \includegraphics[keepaspectratio, scale=0.25]{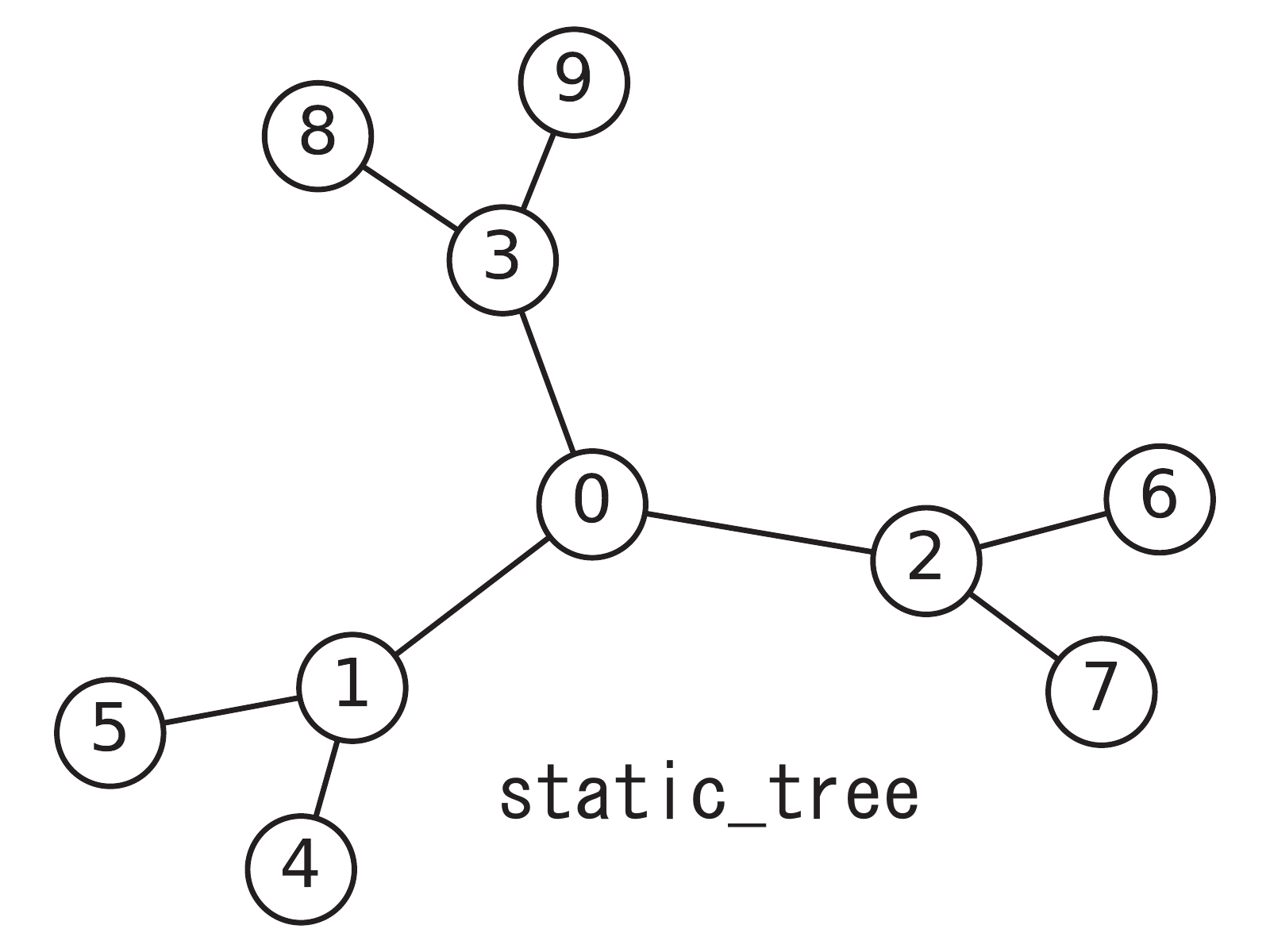}\\
  \end{minipage}
  \begin{minipage}[b]{0.22\textwidth}
  \raggedright
  \includegraphics[keepaspectratio, scale=0.25]{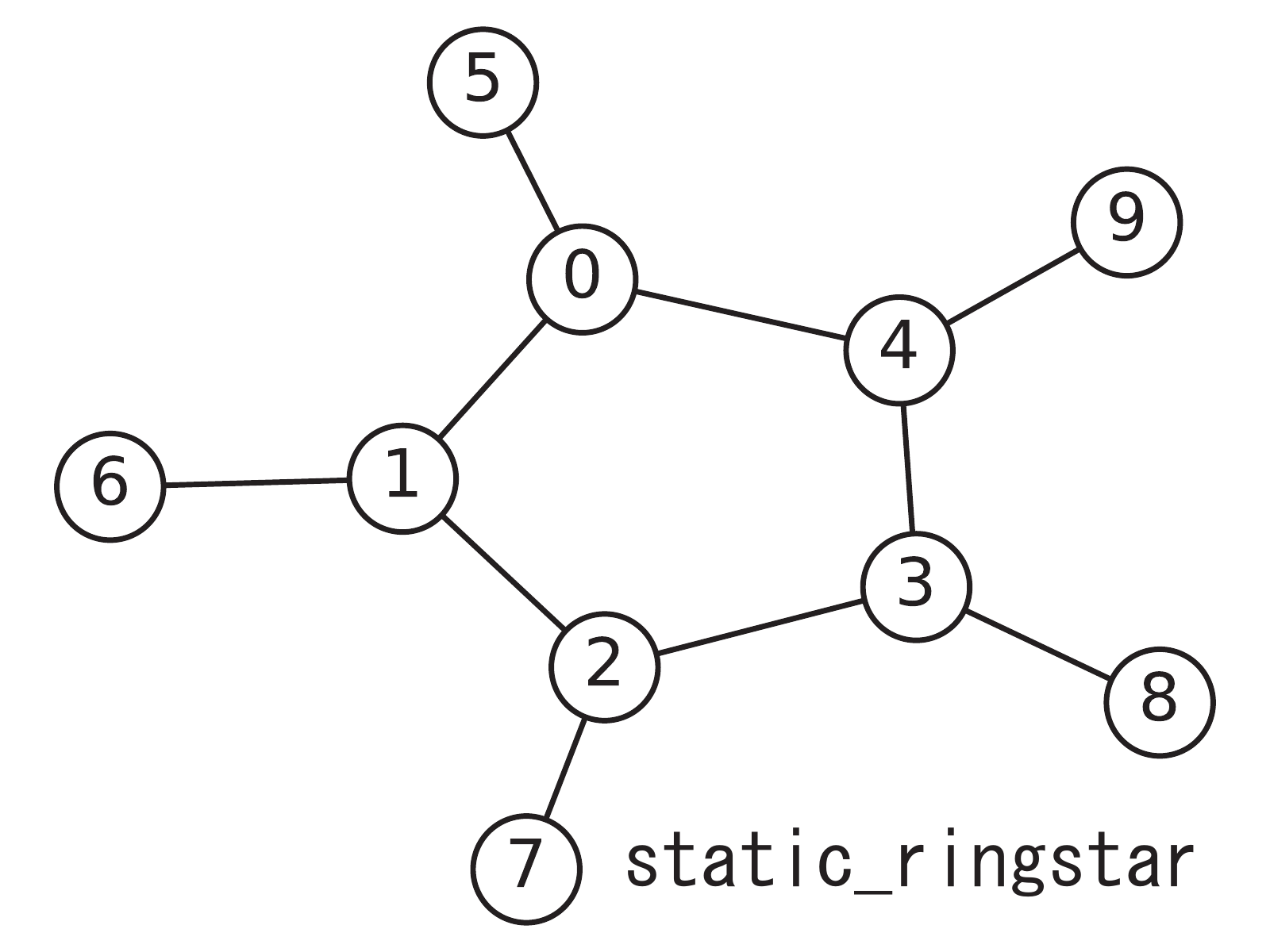}\\
  \end{minipage}
  \begin{minipage}[b]{0.22\textwidth}
  \raggedright
  \includegraphics[keepaspectratio, scale=0.25]{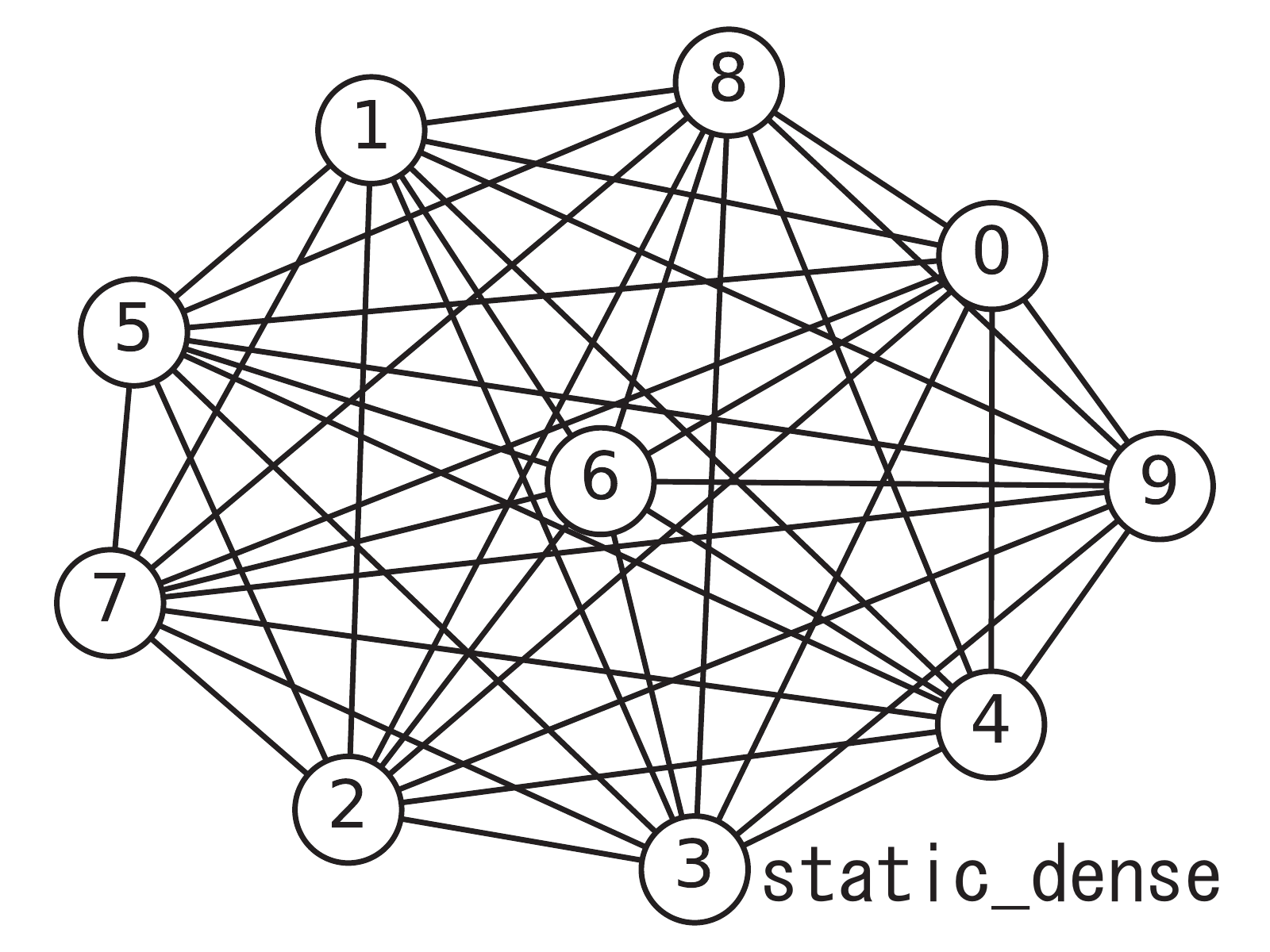}\\
  \end{minipage}
  
 \begin{minipage}[b]{0.22\textwidth}
  \raggedright
  \includegraphics[keepaspectratio, scale=0.25]{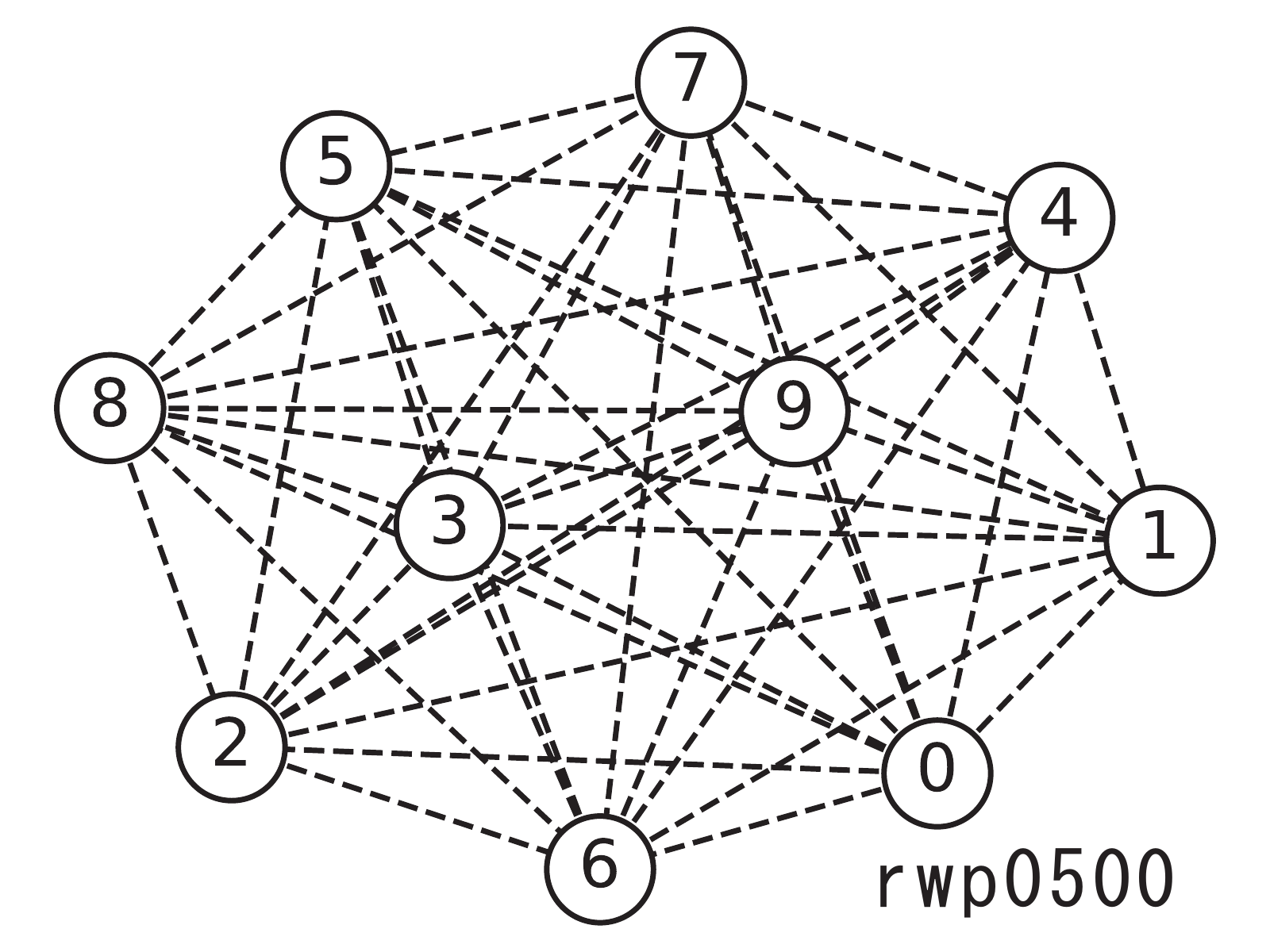}\\
  \end{minipage}
  \begin{minipage}[b]{0.22\textwidth}
  \raggedright
  \includegraphics[keepaspectratio, scale=0.25]{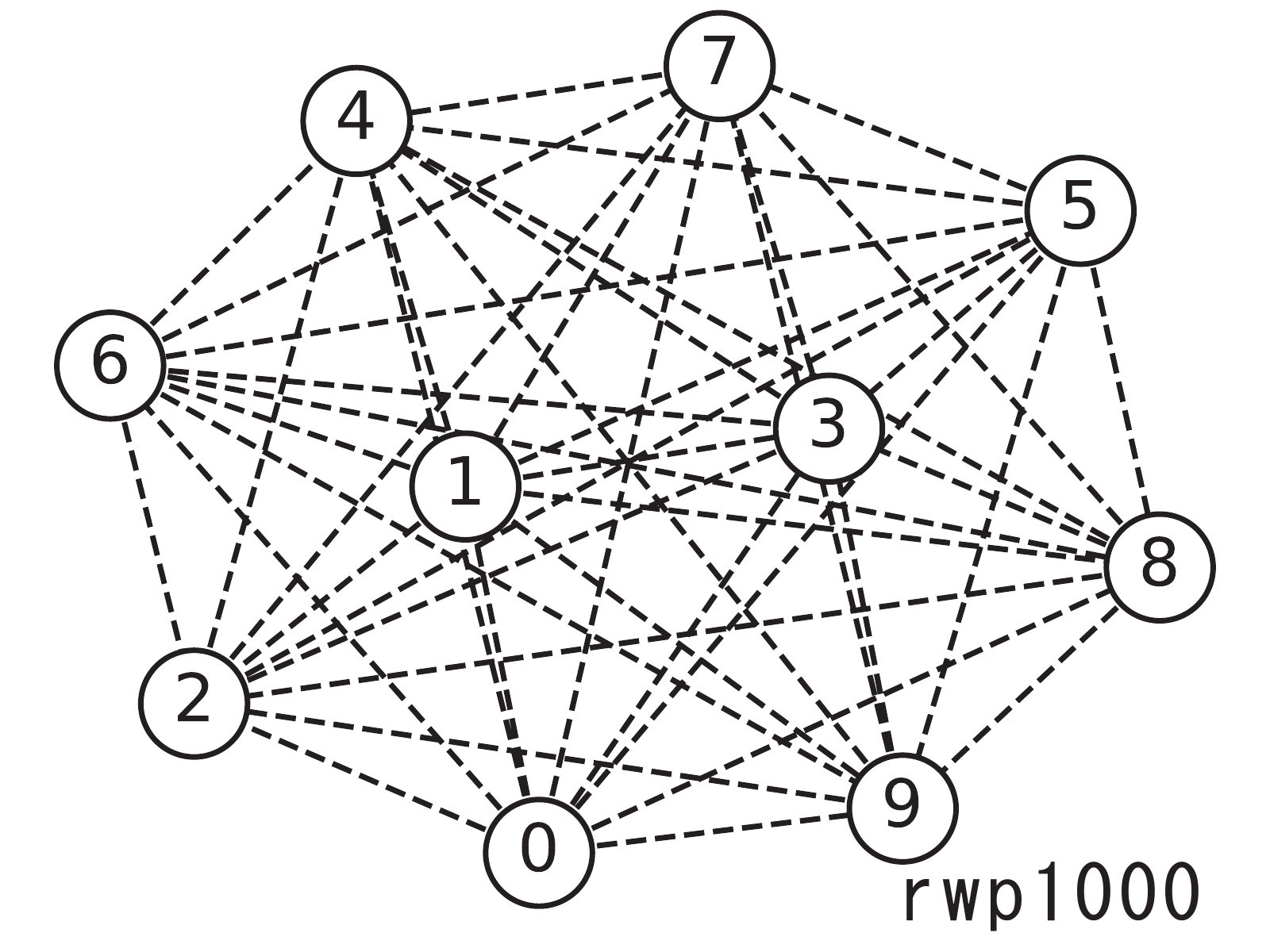}\\
  \end{minipage}
  \begin{minipage}[b]{0.22\textwidth}
  \raggedright
  \includegraphics[keepaspectratio, scale=0.25]{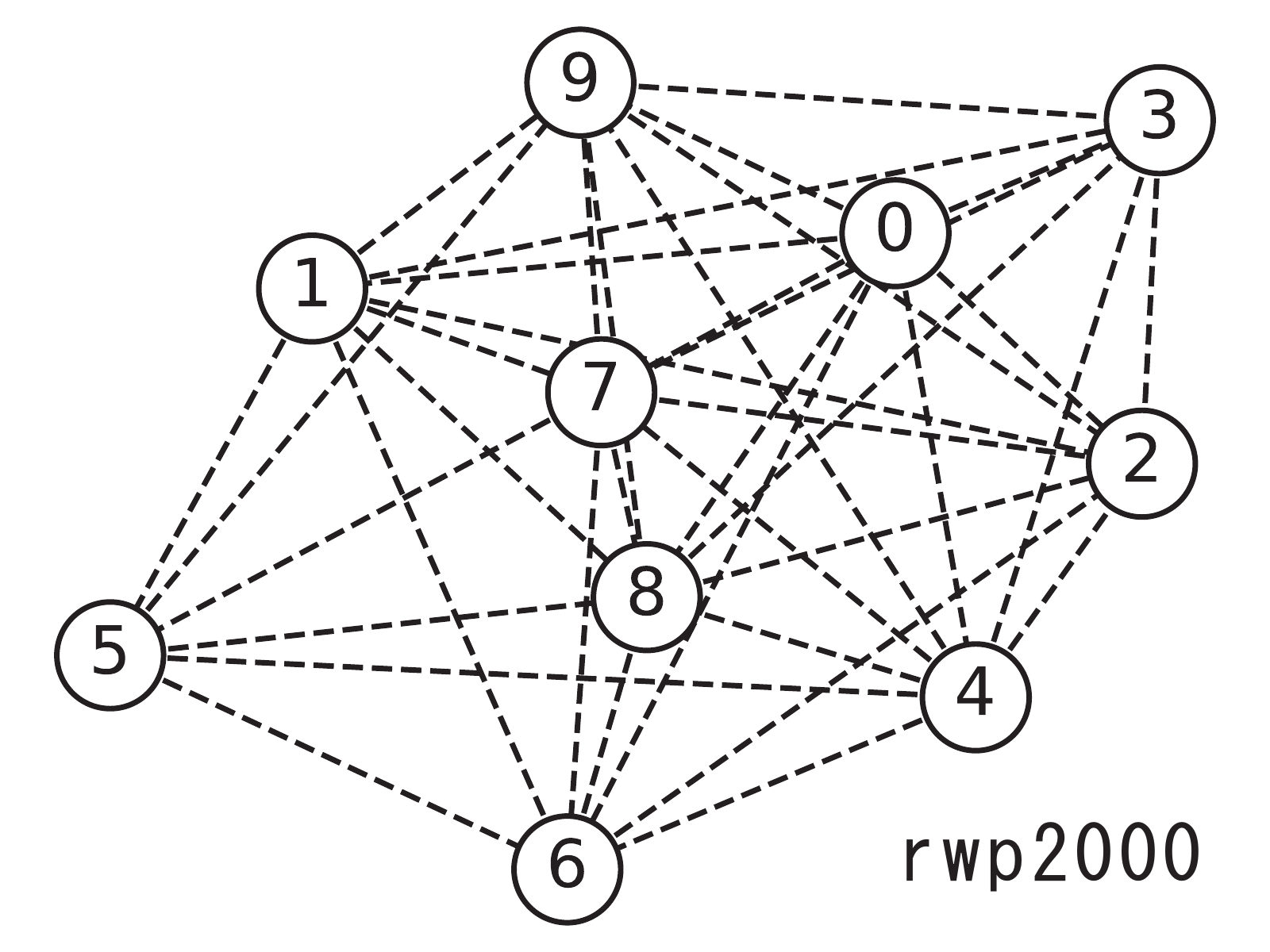}\\
  \end{minipage}
  \begin{minipage}[b]{0.22\textwidth}
  \raggedright
  \includegraphics[keepaspectratio, scale=0.25]{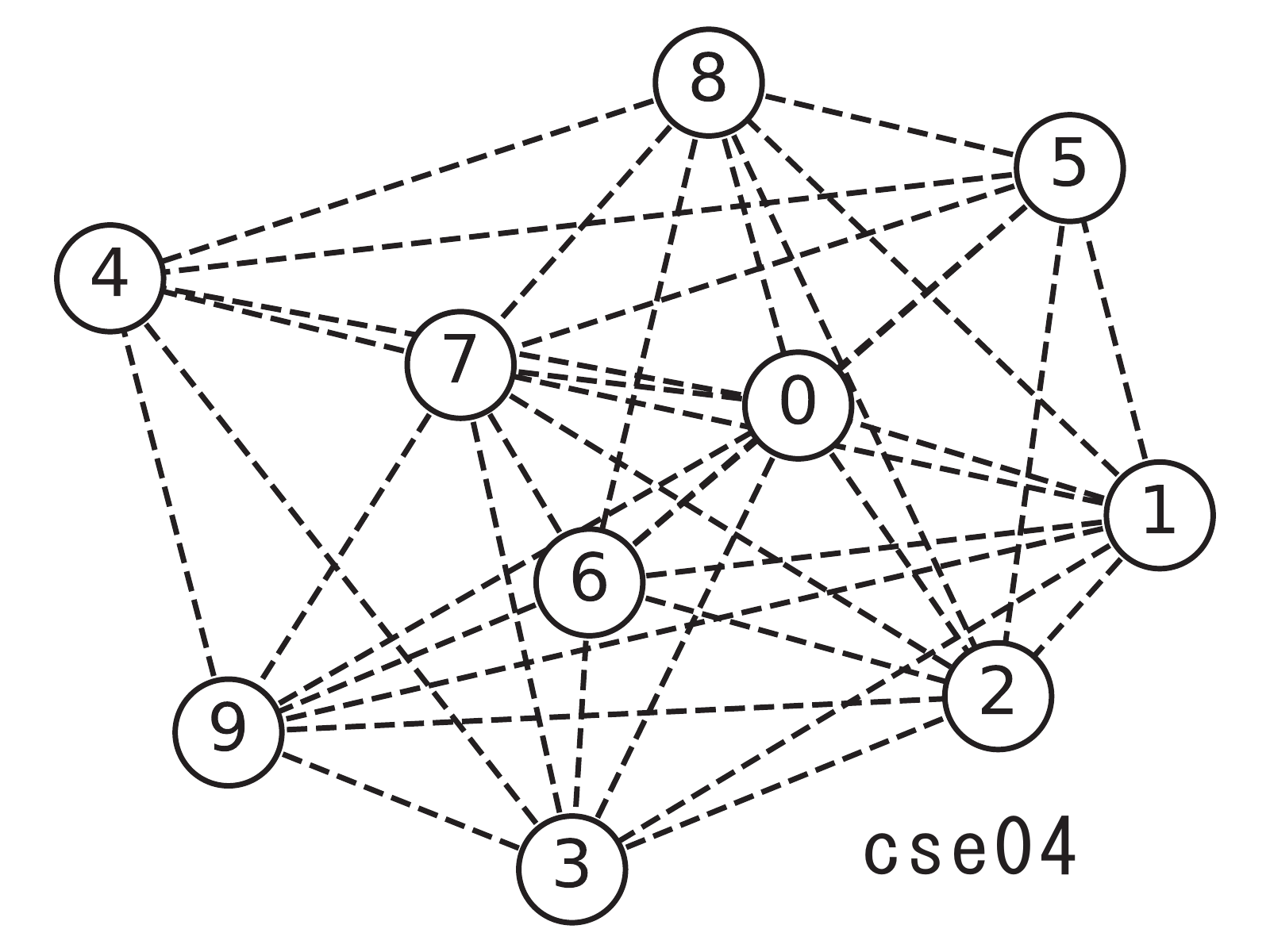}\\
  \end{minipage}

  \begin{minipage}[b]{0.22\textwidth}
  \raggedright
  \includegraphics[keepaspectratio, scale=0.25]{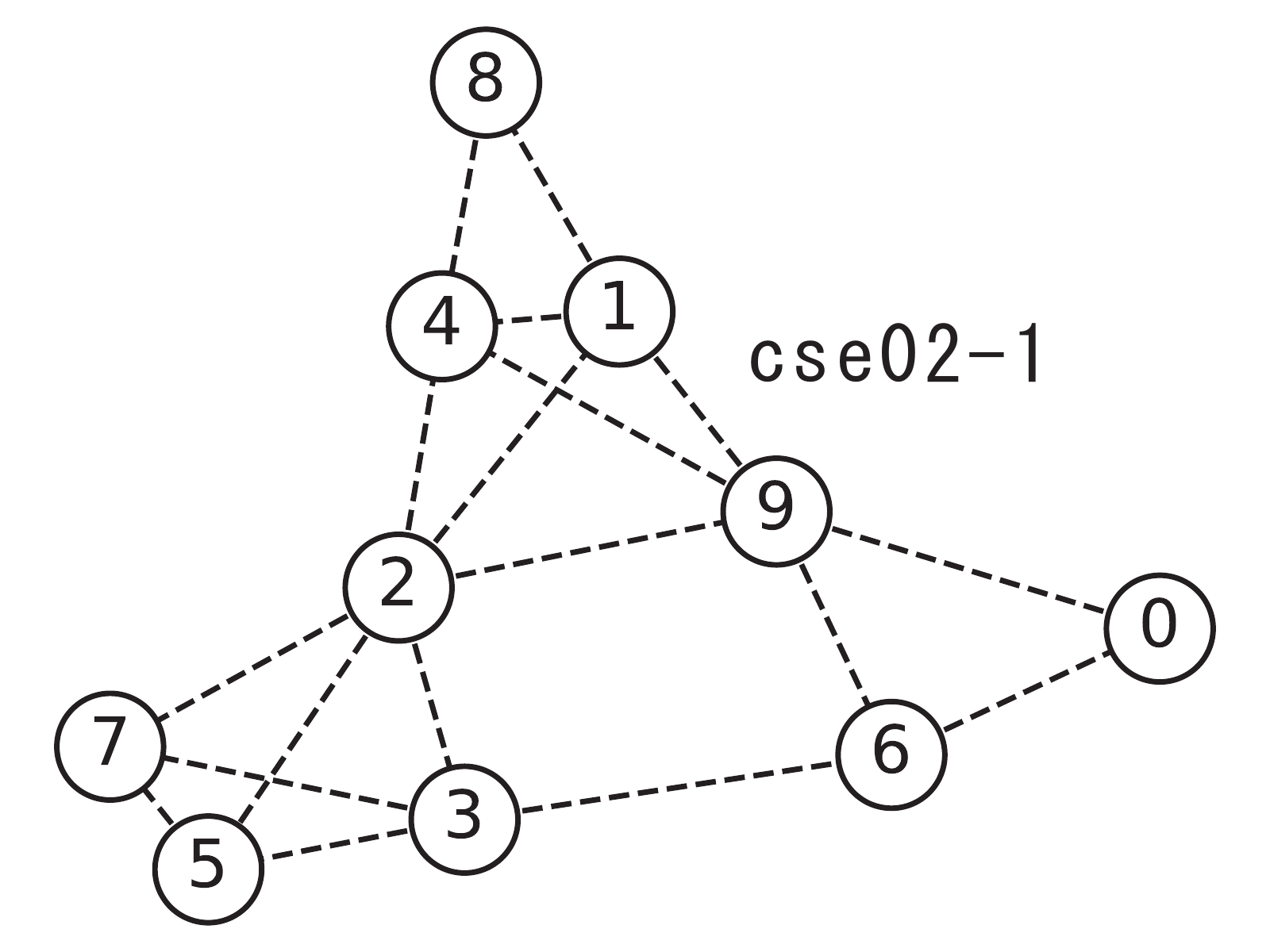}\\
  \end{minipage}
  \begin{minipage}[b]{0.22\textwidth}
  \raggedright
  \includegraphics[keepaspectratio, scale=0.25]{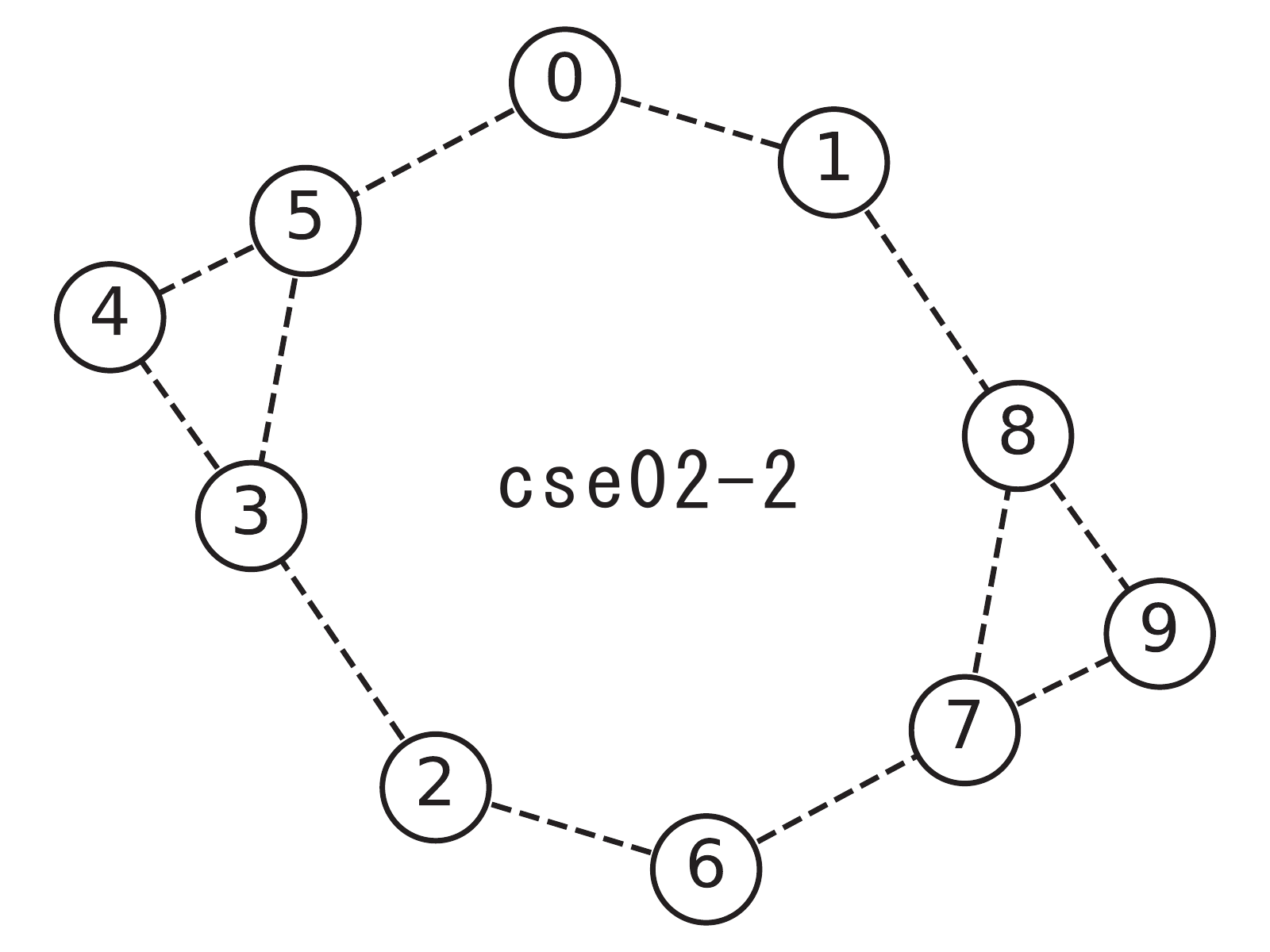}\\
  \end{minipage}
  \begin{minipage}[b]{0.22\textwidth}
  \raggedright
  \includegraphics[keepaspectratio, scale=0.25]{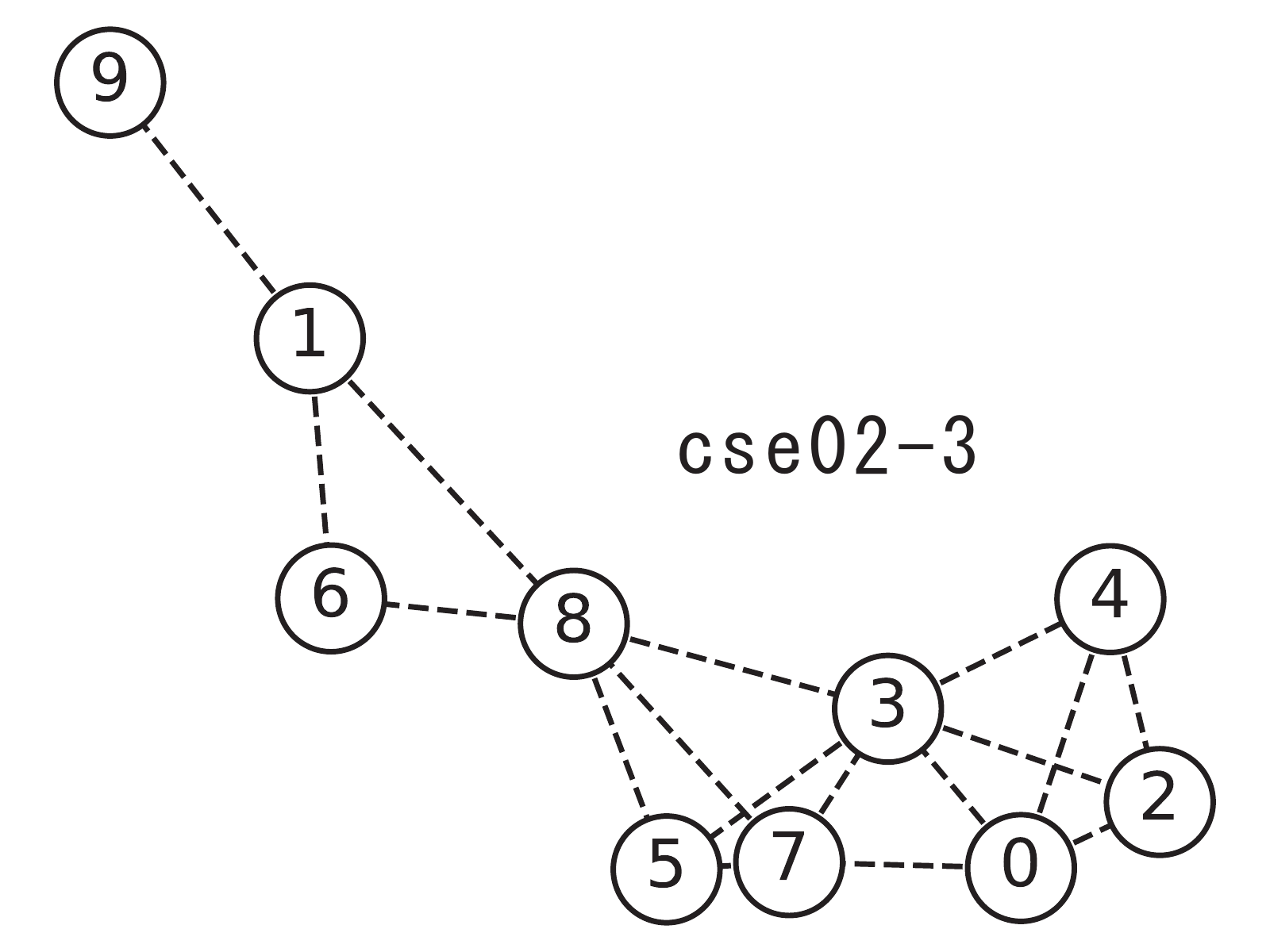}\\
  \end{minipage}
  \begin{minipage}[b]{0.22\textwidth}
  \raggedright
  \includegraphics[keepaspectratio, scale=0.25]{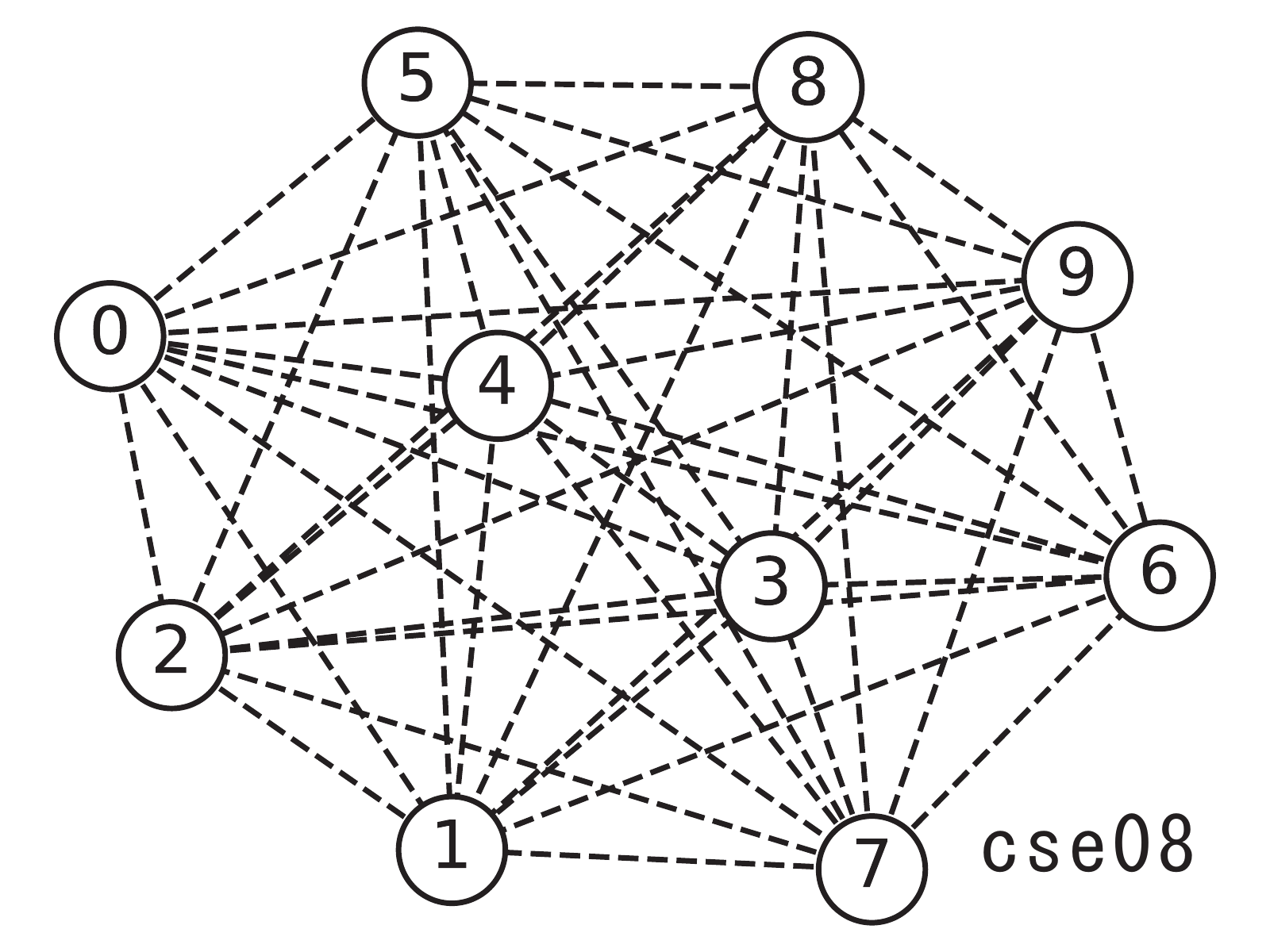}\\
  \end{minipage}

\caption{\label{fig:NetworkModel}Examples of node-to-node contact pattern used for the evaluation. The solid lines indicate static connections, and the dashed lines indicate opportunistic connections. We have prepared four static networks, and two types of dynamic networks -- (1) random waypoint mobility (RWP) and (2) community structured environment (CSE).}
\end{figure*}

We evaluate the characteristics of WAFL, especially regarding the ability of model generalization in Non-IID and ad hoc scenarios. This evaluation includes the analysis of trained model performance such as test accuracies and reductions of confusion over the IID dataset and the convergences of model parameters on varieties of static and dynamic communication networks. In this paper, we carry out benchmark-based experiments by simulation for understanding these characteristics and for reproductivity.

\subsection{Experiment Setting}

\subsubsection{Non-IID Dataset}
In this evaluation, we have configured 10 training nodes ($n=0, 1, \ldots, 9$) to have 90\% Non-IID MNIST datasets each; i.e., 90\% of node $n$'s data is composed of label $n$'s samples, and the other 10\% is uniformly composed of the other label's samples without any overlaps among the nodes. Table \ref{tab:noniid_distribution} describes the detail of data distributions. The distribution parameter 90\% is more severe than the settings given by \cite{wang2020optimizing}, which was 80\%. 

\subsubsection{Node Mobility}

We have generated node-to-node contact patterns of static and dynamic networks as Fig. \ref{fig:NetworkModel}. We prepared four static networks (1) static\_line, (2) static\_tree, (3) static\_ringstar, and (4) static\_dense, and two types of dynamic networks based on (1) random waypoint (RWP) and (2) community structured environment (CSE). 

RWP mobility \cite{camp2002survey} is often used in wireless ad hoc network simulations. The nodes walk around the given area with posing some duration at temporal locations. This mobility model allows a node to contact with any other nodes in the area. This mobility model greatly fits to the movement of people in a shopping mall or in an amusement park. 

CSE mobility \cite{ochiai2008mobility} simulates node movement within a given structure of communities. CSE is based on the idea that there are several communities in the real society to which people belong, and that they usually meet the people who belong to the same community. CSE assumes that a node belongs to several communities and moves from one community to another. Here, the community is a physical rendezvous point. If two nodes are in the same community at the same time, they can communicate with each other.

In the case of RWP mobility, depending on the size of the node movable areas, we generated three categories of mobility patterns: i.e., rwp0500, rwp1000, and rwp2000. Here, the name indicates the size of the movable area, for example, rwp0500 was generated by simulating node movements in 500 [m]$\times$500 [m] square. We assumed 100 [m] for radio range and 10 [epoch] for posing time. We have chosen the travel speed uniformly randomly from the interval of [3.0, 7.0] [m/epoch].

In the case of CSE mobility, depending on the number of communities a node belongs to, we generated three categories of mobility patterns: i.e., cse02, cse04, and cse08. The number indicates the number of communities a node belongs to, for example, cse02 means that a node belongs to two communities in the world of communities. In this evaluation, we have assumed 10 communities as the world of communities, transit time from one community to another as 10 [epoch], and transit starting probability at a community as 5\%. 

Fig. \ref{fig:NetworkModel} shows topology examples of node-to-node contacts generated in this way and used in our evaluation. The solid lines indicate static connections -- the nodes were always connected and could communicate all the time. The dashed lines indicate opportunistic connections -- the nodes have seen temporal connections and could communicate while the connections were alive.

For the dynamic cases, i.e., RWPs and CSEs, we have generated three mobility patterns for each individual category with different random seeds. For example, as Fig. \ref{fig:NetworkModel} shows, cse02 category has three network topology: cse02-1, cse02-2, and cse02-3.

In this evaluation, we have configured it so that a time unit corresponds to an epoch. A single training epoch is assumed to be completed within a single time unit, e.g., in a minute or 10 seconds. We have carried out simulations of each case up to 5000 epochs.

\subsubsection{Machine Learning Model}

As for the target model in this study, we have used a 2-layer fully connected neural network. The input of an MNIST data sample (28$\times$28) is first flattened into 784 elements, and then the first fully connected layer (fc1) with the output of 128 elements comes. Then, the 128 elements go through the ReLU function, and the second fully connected layer (fc2) gives the output of 10 elements. This model predicts the label of the sample by finding the index of the maximum. We have given random values as the initial parameters before the pre-training. After 50 epochs for pre-training, we applied WAFL. We used CrossEntropy as a loss function and Adam as an optimizer. We have used 32 as the batch size and evaluated with 0.001 and 0.0001 as the learning rate $\eta$. We have also studied with coefficients $\lambda$ = 1.0 and 0.1.

We admit that this model configuration is very basic in today's evolution of machine learning research, but we emphasize again that our evaluation focus is on the study of WAFL on the varieties of node-to-node contact patterns.

\subsubsection{Other Learning Schemes for Comparison}

To compare the performance of WAFL, we have also evaluated developed models by self-training, federated learning, and IPLS\cite{pappas2021ipls}.

In self-training, in this paper, we assumed that nodes are closed and make training lonely. They do not make any communications including a model exchange with others. The nodes make continuous training with the local Non-IID data they have individually. In this training scheme, they don't have coefficient $\lambda$, but it has a learning rate $\eta$. We have tested 90\% and 80\% Non-IID cases to check the severity of the 90\% Non-IID case.

In federated learning, we assumed a parameter server virtually that collects local models from all the nodes, aggregates them to the global model, and redistributes the aggregated model to the nodes. In this training scheme, they have coefficient $\lambda$ and learning rate $\eta$.

As for IPLS, the original framework assumes Internet connectivity and pub/sub functionality as a communication backbone for decentralized federated learning. We adjusted the framework to wireless ad hoc networking scenarios for fair comparisons. We implemented the key idea of IPLS, i.e., responsibility, model partitioning, and distribution, and we evaluated our scenario. We have set the number of partitions $K = 2$, the minimum partitions per node $\pi = 1$, and the maximum replicas per partition $\rho=2$ and $3$.

In the above training methods, we have initialized model parameters at random and carried out 50 epoch pre-training as the other cases do. Other settings (i.e., data distributions, model structure, loss functions, Adam optimizer) were also the same.

\subsection{Performance Overview}

\begin{table*}
\caption{Performance evaluation of WAFL on different contact patterns with coefficient $\lambda$ and learning rate $\eta$.\\ Compared to self-train and IPLS, WAFL has achieved better performance, almost nearly the conventional federated learning. }
\label{tab:accuracy}
    \centering
    \begin{tabular}{l|cccc}

    \hline
    \hline
        \textbf{Experiment Settings} & 
        \textbf{Accuracy[\%]}&\textbf{Precision[\%]}&\textbf{Recall[\%]}&\textbf{F1-score[\%]} \\
        
    \hline
        WAFL-static\_line ($\lambda$=1.0, $\eta$=0.001)&\textbf{96.337}$\pm$0.595&96.388$\pm$2.429&96.307$\pm$2.066&96.313$\pm$1.464 \\
        WAFL-static\_line ($\lambda$=1.0, $\eta$=0.0001)&87.454$\pm$3.253&90.754$\pm$11.869&87.350$\pm$9.208&88.020$\pm$7.873\\
        WAFL-static\_line ($\lambda$=0.1, $\eta$=0.001)&95.081$\pm$1.168&95.388$\pm$4.644&95.039$\pm$3.563&95.082$\pm$2.791 \\
        WAFL-static\_line ($\lambda$=0.1, $\eta$=0.0001)&91.950$\pm$1.054&93.128$\pm$8.209&91.877$\pm$5.137&92.108$\pm$4.506 \\
    \hline
        WAFL-static\_tree ($\lambda$=1.0, $\eta$=0.001)&\textbf{95.906}$\pm$1.196&96.023$\pm$3.242&95.861$\pm$2.900&95.880$\pm$2.254 \\
        WAFL-static\_tree ($\lambda$=1.0, $\eta$=0.0001)&88.568$\pm$1.610&91.211$\pm$11.115&88.472$\pm$7.750&89.012$\pm$6.716 \\
        WAFL-static\_tree ($\lambda$=0.1, $\eta$=0.001)&95.389$\pm$1.080&95.640$\pm$4.334&95.333$\pm$3.152&95.378$\pm$2.491 \\
        WAFL-static\_tree ($\lambda$=0.1, $\eta$=0.0001)&91.800$\pm$1.099&93.038$\pm$8.387&91.714$\pm$5.361&91.961$\pm$4.691 \\       
    \hline
        WAFL-static\_ringstar ($\lambda$=1.0, $\eta$=0.001)&\textbf{96.110}$\pm$0.745&96.181$\pm$2.747&96.066$\pm$2.458&96.082$\pm$1.892 \\
        WAFL-static\_ringstar ($\lambda$=1.0, $\eta$=0.0001)&87.391$\pm$2.889&90.671$\pm$11.913&87.253$\pm$9.489&87.905$\pm$8.034 \\
        WAFL-static\_ringstar ($\lambda$=0.1, $\eta$=0.001)&95.267$\pm$0.773&95.518$\pm$4.433&95.217$\pm$2.979&95.262$\pm$2.410 \\
        WAFL-static\_ringstar ($\lambda$=0.1, $\eta$=0.0001)&91.644$\pm$1.215&92.970$\pm$8.688&91.540$\pm$5.542&91.814$\pm$4.937 \\       
    \hline
        WAFL-static\_dense
        ($\lambda$=1.0, $\eta$=0.001)&95.585$\pm$1.376&95.788$\pm$3.886&95.538$\pm$3.513&95.571$\pm$2.757 \\
        WAFL-static\_dense
        ($\lambda$=1.0, $\eta$=0.0001)&86.497$\pm$3.585&90.225$\pm$12.342&86.389$\pm$9.910&87.124$\pm$8.319 \\
        WAFL-static\_dense
        ($\lambda$=0.1, $\eta$=0.001)&\textbf{95.643}$\pm$1.036&95.840$\pm$3.919&95.599$\pm$2.841&95.636$\pm$2.333 \\
        WAFL-static\_dense ($\lambda$=0.1, $\eta$=0.0001)&92.471$\pm$1.099&93.533$\pm$7.894&92.398$\pm$5.114&92.604$\pm$4.447 \\

    \hline
        WAFL-rwp0500
        ($\lambda$=1.0, $\eta$=0.001)&\textbf{95.389}$\pm$1.218&95.655$\pm$4.359&95.349$\pm$3.287&95.395$\pm$2.696 \\
        WAFL-rwp0500
        ($\lambda$=1.0, $\eta$=0.0001)&91.322$\pm$1.445&92.902$\pm$9.266&91.241$\pm$6.055&91.558$\pm$5.446 \\
        WAFL-rwp0500
        ($\lambda$=0.1, $\eta$=0.001)&95.140$\pm$0.859&95.440$\pm$4.749&95.096$\pm$3.175&95.145$\pm$2.584 \\
        WAFL-rwp0500
        ($\lambda$=0.1, $\eta$=0.0001)&92.030$\pm$1.037&93.307$\pm$8.542&91.952$\pm$5.262&92.210$\pm$4.761 \\

    \hline
        WAFL-rwp1000
        ($\lambda$=1.0, $\eta$=0.001)&\textbf{95.230}$\pm$0.981&95.546$\pm$4.725&95.188$\pm$3.400&95.238$\pm$2.622 \\
        WAFL-rwp1000
        ($\lambda$=1.0, $\eta$=0.0001)&91.683$\pm$1.394&93.225$\pm$9.189&91.609$\pm$5.882&91.917$\pm$5.331 \\
        WAFL-rwp1000
        ($\lambda$=0.1, $\eta$=0.001)&94.734$\pm$0.751&95.194$\pm$5.610&94.689$\pm$3.395&94.774$\pm$2.928 \\
        WAFL-rwp1000
        ($\lambda$=0.1, $\eta$=0.0001)&92.563$\pm$1.033&93.790$\pm$8.456&92.484$\pm$4.977&92.741$\pm$4.691 \\       

    \hline
        WAFL-rwp2000
        ($\lambda$=1.0, $\eta$=0.001)&\textbf{93.659}$\pm$0.948&94.403$\pm$6.898&93.602$\pm$4.256&93.738$\pm$3.702 \\
        WAFL-rwp2000
        ($\lambda$=1.0, $\eta$=0.0001)&89.615$\pm$1.746&91.856$\pm$10.554&89.512$\pm$7.118&89.973$\pm$6.389 \\
        WAFL-rwp2000
        ($\lambda$=0.1, $\eta$=0.001)&92.629$\pm$1.004&93.653$\pm$7.803&92.560$\pm$4.838&92.761$\pm$4.274 \\
        WAFL-rwp2000
        ($\lambda$=0.1, $\eta$=0.0001)&89.984$\pm$1.244&91.800$\pm$9.763&89.870$\pm$6.550&90.239$\pm$5.721 \\       

    \hline
        WAFL-cse02
        ($\lambda$=1.0, $\eta$=0.001)&\textbf{94.840}$\pm$0.833&95.204$\pm$5.023&94.786$\pm$3.384&94.852$\pm$2.643 \\
        WAFL-cse02
        ($\lambda$=1.0, $\eta$=0.0001)&90.735$\pm$1.570&92.500$\pm$9.692&90.650$\pm$6.110&91.012$\pm$5.644 \\
        WAFL-cse02
        ($\lambda$=0.1, $\eta$=0.001)&94.118$\pm$1.011&94.740$\pm$6.378&94.058$\pm$3.950&94.177$\pm$3.451 \\
        WAFL-cse02
        ($\lambda$=0.1, $\eta$=0.0001)&92.015$\pm$1.137&93.357$\pm$8.732&91.931$\pm$5.292&92.208$\pm$4.879 \\       

    \hline
        WAFL-cse04
        ($\lambda$=1.0, $\eta$=0.001)&\textbf{95.454}$\pm$1.288&95.792$\pm$4.798&95.418$\pm$3.598&95.473$\pm$2.929 \\
        WAFL-cse04
        ($\lambda$=1.0, $\eta$=0.0001)&91.719$\pm$1.381&93.255$\pm$9.193&91.637$\pm$5.760&91.951$\pm$5.303 \\
        WAFL-cse04
        ($\lambda$=0.1, $\eta$=0.001)&94.918$\pm$0.855&95.337$\pm$5.426&94.869$\pm$3.447&94.945$\pm$2.935 \\
        WAFL-cse04
        ($\lambda$=0.1, $\eta$=0.0001)&92.588$\pm$0.999&93.789$\pm$8.367&92.509$\pm$4.979&92.758$\pm$4.637 \\       
            
    \hline
        WAFL-cse08
        ($\lambda$=1.0, $\eta$=0.001)&\textbf{95.560}$\pm$0.946&95.836$\pm$4.463&95.519$\pm$3.186&95.567$\pm$2.550 \\
        WAFL-cse08
        ($\lambda$=1.0, $\eta$=0.0001)&91.766$\pm$1.379&93.359$\pm$9.329&91.690$\pm$5.715&92.021$\pm$5.361 \\
        WAFL-cse08
        ($\lambda$=0.1, $\eta$=0.001)&94.848$\pm$0.996&95.280$\pm$5.520&94.806$\pm$3.460&94.883$\pm$3.030 \\
        WAFL-cse08
        ($\lambda$=0.1, $\eta$=0.0001)&92.570$\pm$1.007&93.772$\pm$8.381&92.492$\pm$4.918&92.744$\pm$4.629 \\       

    \hline
    \hline
        IPLS-static\_line($\rho$=2)&82.669$\pm$2.726&85.845$\pm$11.667&82.446$\pm$10.369&83.042$\pm$7.854
 \\
        IPLS-static\_line($\rho$=3)&\textbf{84.062}$\pm$3.280&86.741$\pm$11.320&83.898$\pm$10.061&84.349$\pm$7.641 \\
        \hline
        IPLS-static\_tree ($\rho$=2)&81.223$\pm$2.577&85.014$\pm$12.582&81.008$\pm$12.528&81.457$\pm$9.216
 \\
        IPLS-static\_tree ($\rho$=3)&\textbf{82.485}$\pm$1.876&85.184$\pm$11.314&82.254$\pm$11.057&82.533$\pm$7.078
 \\
        \hline
        IPLS-static\_ringstar ($\rho$=2)&\textbf{82.164}$\pm$2.696&85.533$\pm$11.906&81.909$\pm$11.058&82.508$\pm$8.222\\
        IPLS-static\_ringstar ($\rho$=3)&75.811$\pm$4.706&81.242$\pm$15.421&75.552$\pm$16.337&76.236$\pm$13.583\\
        \hline
        IPLS-static\_dense ($\rho$=2)&\textbf{72.762}$\pm$6.423&80.736$\pm$16.914&72.453$\pm$18.295&73.690$\pm$15.059 \\
        IPLS-static\_dense ($\rho$=3)&70.345$\pm$5.261&77.923$\pm$19.447&69.955$\pm$20.828&70.780$\pm$17.947 \\
        \hline
        IPLS-rwp0500        ($\rho$=2)&\textbf{92.035}$\pm$8.897&93.876$\pm$9.760&91.998$\pm$11.390&92.197$\pm$10.495 \\
        IPLS-rwp0500        ($\rho$=3)&88.256$\pm$11.062&91.300$\pm$11.883&88.189$\pm$14.671&88.244$\pm$14.289 \\
        \hline
        IPLS-rwp1000        ($\rho$=2)&\textbf{92.551}$\pm$1.512&93.630$\pm$8.031&92.470$\pm$5.382&92.672$\pm$4.597\\
        IPLS-rwp1000        ($\rho$=3)&92.085$\pm$2.001&93.276$\pm$8.351&92.010$\pm$6.495&92.181$\pm$5.262 \\
        \hline
        IPLS-rwp2000      ($\rho$=2)&\textbf{89.592}$\pm$5.333&91.653$\pm$10.038&89.527$\pm$9.717&89.735$\pm$8.187 \\
        IPLS-rwp2000 ($\rho$=3)&88.963$\pm$3.353&90.891$\pm$10.091&88.852$\pm$9.643&89.023$\pm$7.482\\
    
        \hline
        IPLS-cse02 ($\rho$=2)&\textbf{83.633}$\pm$14.698&86.755$\pm$16.831&83.480$\pm$18.642&83.560$\pm$17.437\\
        IPLS-cse02 ($\rho$=3)&83.606$\pm$14.929&86.641$\pm$17.385&83.469$\pm$18.031&83.639$\pm$17.153\\        
        \hline
        IPLS-cse04 ($\rho$=2)&\textbf{93.406}$\pm$1.264&94.296$\pm$7.395&93.340$\pm$4.603&93.516$\pm$4.135 \\
        IPLS-cse04 ($\rho$=3)&92.755$\pm$1.435&93.700$\pm$7.627&92.695$\pm$5.274&92.851$\pm$4.385\\
        \hline
        IPLS-cse08 ($\rho$=2)&\textbf{93.578}$\pm$1.381&94.461$\pm$7.441&93.513$\pm$4.610&93.685$\pm$4.185 \\
        IPLS-cse08 ($\rho$=3)&92.226$\pm$1.434&93.389$\pm$8.164&92.159$\pm$5.681&92.363$\pm$4.679 \\
    \hline
    \hline
        federated
        ($\lambda$=1.0, $\eta$=0.001)&\textbf{96.756}$\pm$0.928&96.806$\pm$2.518&96.724$\pm$2.433&96.728$\pm$1.876  \\
        federated
        ($\lambda$=1.0, $\eta$=0.0001)&85.882$\pm$4.392&90.577$\pm$13.357&85.760$\pm$10.897&86.748$\pm$9.392  \\
        federated
        ($\lambda$=0.1, $\eta$=0.001)&94.261$\pm$1.681&95.073$\pm$7.074&94.217$\pm$4.816&94.367$\pm$4.355  \\
        federated
        ($\lambda$=0.1, $\eta$=0.0001)&91.204$\pm$1.830&93.220$\pm$10.223&91.117$\pm$6.680&91.534$\pm$6.213
  \\

    \hline
    \hline
    
        self-train
        ($\eta$=0.001, 90\% Non-IID)&\textbf{84.663}$\pm$1.285&86.363$\pm$9.356&84.445$\pm$9.319&84.663$\pm$6.181  \\
        self-train
        ($\eta$=0.0001, 90\% Non-IID)&82.160$\pm$1.543&84.731$\pm$11.003&81.922$\pm$10.684&82.281$\pm$7.253
  \\

    \hline

        self-train
        ($\eta$=0.001, 80\% Non-IID)&\textbf{89.460}$\pm$0.510&89.979$\pm$6.083&89.313$\pm$5.881&89.377$\pm$3.857 \\
        
        self-train       ($\eta$=0.0001, 80\% Non-IID)&86.484$\pm$0.544&87.438$\pm$7.716&86.288$\pm$7.716&86.404$\pm$5.115
    \\

    \hline
    \hline
    \end{tabular}
\end{table*}

\begin{figure*}
\centering

 \begin{minipage}[b]{0.32\textwidth}
  \raggedright
  \includegraphics[keepaspectratio, scale=0.38]{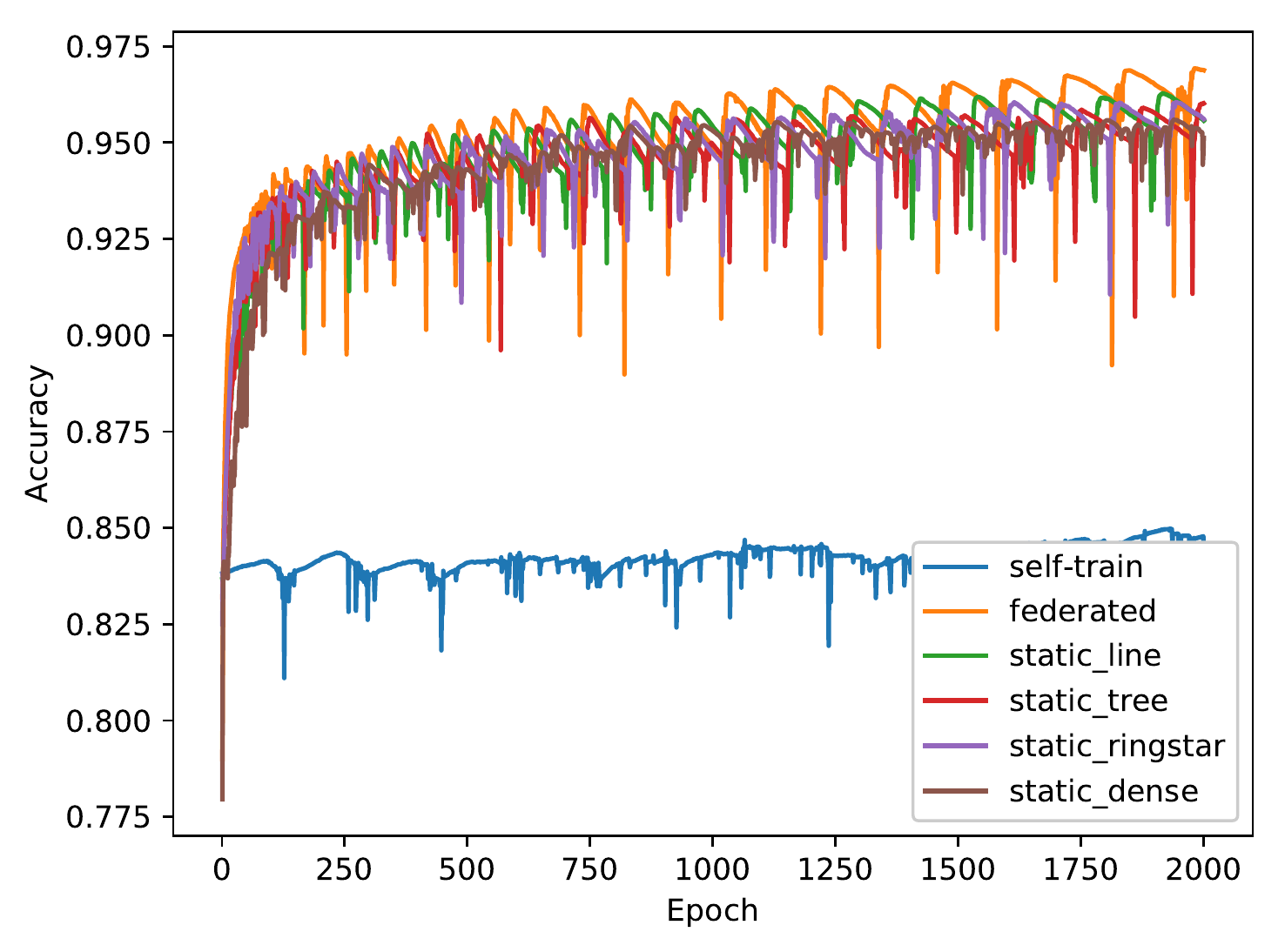}\\
  \scriptsize{\hspace{5em}(a) Static Network (upto 2000 epochs)}
  \end{minipage}
  \begin{minipage}[b]{0.32\textwidth}
  \raggedright
  \includegraphics[keepaspectratio, scale=0.38]{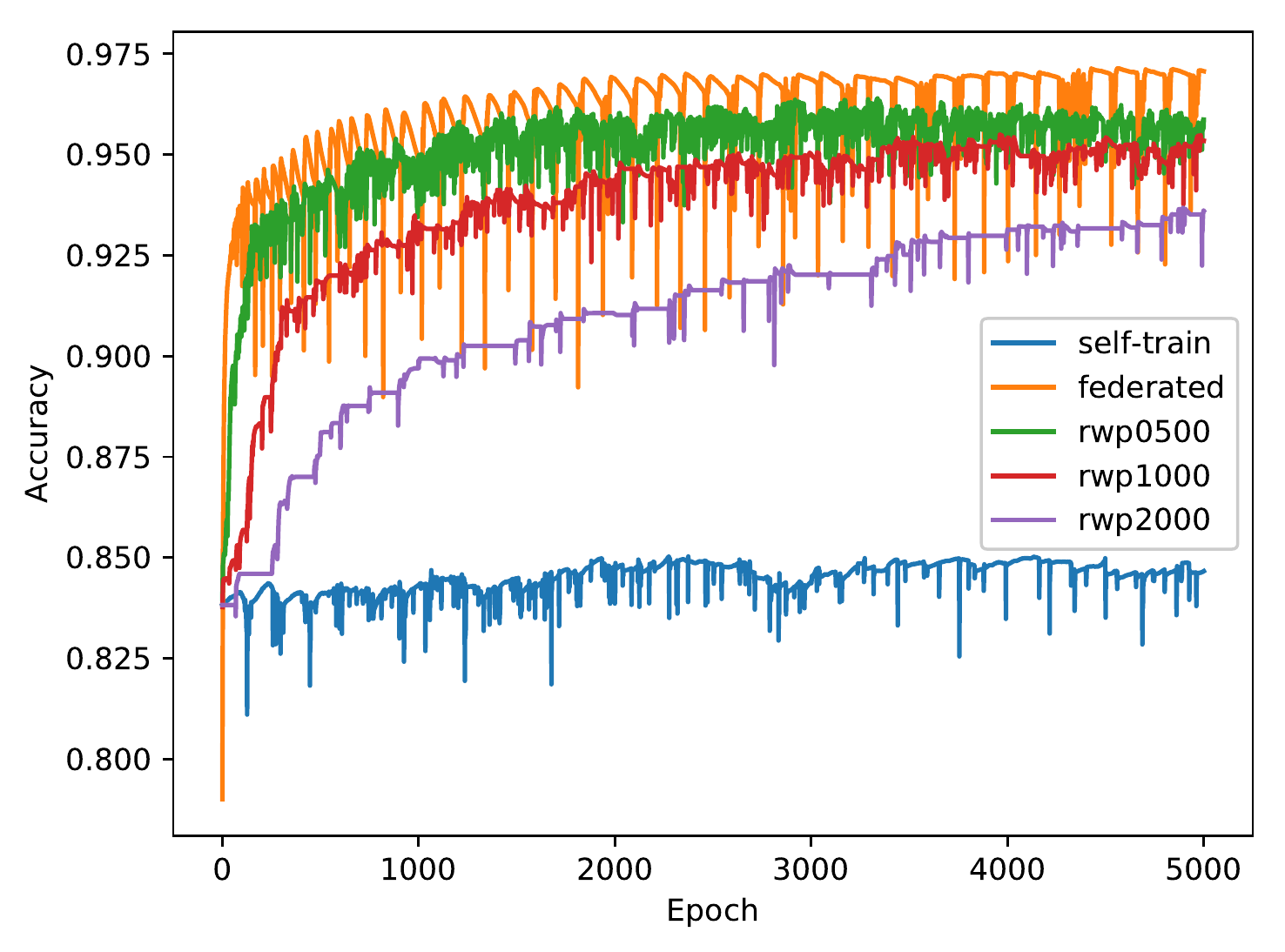}\\
  \scriptsize{\hspace{5em}(b) Dynamic Network (RWP Mobility)}
  \end{minipage}
  \begin{minipage}[b]{0.32\textwidth}
  \raggedright
  \includegraphics[keepaspectratio, scale=0.38]{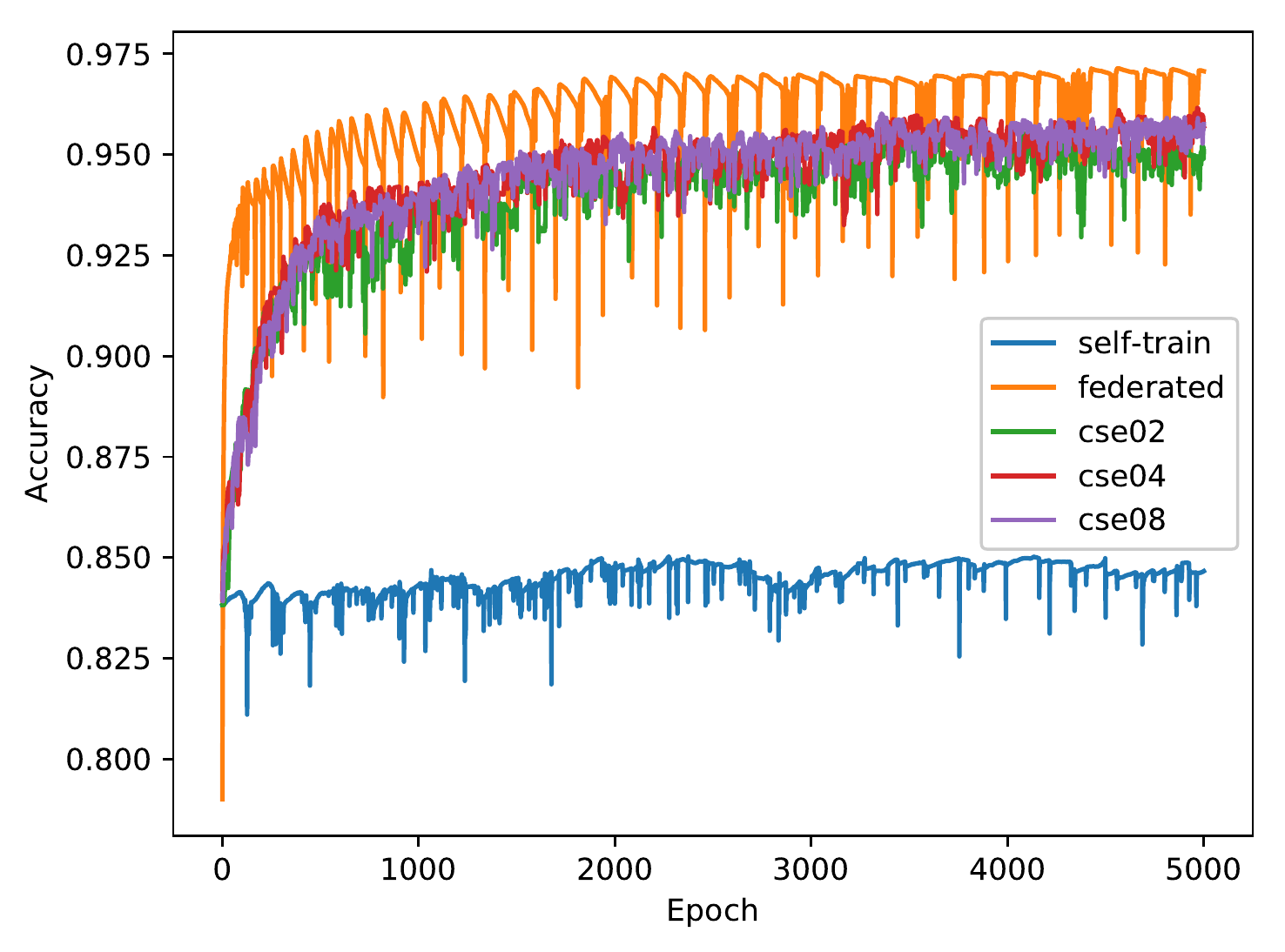}\\
  \scriptsize{\hspace{5em}(c) Dynamic Network (CSE Mobility)}
  \end{minipage}
  
\caption{Accuracy trend on (a) static network, (b) RWP mobility, and (c) CSE mobility with the best hyperparameters (Table \ref{tab:accuracy}). The static network allowed fast training as the federated learning did. We could observe slower training speeds in dynamic cases. This is because nodes skipped training while they were not in contact.}
\label{fig:Accuracy}
\end{figure*}

Table \ref{tab:accuracy} shows the performance of WAFL on the test data on different contact patterns, coefficient $\lambda$, and learning rate $\eta$. 

For this analysis, we have taken the models of the last 100 epochs, i.e., from epoch 4901 to 5000, and calculated averaged accuracy, precision, recall and F1-score with the test MNIST dataset in the following manner. As this is a multi-class classification task, we have first calculated those metrics by each node, each class, and each epoch, which we call micro-accuracy, micro-precision, micro-recall, and micro-F1-score. Then, we got 10$\times$10$\times$100 = 10000 micro-accuracies for each static case. We did the same calculations for the other metrics, i.e., precision, recall, and F1-score. As for the dynamic cases, we got 10$\times$10$\times$100$\times$3 = 30000 micro-accuracies because we have three contact patterns for each case. The discussion of this approach - the arithmetic mean over individual micro-metrics is provided in \cite{opitz2019macro}. Here, the error values on the table indicate the standard deviation of the micro-accuracies, precisions, recalls, and F1 scores. 

From these results, we have observed that self-train could achieve only 84.7\% accuracy whereas WAFL overall (excluding rwp2000) could achieve around 94.8-96.3\%, which is close to the conventional federated learning case 96.8\%, improving 10.2-11.7\% from the self-train (i.e., lonely) case. Please see the self-train (90\% Non-IID) case, here for a fair comparison. The accuracy of rwp2000 is low because of the sparseness and potentially fewer contacts among the nodes compared to the other cases, which we describe in the next subsection.

We observed that static cases achieved higher accuracy (96.0\% on average) than dynamic cases did (95.3\% on average, excluding rwp2000). The differences were only 0.7\%, meaning that WAFL is tolerant of network topology changes, especially in the opportunistic network cases that we consider.

\begin{figure*}
\centering
\includegraphics[width=0.9\textwidth]{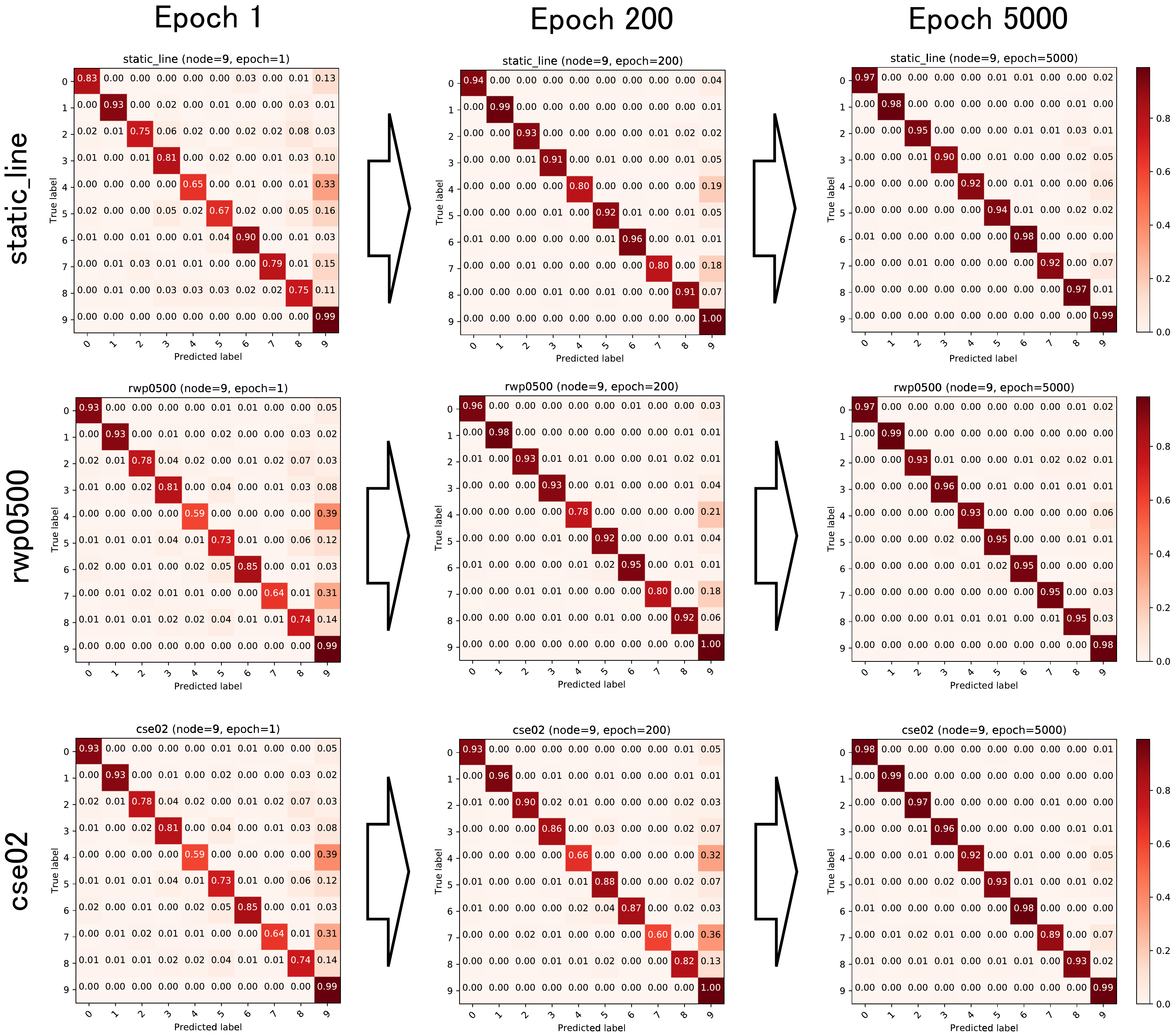}
\caption{Reductions of confusion in static and dynamic contact examples. These results show the confusion matrices of node 9 with the test data which is true labels uniformly randomly. At the beginning of epoch 1, the model had confusion, especially for labels 4 and 8. These confusions decreased gradually, and the model could improve overall classification accuracy finally at epoch 5000. \label{fig:confusion_matrix}}
\end{figure*}

Regarding coefficient $\lambda$ and learning rate $\eta$, we observed that except static-dense, in both static and dynamic networks including federated learning provided better accuracy with $\lambda=1.0$ rather than $\lambda=0.1$. Lower learning rates did not contribute to higher accuracy.



As an ablation, we have also studied the accuracies of the models trained without local mini-batches (i.e., aggregation only). However, we have found that aggregation-only did not contribute to the improvement of performance from self-train, for example, the accuracies were 77.2\% (static\_line), 78.5\% (static\_dense), 70.7\% (rwp0500), and 77.9\% (cse02).

Regarding the comparison to the IPLS framework, they have achieved accuracies of 82.2\%-93.6\% (and 72.8\% in static\_dense), which was lower than WAFL in all the cases. The reason behind this result may be the Non-IID features of the training dataset. This indicates that model partitioning and responsibility management would not be a good strategy in the case of Non-IID.

Please note that in these experiments we have also confirmed the severity of the 90\% Non-IID. With 80\% Non-IID, self-training could achieve the accuracy of 89.5\% but with 90\% Non-IID, it could achieve only 84.7\%, i.e., about 4.8\% difference was observed.

\begin{figure*}
\centering

  \begin{minipage}[b]{0.30\textwidth}
  \raggedright
  \includegraphics[keepaspectratio, scale=0.36]{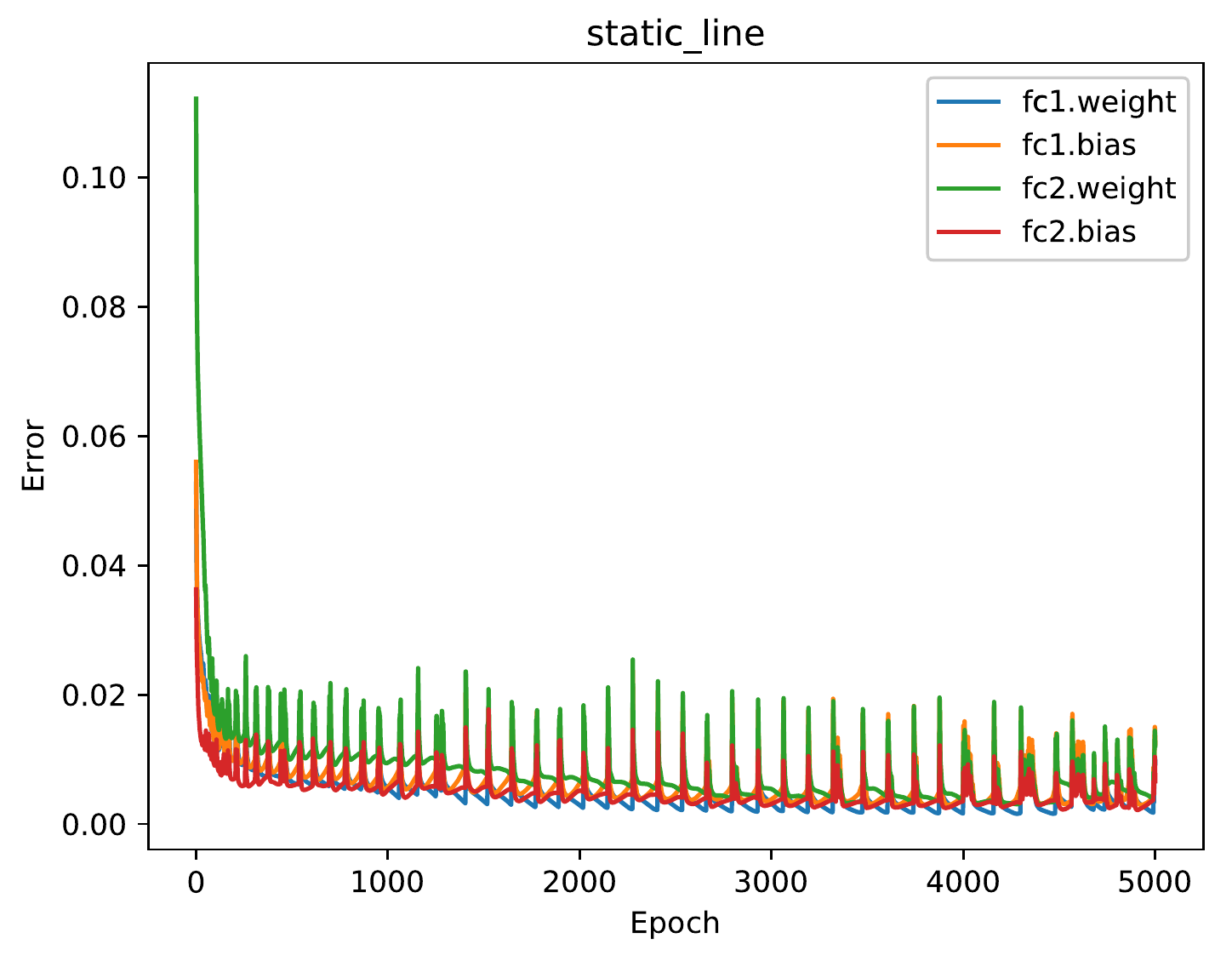}\\
  \end{minipage}
  \begin{minipage}[b]{0.30\textwidth}
  \raggedright
  \includegraphics[keepaspectratio, scale=0.36]{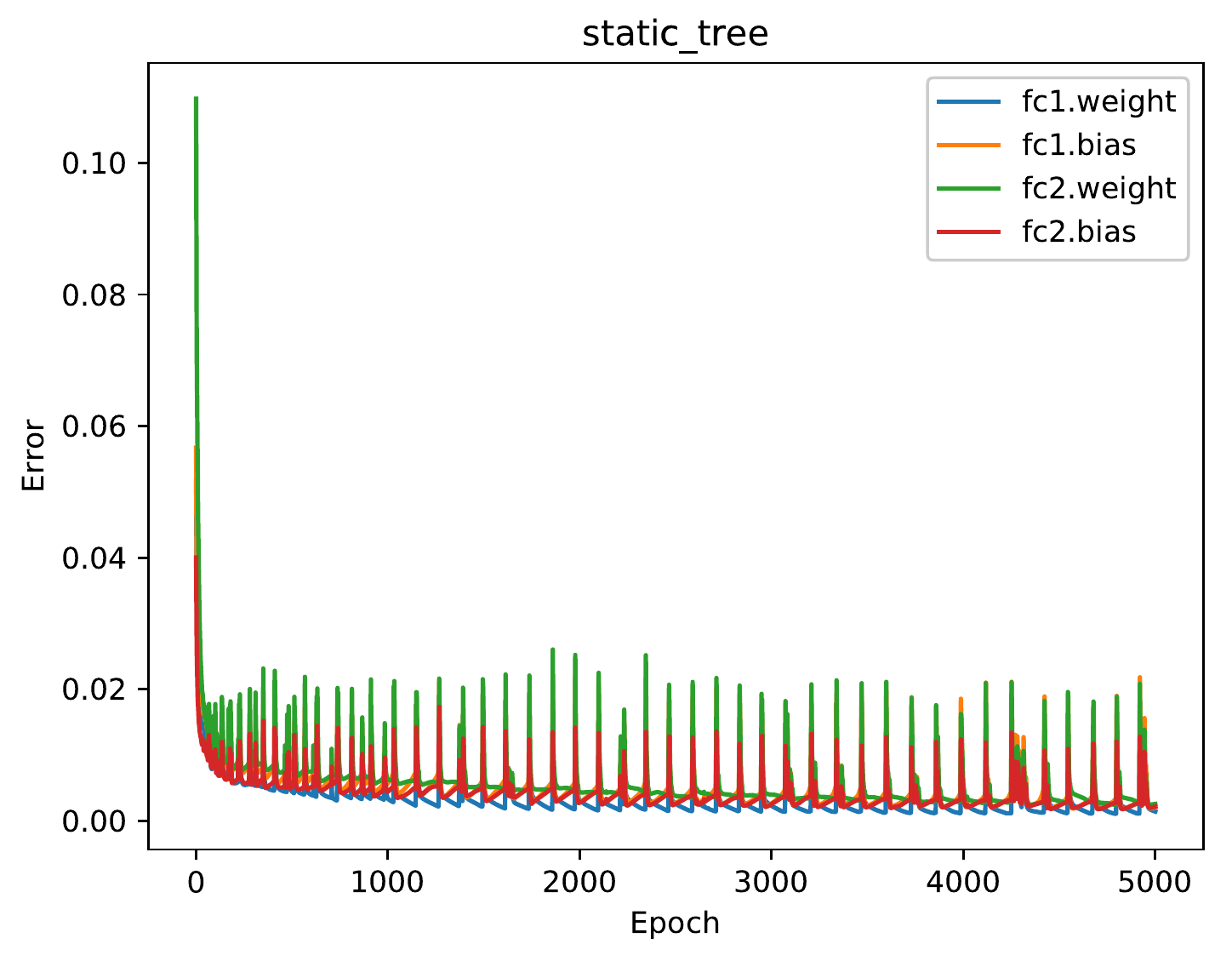}\\
  \end{minipage}
  \begin{minipage}[b]{0.30\textwidth}
  \raggedright
  \includegraphics[keepaspectratio, scale=0.36]{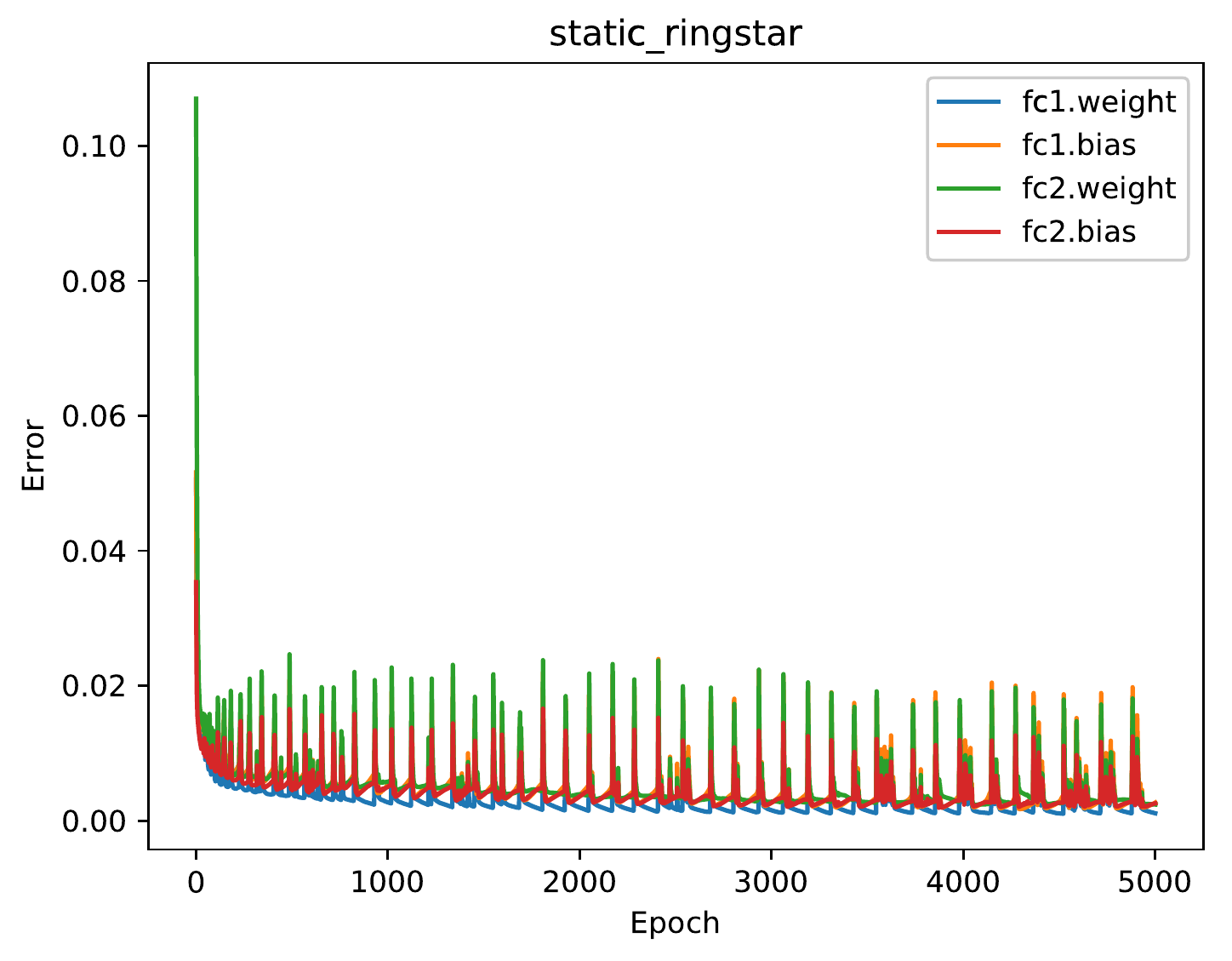}\\
  \end{minipage}

  \vspace{0.5em}

  \begin{minipage}[b]{0.30\textwidth}
  \raggedright
  \includegraphics[keepaspectratio, scale=0.36]{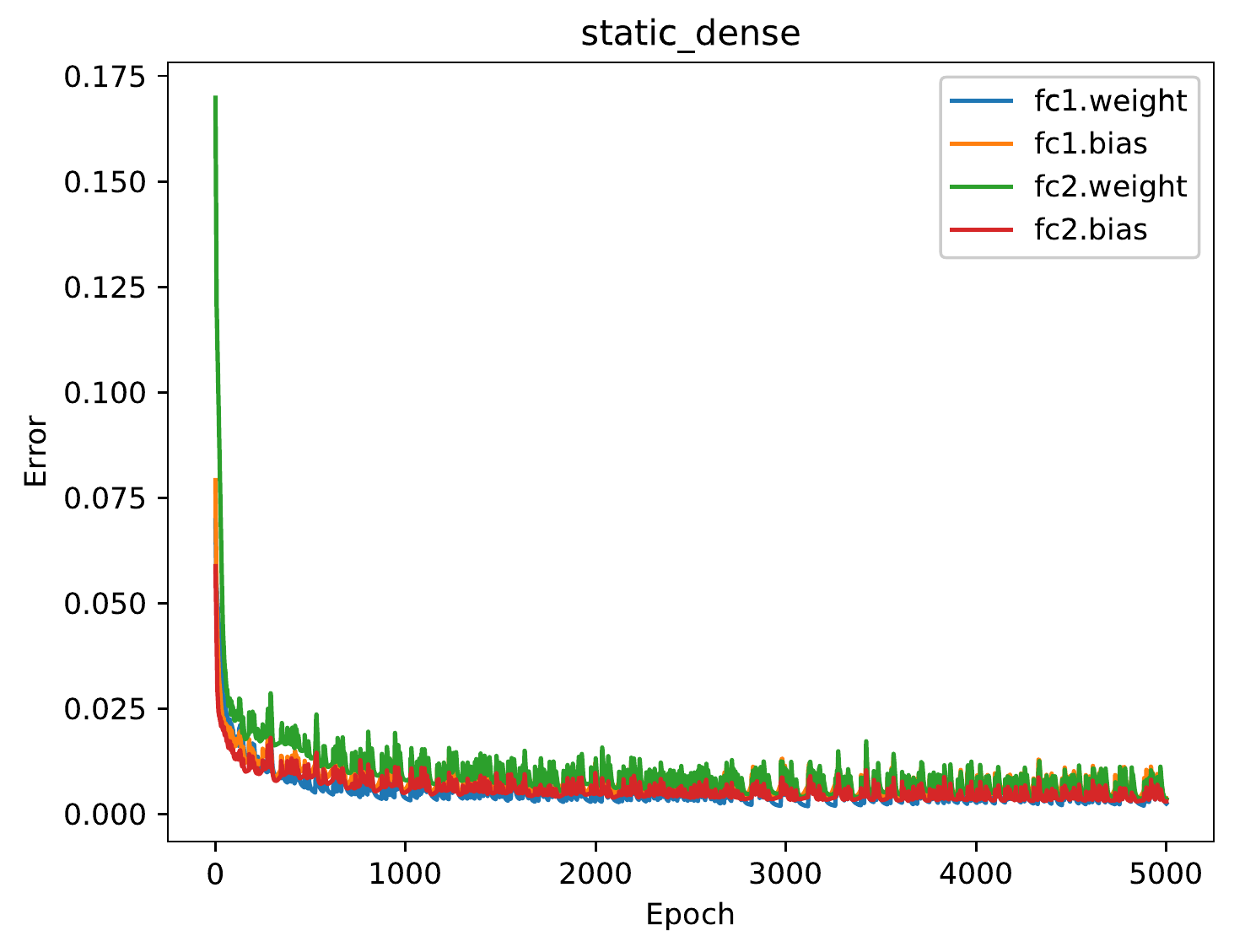}\\
  \end{minipage}
  \begin{minipage}[b]{0.30\textwidth}
  \raggedright
  \includegraphics[keepaspectratio, scale=0.36]{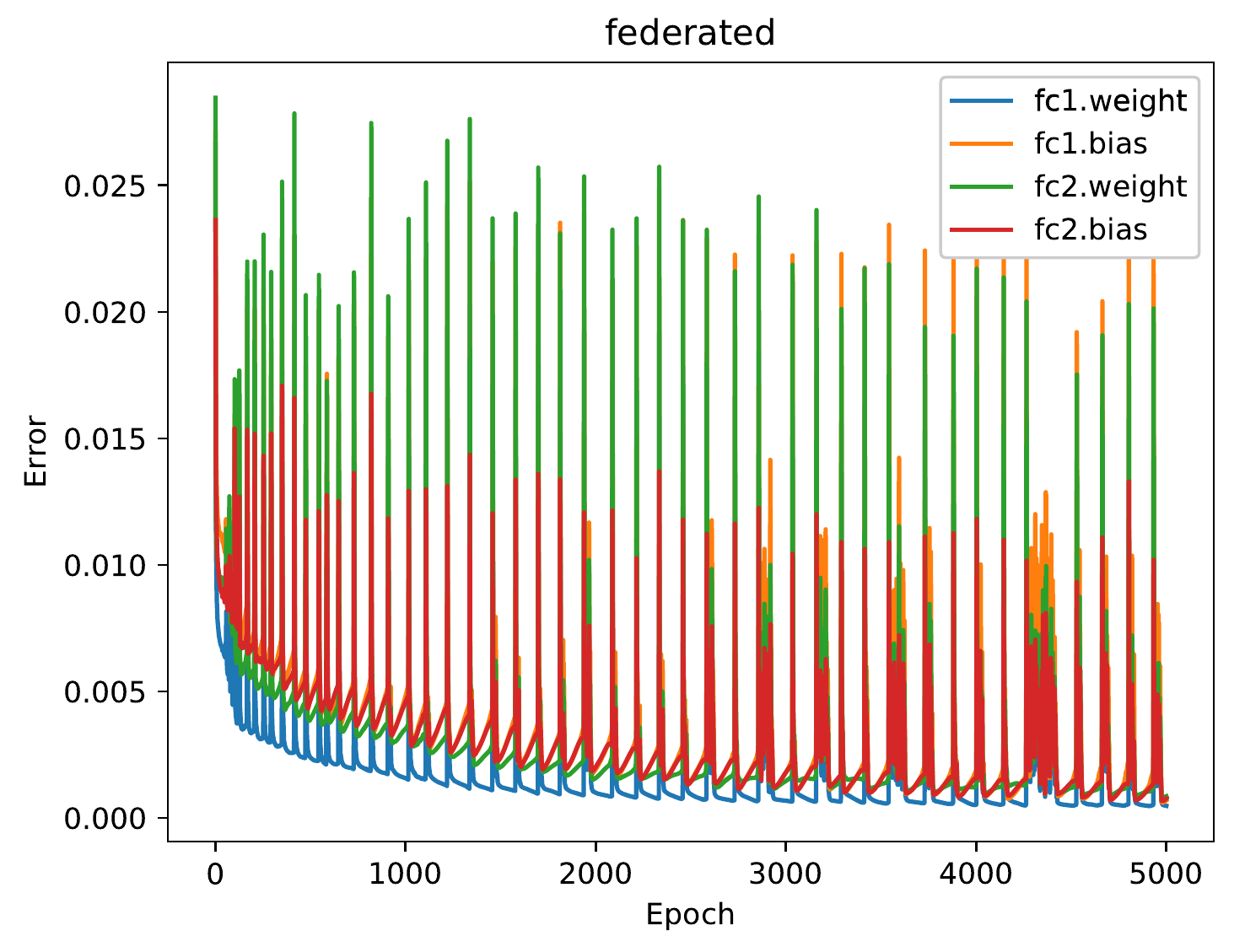}\\
  \end{minipage}
  \begin{minipage}[b]{0.30\textwidth}
  \raggedright
  \includegraphics[keepaspectratio, scale=0.36]{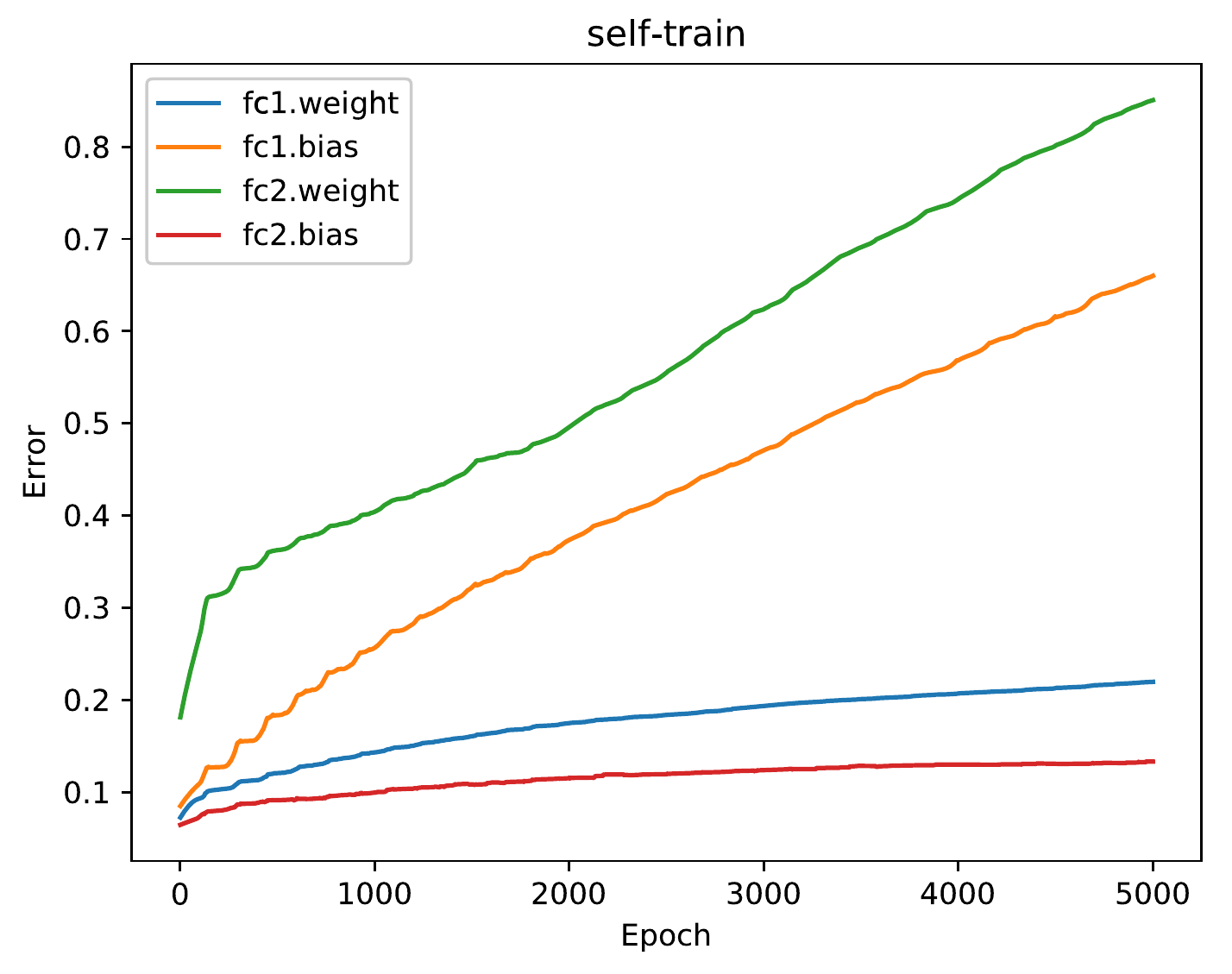}\\
  \end{minipage}

  \vspace{0.5em}

  \begin{minipage}[b]{0.30\textwidth}
  \raggedright
  \includegraphics[keepaspectratio, scale=0.36]{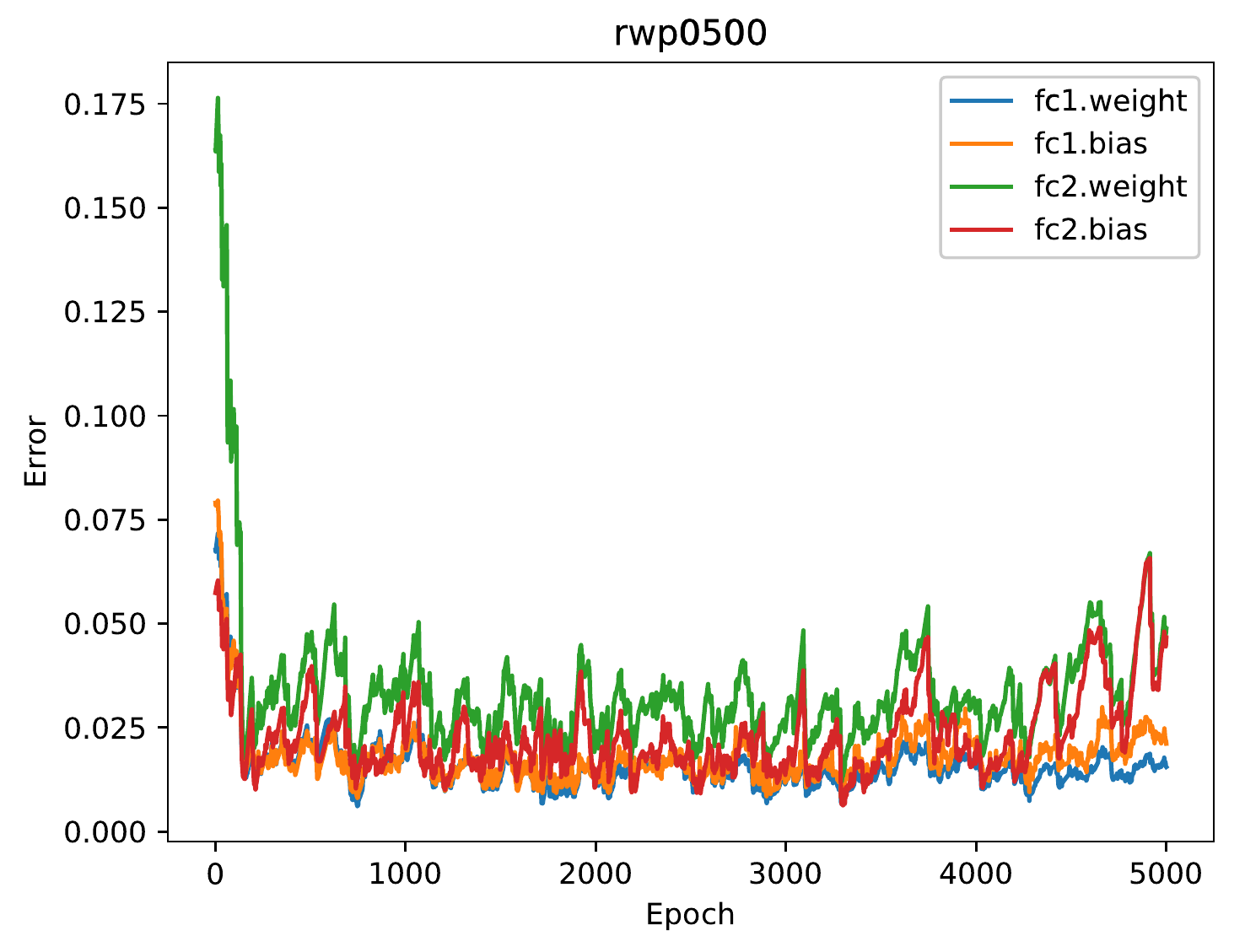}\\
  \end{minipage}
  \begin{minipage}[b]{0.30\textwidth}
  \raggedright
  \includegraphics[keepaspectratio, scale=0.36]{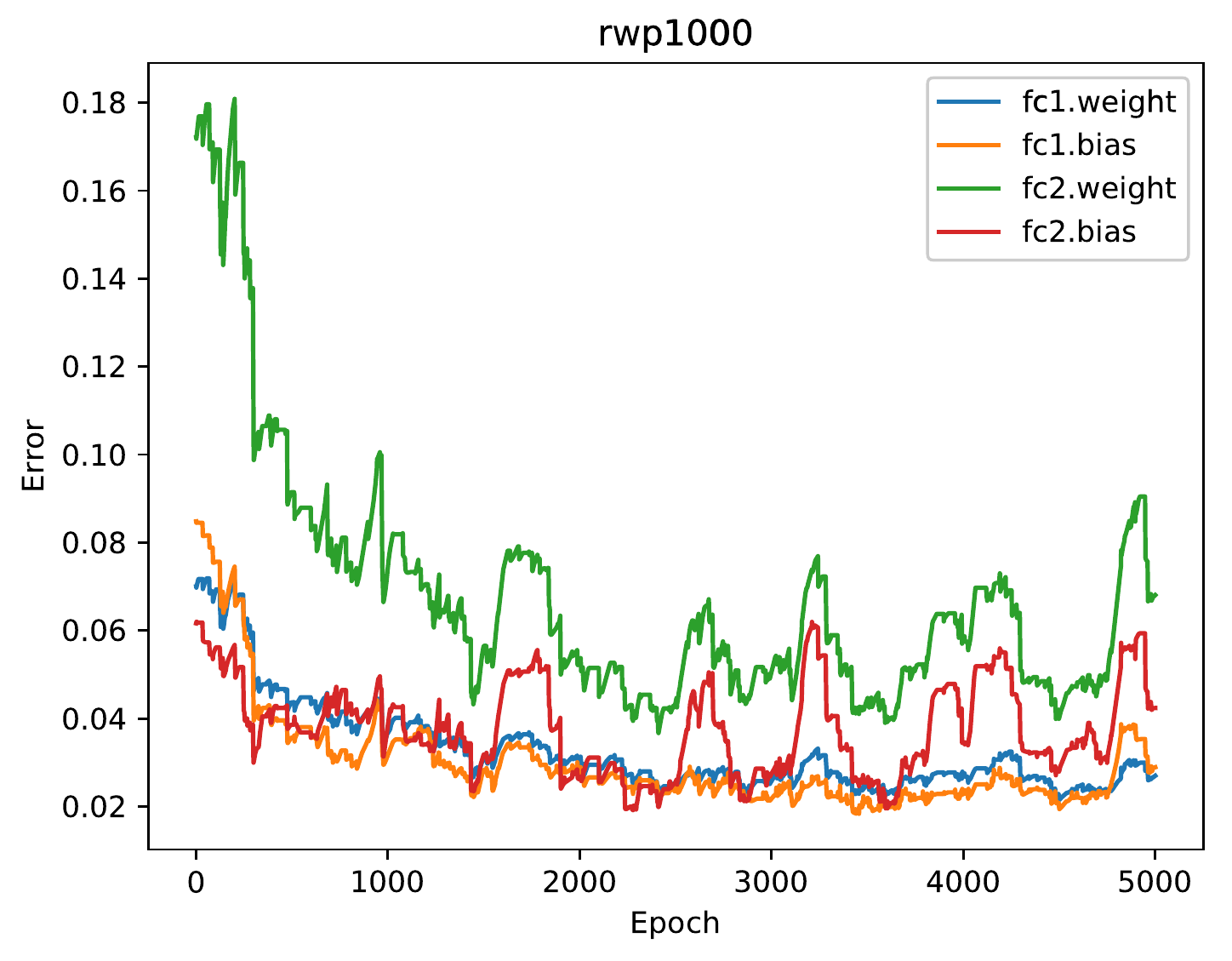}\\
  \end{minipage}
  \begin{minipage}[b]{0.30\textwidth}
  \raggedright
  \includegraphics[keepaspectratio, scale=0.36]{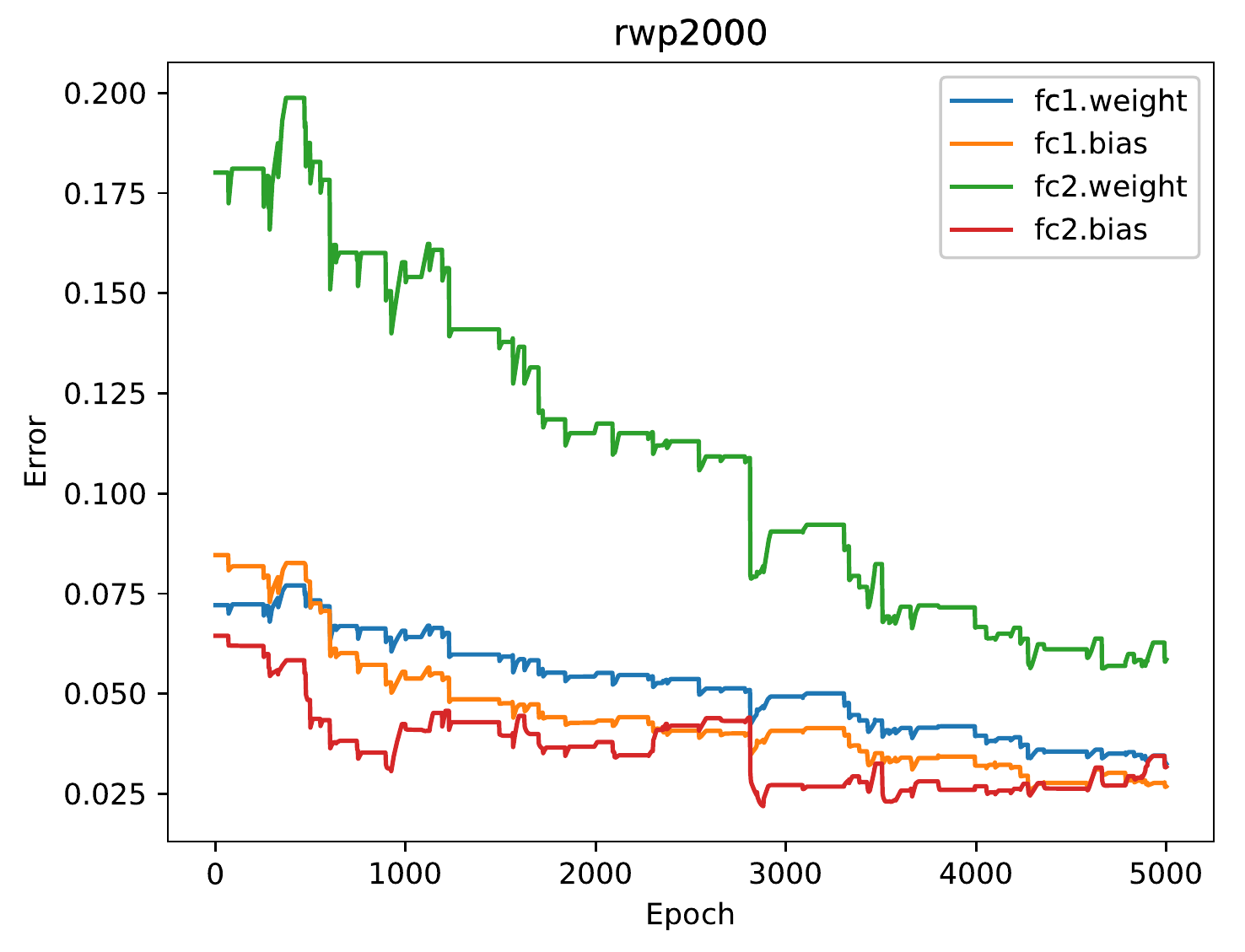}\\
  \end{minipage}

  \vspace{0.5em}

  \begin{minipage}[b]{0.30\textwidth}
  \raggedright
  \includegraphics[keepaspectratio, scale=0.36]{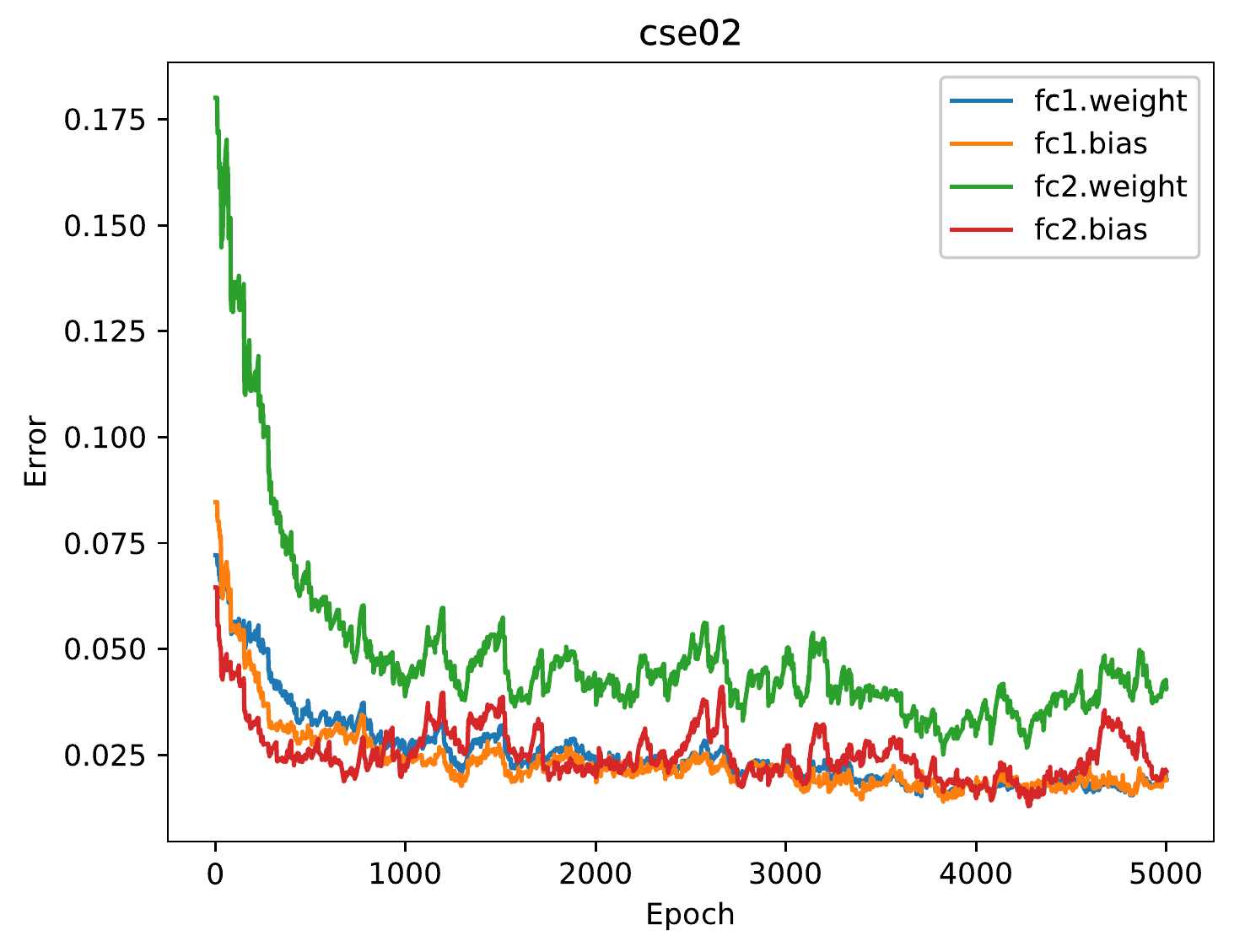}\\
  \end{minipage}
  \begin{minipage}[b]{0.30\textwidth}
  \raggedright
  \includegraphics[keepaspectratio, scale=0.36]{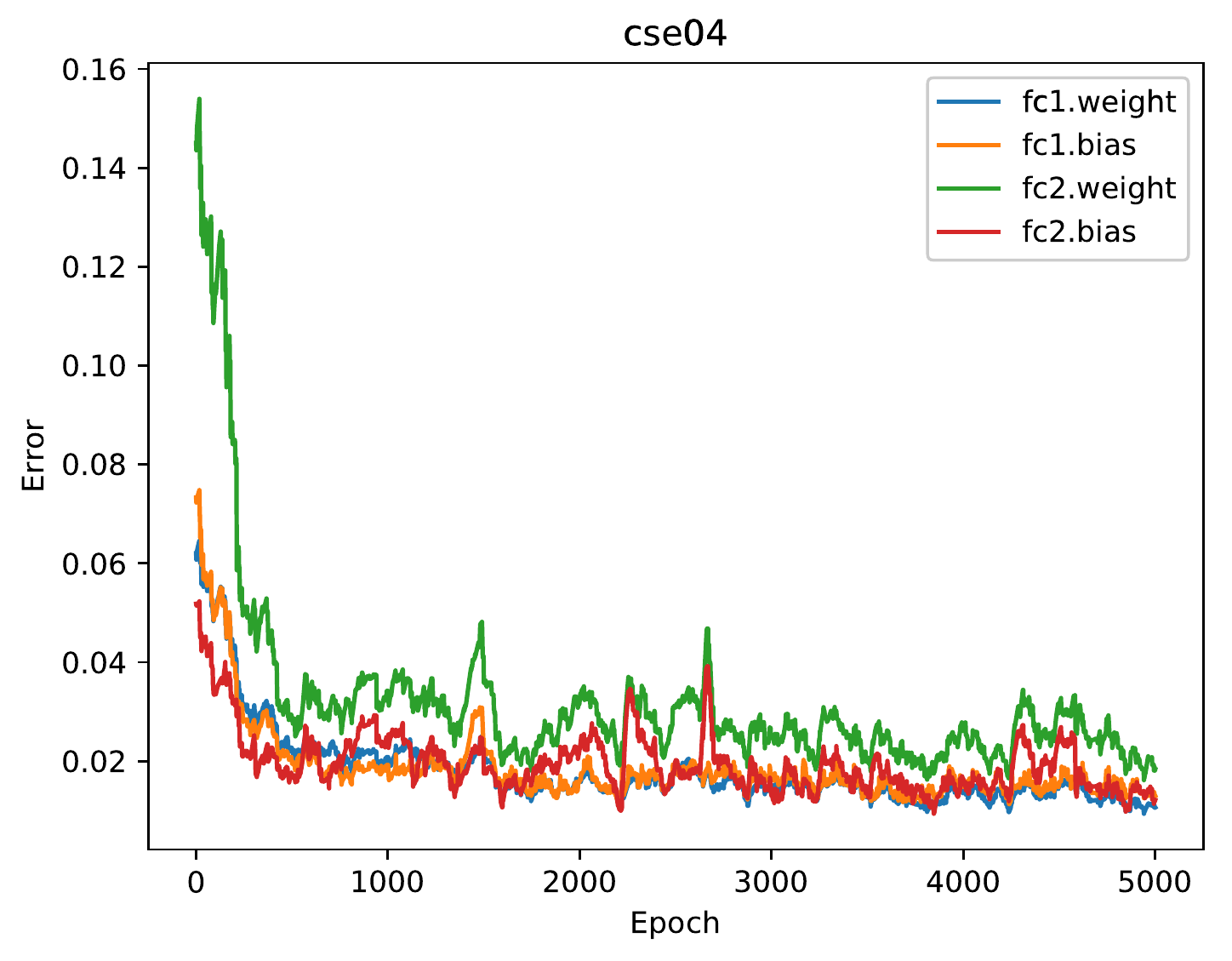}\\
  \end{minipage}
  \begin{minipage}[b]{0.30\textwidth}
  \raggedright
  \includegraphics[keepaspectratio, scale=0.36]{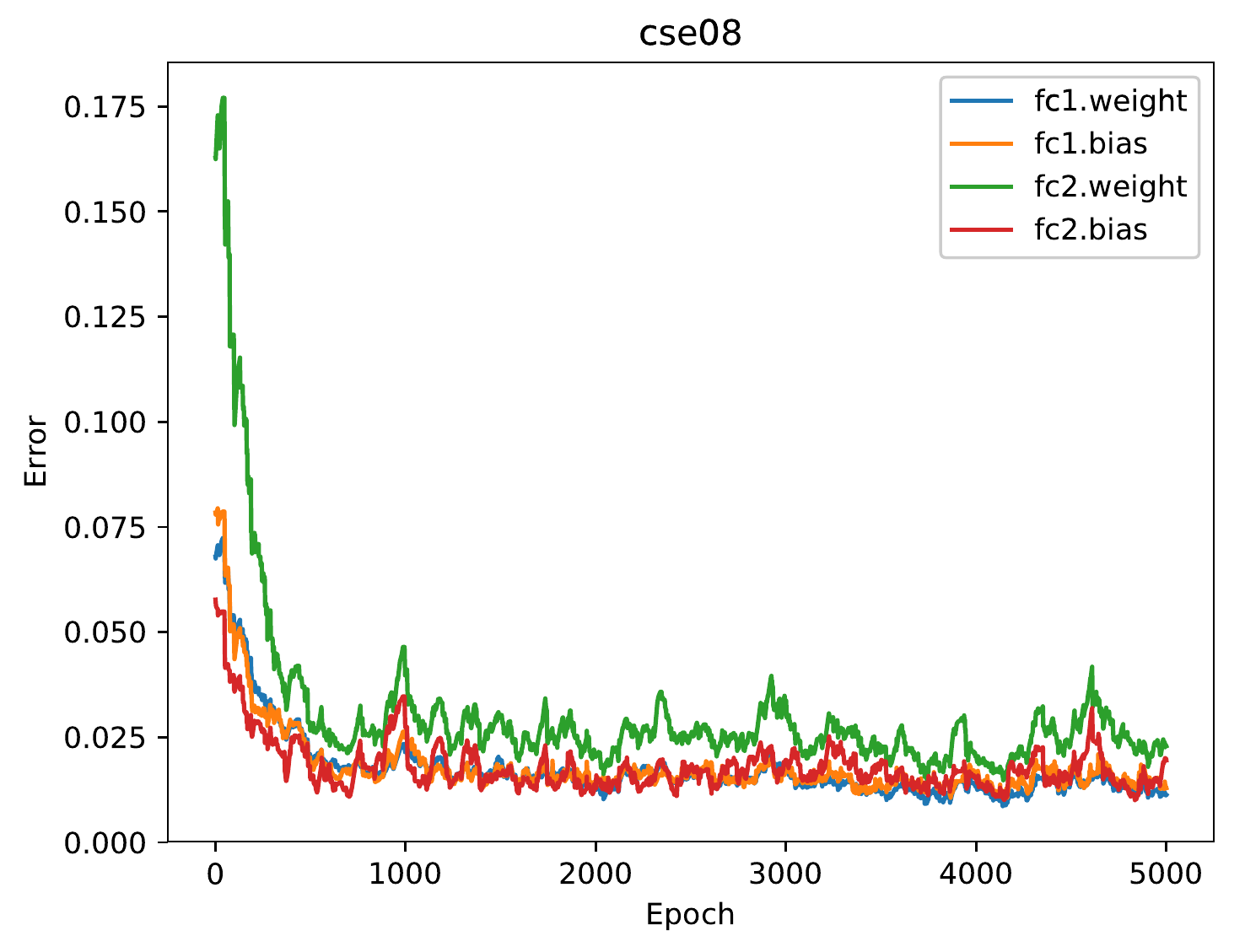}\\
  \end{minipage}

\caption{\label{fig:ModelDistance} The trends of model convergence errors. The static cases and the conventional federated learning provided similar trends including the values of the error. The dynamic cases also allowed gradual error reductions, indicating model aggregation in WAFL has enabled cooperative learning. If not cooperative, i.e., in the self-training case, the convergence errors increased.}
\end{figure*}

\begin{figure*}
\centering

  \begin{minipage}[b]{0.30\textwidth}
  \raggedright
  \includegraphics[keepaspectratio, scale=0.36]{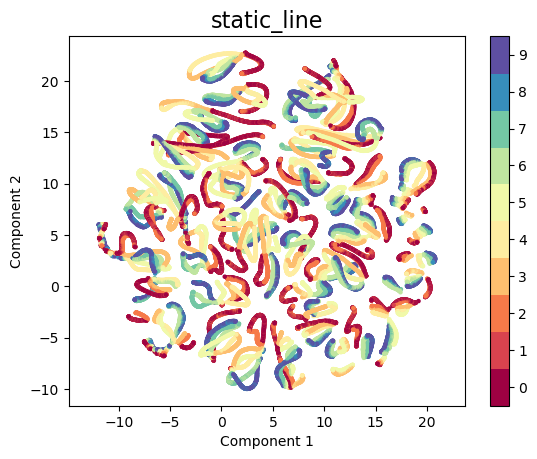}\\
  \end{minipage}
  \begin{minipage}[b]{0.30\textwidth}
  \raggedright
  \includegraphics[keepaspectratio, scale=0.36]{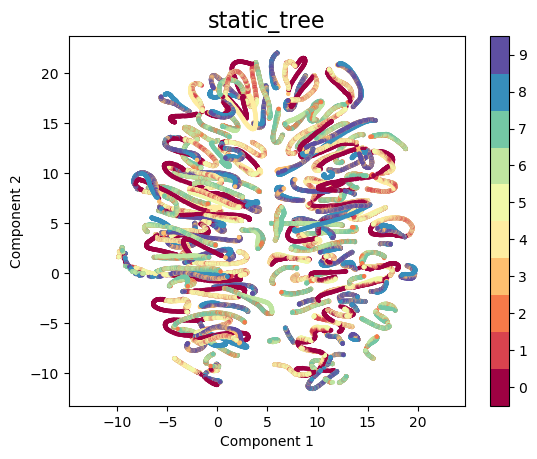}\\
  \end{minipage}
  \begin{minipage}[b]{0.30\textwidth}
  \raggedright
  \includegraphics[keepaspectratio, scale=0.36]{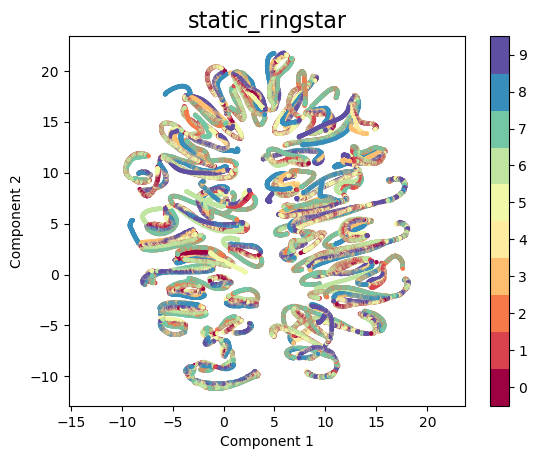}\\
  \end{minipage}

  \vspace{0.5em}
  \begin{minipage}[b]{0.30\textwidth}
  \raggedright
  \includegraphics[keepaspectratio, scale=0.36]{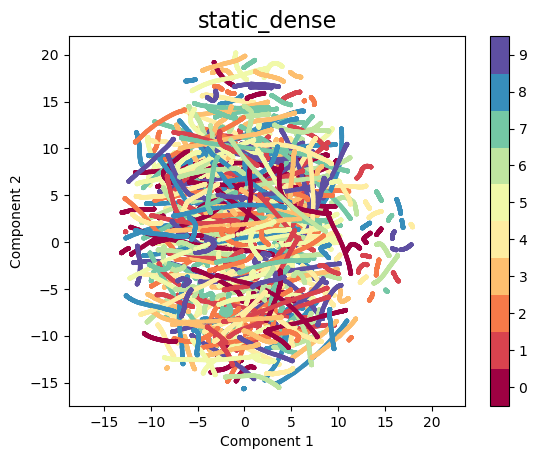}\\
  \end{minipage}
  \begin{minipage}[b]{0.30\textwidth}
  \raggedright
  \includegraphics[keepaspectratio, scale=0.36]{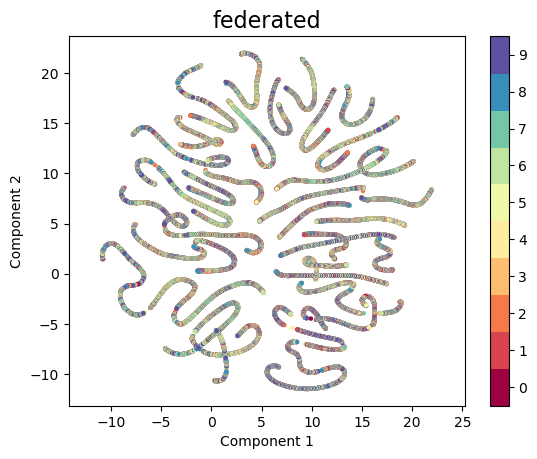}\\
  \end{minipage}
  \begin{minipage}[b]{0.30\textwidth}
  \raggedright
  \includegraphics[keepaspectratio, scale=0.36]{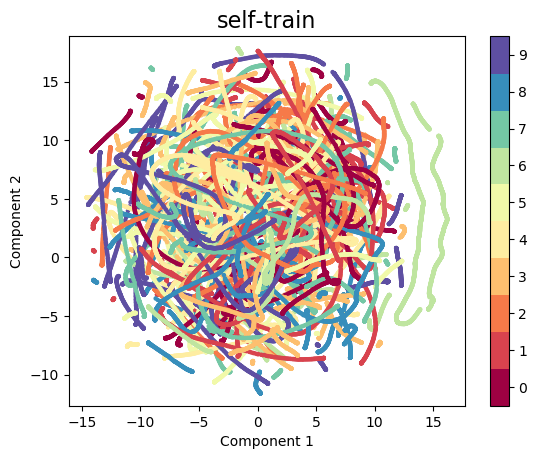}\\
  \end{minipage}

  \vspace{0.5em}

  \begin{minipage}[b]{0.30\textwidth}
  \raggedright
  \includegraphics[keepaspectratio, scale=0.36]{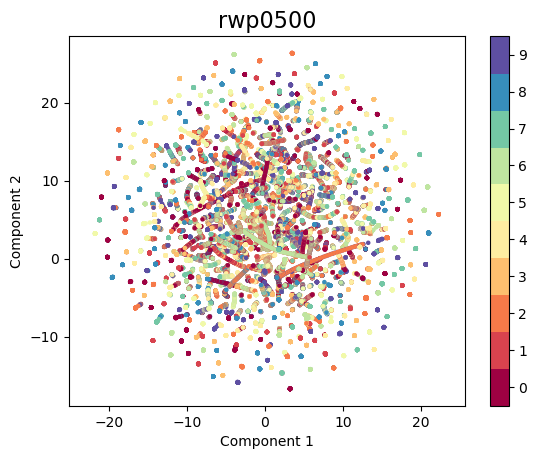}\\
  \end{minipage}
  \begin{minipage}[b]{0.30\textwidth}
  \raggedright
  \includegraphics[keepaspectratio, scale=0.36]{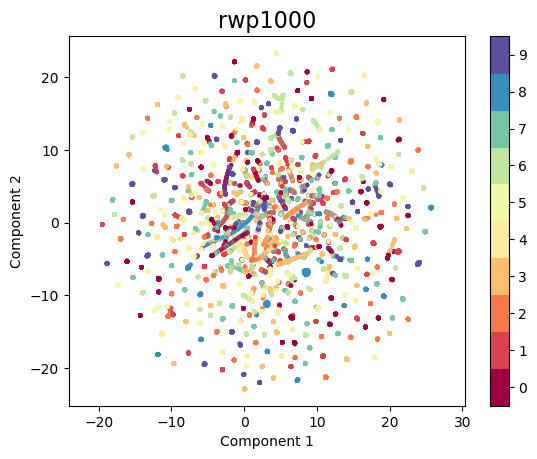}\\
  \end{minipage}
  \begin{minipage}[b]{0.30\textwidth}
  \raggedright
  \includegraphics[keepaspectratio, scale=0.36]{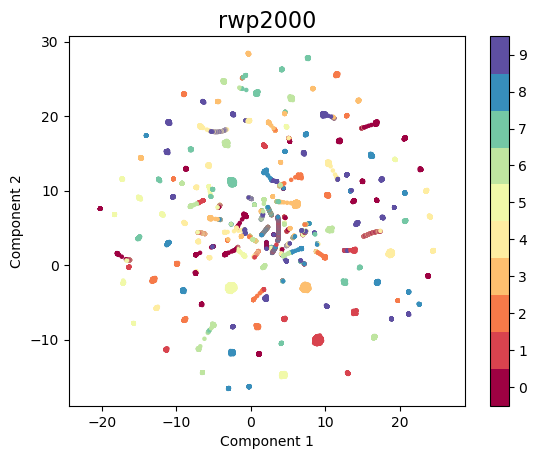}\\
  \end{minipage}

  \vspace{0.5em}

  \begin{minipage}[b]{0.30\textwidth}
  \raggedright
  \includegraphics[keepaspectratio, scale=0.36]{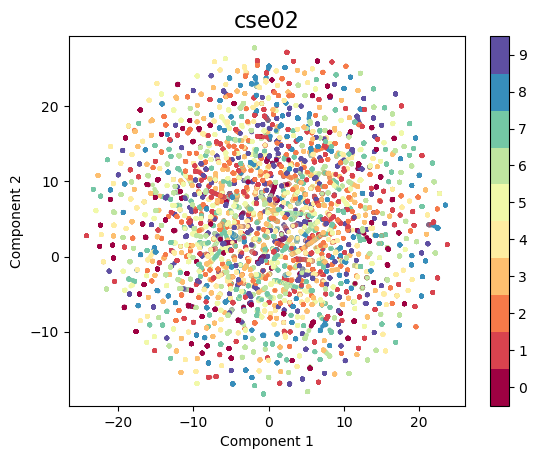}\\
  \end{minipage}
  \begin{minipage}[b]{0.30\textwidth}
  \raggedright
  \includegraphics[keepaspectratio, scale=0.36]{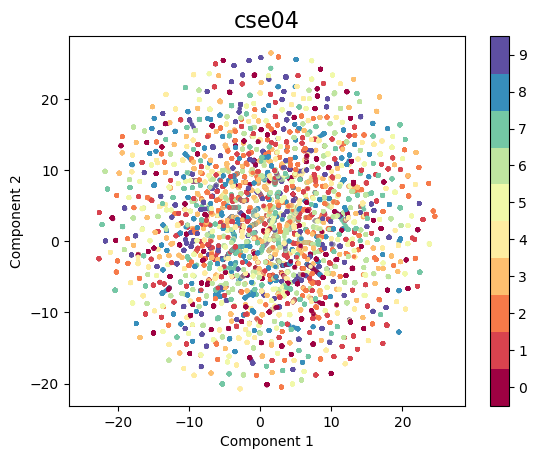}\\
  \end{minipage}
  \begin{minipage}[b]{0.30\textwidth}
  \raggedright
  \includegraphics[keepaspectratio, scale=0.36]{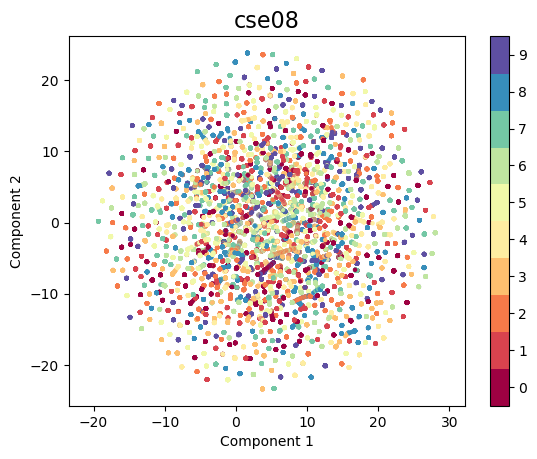}\\
  \end{minipage}

\caption{UMAP projections of model parameters (fc2.weight) by node from epoch = 1 to epoch = 5000. The color indicates the node number. }
\label{fig:UMAP}
\end{figure*}

\subsection{Trend of Accuracy}

Fig. \ref{fig:Accuracy} shows the trend of the test accuracy with the best hyperparameters shown in Table \ref{tab:accuracy}. In static networks, as nodes were always connected and could always exchange model parameters among them, they rapidly trained the model with increasing accuracy just the same as conventional federated learning. In dynamic networks, the training speed was low compared to static cases and federated learning. This is simply because in dynamic networks, during the nodes were alone, they did not make any training process. Especially, the training speeds on RWP mobility cases were rwp0500 $>$ rwp1000 $>$ rwp2000. This corresponds to the density of nodes in the mobility area. Higher density allowed faster training. This shows that the rwp2000 case requires a longer time to have more contacts and to obtain enough accuracy.

\subsection{Reductions of Confusion}

Fig. \ref{fig:confusion_matrix} shows some examples of confusion reductions. These examples are the confusion matrices of the model of node 9 for static\_line, rwp0500, and cse02 cases.

At epoch = 1, i.e., after the single round of WAFL algorithm with encountered nodes, we can observe that the models had wrong predictions regarding some data samples as label 9's samples. For example, about 33-39\% of label 4's samples were predicted as 9, and about 15-31\% of label 7's samples were as 9. This is because just after the pre-self-training process with only a single WAFL epoch, it was over-fitted to the local training data at node 9, which has much larger portions of label 9's samples compared to the other samples.

We can also observe that label 2's samples were predicted as label 3 (4-6\%) or 8 (7-8\%), label 5 as 3 (4-5\%) or 8 (5-6\%), and label 6 as 5 (4-5\%). These confusions were made probably because of the much fewer data samples regarding label 0-8's data for training. 

As the training proceeds, at epoch = 200 and epoch = 5000, we can observe that they could gradually reduce confusion, achieving better performance in general.

\subsection{Model Convergence}

We have investigated the convergences of model parameters. In this evaluation, we consider the distances among model parameter vectors of the ten nodes. Let us consider the ten model parameter vectors $\theta^{(n)} (n=0, \ldots, 9)$, and calculate distances from the mean of these parameter vectors. Here, $\theta^{(n)}$ can be one of fc1.weight, fc1.bias, fc2.weight, and fc2.bias. Then, we calculate the averaged distance as the metric of model convergence errors.

More formally, at a specific epoch, we first calculate the mean of model parameter vectors by node,

\begin{equation}
\bar{\theta}=\frac{1}{N}{\sum_{n=1}^{N} \theta^{(n)}}.
\end{equation}

Then, we calculate the averaged distance from the mean parameter $\bar{\theta}$ as convergence error $E$:

\begin{equation}
E=\frac{1}{N}{\sum_{n=1}^{N} Distance(\theta^{(n)},\bar{\theta})}.
\end{equation}

Here, the distance between two model parameters $\theta_1, \theta_2$ is defined by the following formula:
\begin{equation}
Distance(\theta_1, \theta_2) = \frac{1}{\vert \theta \vert}{\sqrt{\sum_{i=1}^{\vert \theta \vert} (\theta_1[i]-\theta_2[i])^2}}.
\end{equation}

Fig. \ref{fig:ModelDistance} shows the trend of model convergence errors calculated in this way. We have chosen the models developed with the best $\lambda$ and $\eta$ configurations studied in Table \ref{tab:accuracy}. The static cases and the conventional federated learning provided similar trends including the values of the error. Here, we have observed that static\_dense provided smoother patterns. This is probably because static\_dense's $\lambda$ was 0.1 whereas others' $\lambda$ was 1.0. Conventional federated learning has shown a small error even from the epoch = 1. This is because of the nature of aggregating into a single global model. 

As for the self-training case, the errors were always increasing. This indicates that long-term lonely learning diverges its model parameters.

The dynamic cases also allowed gradual error reductions although there were some temporal error increases. This indicates that model aggregation in WAFL has enabled cooperative learning. If not, the model parameters will diverge as the self-training case does.

\subsection{Model Visualization}

In order to investigate more into the model parameters, we have visualized them as Fig. \ref{fig:UMAP}. As model parameters are the elements of a multi-dimensional vector, we have used uniform manifold approximation and projection (UMAP) \cite{mcinnes2020umap} for mapping them to a 2-dimension space. In this visualization, each epoch has ten vectors associated with the ten nodes. We have assigned different colors for those ten nodes from 0 to 9. We have studied all of fc1.weight, fc1.bias, fc2.weight, and fc2.bias from epoch = 1 to epoch = 5000, but we present only the fc2.weight case in the figure because the results were similar. Please note that fc2.weight has 1280 parameters, i.e., this projection was made from the 1280-dimension space to a 2-dimension space. The configurations of $\lambda$ and $\eta$ correspond to the best cases shown in Table \ref{tab:accuracy}.

From these results, we can observe that the conventional federated learning has shown clear lines, meaning that different nodes are plotted at the same location and steadily moving in the space. The lines are sometimes broken, probably because of the sudden changes made by the optimizer. These breaks correspond to the jumps of errors in Fig. \ref{fig:ModelDistance}.

We can also observe that static\_ringstar, static\_tree, and static\_line have shown similar patterns. However, especially static\_line case, nodes were plotted not at the exact same locations but nearby. The relationships of the locations corresponded to the network topology of static\_line (Fig. \ref{fig:NetworkModel}).

As for static\_dense, it was very different from other static cases. This is most probably because of the differences in coefficient $\lambda$. If $\lambda=0.1$ instead of $\lambda=1.0$, the parameters of the models did move around without concentrating on a specific location at each epoch. 

In RWP cases, rwp0500 has more visible plots than rwp1000 and rwp2000. This came from the differences in opportunities for node-to-node contacts. As the nodes were denser, they could get higher chances of encountering each other, and the model parameters had higher chances of changing. As for CSEs, there were no clear differences among them. Those dots were scattered like rwp0500.

As for the self-training case, the model parameters of each node independently moved around drawing multiple lines. 

\section{Open Research Directions}

In this paper, we have proposed WAFL and shown their basic characteristics on a simple fully-connected neural network with the basic MNIST dataset rather than today's advanced neural networks and complex datasets. This is because our focus was on the analysis of model aggregations with simple layer settings with a baseline dataset on varieties of mobility scenarios.

WAFL algorithm itself can be also applied into other kinds of deep learning models such as CNN\cite{lecun1995convolutional}, GNN\cite{kipf2016semi}, LSTM\cite{hochreiter1997long}, Autoencoder\cite{hinton2006reducing}, GAN\cite{goodfellow2014generative}, Word2vec\cite{mikolov2013efficient}, A3C\cite{mnih2016asynchronous}, including more complex and heavy deep learning models such as SSD\cite{liu2016ssd}, BERT\cite{devlin2018bert}, and MDETR\cite{kamath2021mdetr}. Such further applications of WAFL and analysis are open problems. 

Actually, simple neural networks are very practical in today's wireless ad hoc scenarios where nodes are operated with battery-powered personal computers or smartphones.


Some practical use cases of WAFL may be realized in the context of transfer learning with very deep neural networks such as VGG\cite{simonyan2014very} and ResNet\cite{he2016deep}.
Transfer learning uses pre-trained models, and only around the output layers are expected to be trained. In WAFL's point of view, the nodes exchange such a small number of trainable parameters. This could be acceptable in today's wireless ad hoc scenarios.

Security is another important topic. Security itself is a common issue in autonomous and cooperative distributed systems, especially when it comes to open systems. In wireless ad hoc networking, a friend mechanism \cite{abd2008friend} can be helpful for acquiring basic security. Anomaly detection, adversarial detection, or trust mechanisms should be studied as \cite{bhagoji2019analyzing, bagdasaryan2020backdoor} did.

\section{Conclusion}

We have proposed a wireless ad hoc federated learning (WAFL) -- a fully distributed cooperative machine learning organized by the nodes physically nearby. WAFL can develop generalized models from Non-IID datasets stored in distributed nodes by exchanging and aggregating them with each other over opportunistic direct node-to-node contacts.


In our benchmark-based evaluation with various opportunistic networks, WAFL has achieved higher accuracy of 94.8-96.3\% than the self-training case of 84.7\%. All of our evaluation results show that WAFL can train and converge the model parameters from highly-partitioned Non-IID datasets without any centralized or third-party mechanisms.



\bibliographystyle{unsrt}
\bibliography{tai.bib}

\begin{thebibliography}{10}

\bibitem{hashem2015rise}
Ibrahim Abaker~Targio Hashem, Ibrar Yaqoob, Nor~Badrul Anuar, Salimah Mokhtar,
  Abdullah Gani, and Samee~Ullah Khan.
\newblock The rise of {“Big Data”} on cloud computing: Review and open
  research issues.
\newblock {\em Information systems}, 47:98--115, 2015.

\bibitem{l2017machine}
Alexandra L’heureux, Katarina Grolinger, Hany~F Elyamany, and Miriam~AM
  Capretz.
\newblock Machine learning with {Big Data}: Challenges and approaches.
\newblock {\em IEEE Access}, 5:7776--7797, 2017.

\bibitem{hessel2020regulation}
Stefan Hessel and Andreas Rebmann.
\newblock Regulation of {Internet-of-Things} cybersecurity in {Europe and
  Germany} as exemplified by devices for children.
\newblock {\em International Cybersecurity Law Review}, 1(1):27--37, 2020.

\bibitem{voigt2017eu}
Paul Voigt and Axel Von~dem Bussche.
\newblock The {EU} general data protection regulation {(GDPR)}.
\newblock {\em A Practical Guide, 1st Ed., Cham: Springer International
  Publishing}, 10(3152676):10--5555, 2017.

\bibitem{xiao2012security}
Zhifeng Xiao and Yang Xiao.
\newblock Security and privacy in cloud computing.
\newblock {\em IEEE communications surveys \& tutorials}, 15(2):843--859, 2012.

\bibitem{pearson2013privacy}
Siani Pearson.
\newblock Privacy, security and trust in cloud computing.
\newblock In {\em Privacy and security for cloud computing}, pages 3--42.
  Springer, 2013.

\bibitem{konevcny2016federated}
Jakub Kone{\v{c}}n{\`y}, H~Brendan McMahan, Felix~X Yu, Peter Richt{\'a}rik,
  Ananda~Theertha Suresh, and Dave Bacon.
\newblock Federated learning: Strategies for improving communication
  efficiency.
\newblock {\em arXiv preprint arXiv:1610.05492}, 2016.

\bibitem{li2020federated}
Tian Li, Anit~Kumar Sahu, Ameet Talwalkar, and Virginia Smith.
\newblock Federated learning: Challenges, methods, and future directions.
\newblock {\em IEEE Signal Processing Magazine}, 37(3):50--60, 2020.

\bibitem{tan2022towards}
Alysa~Ziying Tan, Han Yu, Lizhen Cui, and Qiang Yang.
\newblock Towards personalized federated learning.
\newblock {\em IEEE Transactions on Neural Networks and Learning Systems},
  2022.

\bibitem{festag2014cooperative}
Andreas Festag.
\newblock Cooperative intelligent transport systems standards in europe.
\newblock {\em IEEE communications magazine}, 52(12):166--172, 2014.

\bibitem{zhao2018federated}
Yue Zhao, Meng Li, Liangzhen Lai, Naveen Suda, Damon Civin, and Vikas Chandra.
\newblock Federated learning with non-iid data.
\newblock {\em arXiv preprint arXiv:1806.00582}, 2018.

\bibitem{bettstetter2004stochastic}
Christian Bettstetter, Hannes Hartenstein, and Xavier P{\'e}rez-Costa.
\newblock Stochastic properties of the random waypoint mobility model.
\newblock {\em Wireless networks}, 10(5):555--567, 2004.

\bibitem{ochiai2008mobility}
Hideya Ochiai and Hiroshi Esaki.
\newblock Mobility entropy and message routing in community-structured delay
  tolerant networks.
\newblock In {\em Proceedings of the 4th Asian Conference on Internet
  Engineering}, pages 93--102, 2008.

\bibitem{wang2020optimizing}
Hao Wang, Zakhary Kaplan, Di~Niu, and Baochun Li.
\newblock Optimizing federated learning on non-iid data with reinforcement
  learning.
\newblock In {\em IEEE INFOCOM 2020-IEEE Conference on Computer
  Communications}, pages 1698--1707. IEEE, 2020.

\bibitem{lecun1995convolutional}
Yann LeCun, Yoshua Bengio, et~al.
\newblock Convolutional networks for images, speech, and time series.
\newblock {\em The handbook of brain theory and neural networks},
  3361(10):1995, 1995.

\bibitem{kipf2016semi}
Thomas~N Kipf and Max Welling.
\newblock Semi-supervised classification with graph convolutional networks.
\newblock {\em arXiv preprint arXiv:1609.02907}, 2016.

\bibitem{hochreiter1997long}
Sepp Hochreiter and J{\"u}rgen Schmidhuber.
\newblock Long short-term memory.
\newblock {\em Neural computation}, 9(8):1735--1780, 1997.

\bibitem{hinton2006reducing}
Geoffrey~E Hinton and Ruslan~R Salakhutdinov.
\newblock Reducing the dimensionality of data with neural networks.
\newblock {\em science}, 313(5786):504--507, 2006.

\bibitem{goodfellow2014generative}
Ian Goodfellow, Jean Pouget-Abadie, Mehdi Mirza, Bing Xu, David Warde-Farley,
  Sherjil Ozair, Aaron Courville, and Yoshua Bengio.
\newblock Generative adversarial nets.
\newblock {\em Advances in neural information processing systems}, 27, 2014.

\bibitem{mikolov2013efficient}
Tomas Mikolov, Kai Chen, Greg Corrado, and Jeffrey Dean.
\newblock Efficient estimation of word representations in vector space.
\newblock {\em arXiv preprint arXiv:1301.3781}, 2013.

\bibitem{mnih2016asynchronous}
Volodymyr Mnih, Adria~Puigdomenech Badia, Mehdi Mirza, Alex Graves, Timothy
  Lillicrap, Tim Harley, David Silver, and Koray Kavukcuoglu.
\newblock Asynchronous methods for deep reinforcement learning.
\newblock In {\em International conference on machine learning}, pages
  1928--1937. PMLR, 2016.

\bibitem{frodigh2000wireless}
Magnus Frodigh, Per Johansson, and Peter Larsson.
\newblock Wireless ad hoc networking: the art of networking without a network.
\newblock {\em Ericsson review}, 4(4):249, 2000.

\bibitem{schollmeier2001definition}
R{\"u}diger Schollmeier.
\newblock A definition of peer-to-peer networking for the classification of
  peer-to-peer architectures and applications.
\newblock In {\em Proceedings First International Conference on Peer-to-Peer
  Computing}, pages 101--102. IEEE, 2001.

\bibitem{abolhasan2004review}
Mehran Abolhasan, Tadeusz Wysocki, and Eryk Dutkiewicz.
\newblock A review of routing protocols for mobile ad hoc networks.
\newblock {\em Ad hoc networks}, 2(1):1--22, 2004.

\bibitem{hartenstein2008tutorial}
Hannes Hartenstein and LP~Laberteaux.
\newblock A tutorial survey on vehicular ad hoc networks.
\newblock {\em IEEE Communications magazine}, 46(6):164--171, 2008.

\bibitem{fall2003delay}
Kevin Fall.
\newblock A delay-tolerant network architecture for challenged {Internets}.
\newblock In {\em Proceedings of the 2003 conference on Applications,
  technologies, architectures, and protocols for computer communications},
  pages 27--34, 2003.

\bibitem{nguyen2021federated}
Dinh~C Nguyen, Ming Ding, Pubudu~N Pathirana, Aruna Seneviratne, Jun Li, and
  H~Vincent Poor.
\newblock Federated learning for internet of things: A comprehensive survey.
\newblock {\em IEEE Communications Surveys \& Tutorials}, 2021.

\bibitem{roy2019braintorrent}
Abhijit~Guha Roy, Shayan Siddiqui, Sebastian P{\"o}lsterl, Nassir Navab, and
  Christian Wachinger.
\newblock {BrainTorrent}: A peer-to-peer environment for decentralized
  federated learning.
\newblock {\em arXiv preprint arXiv:1905.06731}, 2019.

\bibitem{li2020blockchain}
Yuzheng Li, Chuan Chen, Nan Liu, Huawei Huang, Zibin Zheng, and Qiang Yan.
\newblock A blockchain-based decentralized federated learning framework with
  committee consensus.
\newblock {\em IEEE Network}, 35(1):234--241, 2020.

\bibitem{bao2019flchain}
Xianglin Bao, Cheng Su, Yan Xiong, Wenchao Huang, and Yifei Hu.
\newblock {FLchain}: A blockchain for auditable federated learning with trust
  and incentive.
\newblock In {\em 2019 5th International Conference on Big Data Computing and
  Communications (BIGCOM)}, pages 151--159. IEEE, 2019.

\bibitem{pokhrel2020federated}
Shiva~Raj Pokhrel and Jinho Choi.
\newblock Federated learning with blockchain for autonomous vehicles: Analysis
  and design challenges.
\newblock {\em IEEE Transactions on Communications}, 68(8):4734--4746, 2020.

\bibitem{ur2020towards}
Muhammad~Habib ur~Rehman, Khaled Salah, Ernesto Damiani, and Davor Svetinovic.
\newblock Towards blockchain-based reputation-aware federated learning.
\newblock In {\em IEEE Conference on Computer Communications Workshops (INFOCOM
  WKSHPS)}, pages 183--188. IEEE, 2020.

\bibitem{warnat2021swarm}
Stefanie Warnat-Herresthal, Hartmut Schultze, Krishnaprasad~Lingadahalli
  Shastry, Sathyanarayanan Manamohan, et~al.
\newblock Swarm learning for decentralized and confidential clinical machine
  learning.
\newblock {\em Nature}, 594(7862):265--270, 2021.

\bibitem{hu2019decentralized}
Chenghao Hu, Jingyan Jiang, and Zhi Wang.
\newblock Decentralized federated learning: A segmented gossip approach.
\newblock {\em arXiv preprint arXiv:1908.07782}, 2019.

\bibitem{pappas2021ipls}
Christodoulos Pappas, Dimitris Chatzopoulos, Spyros Lalis, and Manolis Vavalis.
\newblock {IPLS}: A framework for decentralized federated learning.
\newblock In {\em 2021 IFIP Networking Conference}, pages 1--6. IEEE, 2021.

\bibitem{haas2006gossip}
Zygmunt~J Haas, Joseph~Y Halpern, and Li~Li.
\newblock Gossip-based ad hoc routing.
\newblock {\em IEEE/ACM Transactions on networking}, 14(3):479--491, 2006.

\bibitem{vahdat2000epidemic}
Amin Vahdat, David Becker, et~al.
\newblock Epidemic routing for partially connected ad hoc networks, 2000.

\bibitem{brik2020federated}
Bouziane Brik, Adlen Ksentini, and Maha Bouaziz.
\newblock Federated learning for {UAVs-enabled} wireless networks: Use cases,
  challenges, and open problems.
\newblock {\em IEEE Access}, 8:53841--53849, 2020.

\bibitem{bekmezci2013flying}
Ilker Bekmezci, Ozgur~Koray Sahingoz, and {\c{S}}amil Temel.
\newblock Flying ad-hoc networks {(FANETs)}: A survey.
\newblock {\em Ad Hoc Networks}, 11(3):1254--1270, 2013.

\bibitem{liu2020federated}
Yi~Liu, Jiangtian Nie, Xuandi Li, Syed~Hassan Ahmed, Wei Yang~Bryan Lim, and
  Chunyan Miao.
\newblock Federated learning in the sky: Aerial-ground air quality sensing
  framework with {UAV} swarms.
\newblock {\em IEEE Internet of Things Journal}, 8(12):9827--9837, 2020.

\bibitem{wang2020learning}
Yuntao Wang, Zhou Su, Ning Zhang, and Abderrahim Benslimane.
\newblock Learning in the air: Secure federated learning for {UAV-assisted}
  crowdsensing.
\newblock {\em IEEE Transactions on network science and engineering},
  8(2):1055--1069, 2020.

\bibitem{pham2021uav}
Quoc-Viet Pham, Ming Zeng, Rukhsana Ruby, Thien Huynh-The, and Won-Joo Hwang.
\newblock {UAV} communications for sustainable federated learning.
\newblock {\em IEEE Transactions on Vehicular Technology}, 70(4):3944--3948,
  2021.

\bibitem{shiri2020communication}
Hamid Shiri, Jihong Park, and Mehdi Bennis.
\newblock Communication-efficient massive {UAV} online path control: Federated
  learning meets mean-field game theory.
\newblock {\em IEEE Transactions on Communications}, 68(11):6840--6857, 2020.

\bibitem{ochiai2011case}
Hideya Ochiai, Kenji Matsuo, Satoshi Matsuura, and Hiroshi Esaki.
\newblock A case study of utmesh: Design and impact of real world experiments
  with wi-fi and bluetooth devices.
\newblock In {\em IEEE/IPSJ International Symposium on Applications and the
  Internet}, pages 433--438. IEEE, 2011.

\bibitem{ahmed2020survey}
Nadeem Ahmed, Regio~A Michelin, Wanli Xue, Sushmita Ruj, Robert Malaney,
  Salil~S Kanhere, Aruna Seneviratne, Wen Hu, Helge Janicke, and Sanjay~K Jha.
\newblock A survey of {COVID-19} contact tracing apps.
\newblock {\em IEEE Access}, 8:134577--134601, 2020.

\bibitem{camp2002survey}
Tracy Camp, Jeff Boleng, and Vanessa Davies.
\newblock A survey of mobility models for ad hoc network research.
\newblock {\em Wireless communications and mobile computing}, 2(5):483--502,
  2002.

\bibitem{opitz2019macro}
Juri Opitz and Sebastian Burst.
\newblock Macro f1 and macro f1.
\newblock {\em arXiv preprint arXiv:1911.03347}, 2019.

\bibitem{mcinnes2020umap}
Leland McInnes, John Healy, and James Melville.
\newblock Umap: uniform manifold approximation and projection for dimension
  reduction.
\newblock 2020.

\bibitem{liu2016ssd}
Wei Liu, Dragomir Anguelov, Dumitru Erhan, Christian Szegedy, Scott Reed,
  Cheng-Yang Fu, and Alexander~C Berg.
\newblock {SSD}: Single shot multibox detector.
\newblock In {\em European conference on computer vision}, pages 21--37.
  Springer, 2016.

\bibitem{devlin2018bert}
Jacob Devlin, Ming-Wei Chang, Kenton Lee, and Kristina Toutanova.
\newblock {BERT}: Pre-training of deep bidirectional transformers for language
  understanding.
\newblock {\em arXiv preprint arXiv:1810.04805}, 2018.

\bibitem{kamath2021mdetr}
Aishwarya Kamath, Mannat Singh, Yann LeCun, Gabriel Synnaeve, Ishan Misra, and
  Nicolas Carion.
\newblock {MDETR-modulated} detection for end-to-end multi-modal understanding.
\newblock In {\em Proceedings of the IEEE/CVF International Conference on
  Computer Vision}, pages 1780--1790, 2021.

\bibitem{simonyan2014very}
Karen Simonyan and Andrew Zisserman.
\newblock Very deep convolutional networks for large-scale image recognition.
\newblock {\em arXiv preprint arXiv:1409.1556}, 2014.

\bibitem{he2016deep}
Kaiming He, Xiangyu Zhang, Shaoqing Ren, and Jian Sun.
\newblock Deep residual learning for image recognition.
\newblock In {\em Proceedings of the IEEE conference on computer vision and
  pattern recognition}, pages 770--778, 2016.

\bibitem{abd2008friend}
Shukor Abd~Razak, Normalia Samian, and Mohd~Aizaini Maarof.
\newblock A friend mechanism for mobile ad hoc networks.
\newblock In {\em 2008 The Fourth International Conference on Information
  Assurance and Security}, pages 243--248. IEEE, 2008.

\bibitem{bhagoji2019analyzing}
Arjun~Nitin Bhagoji, Supriyo Chakraborty, Prateek Mittal, and Seraphin Calo.
\newblock Analyzing federated learning through an adversarial lens.
\newblock In {\em International Conference on Machine Learning}, pages
  634--643. PMLR, 2019.

\bibitem{bagdasaryan2020backdoor}
Eugene Bagdasaryan, Andreas Veit, Yiqing Hua, Deborah Estrin, and Vitaly
  Shmatikov.
\newblock How to backdoor federated learning.
\newblock In {\em International Conference on Artificial Intelligence and
  Statistics}, pages 2938--2948. PMLR, 2020.

\end{thebibliography}

\begin{IEEEbiography}[{\includegraphics[width=1in,height=1.25in,clip,keepaspectratio]{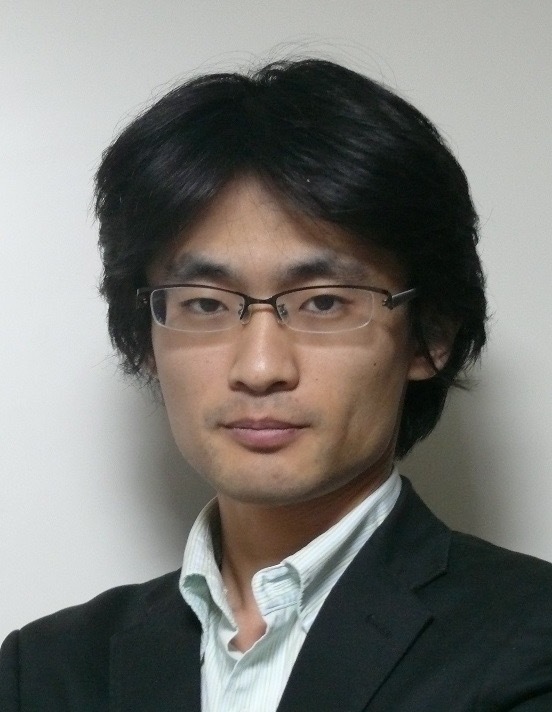}}]{Hideya Ochiai}{\space}(Member, IEEE) received B.E. and master of information science and technology from the University of Tokyo, Japan, in 2006 and 2008, respectively. He received Ph.D. of information science and technology from the University of Tokyo in 2011. He became an assistant professor, and associated professor from 2011, and 2017 of the University of Tokyo respectively. His research ranges from IoT system and protocol designs to peer-to-peer overlay networks, delay-disruption tolerant networks, network security, and decentralized machine learning. He joined standardization activities of IEEE from 2008, and of ISO/IEC JTC1/SC6 from 2012. He is a chair of the board of the Green University of Tokyo Project from 2016, a chair of LAN-security monitoring project from 2018, a chair of decentralized AI project from 2022.
\end{IEEEbiography}

\begin{IEEEbiography}[{\includegraphics[width=1in,height=1.25in,clip,keepaspectratio]{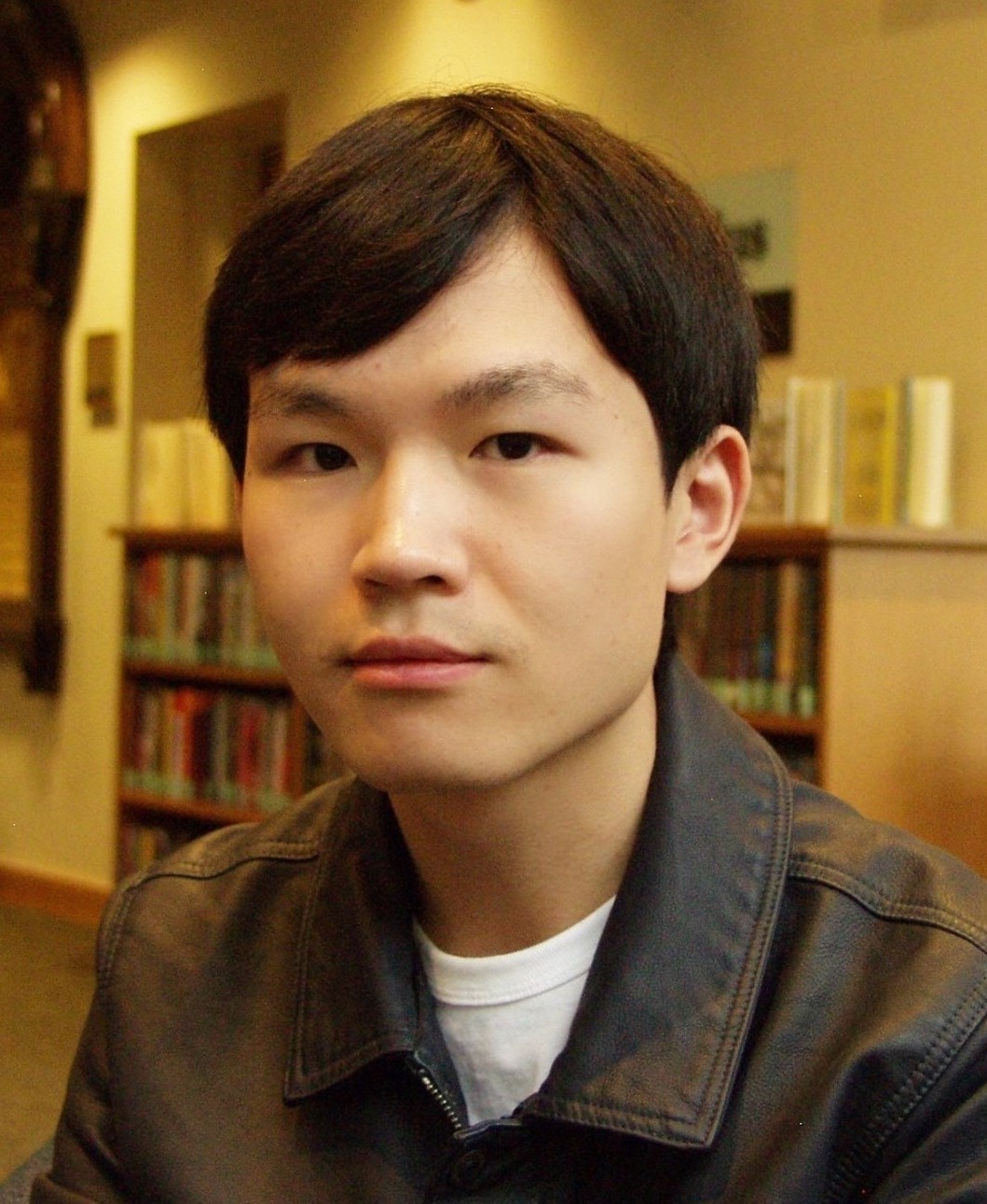}}]{Yuwei Sun}{\space}(Member, IEEE) received the B.E. degree in computer science and technology from North China Electric Power University, in 2018, and the M.E. degree (Hons.) in information and communication engineering from the University of Tokyo, in 2021, where he is currently pursuing the Ph.D. degree with the Graduate School of Information Science and Technology. In 2020, he was the fellow of the Advanced Study Program (ASP) at the Massachusetts Institute of Technology. He has been working with the Campus Computing Centre, United Nations University Centre on Cybersecurity, since 2019. He is a member of the AI Security and Privacy Team with the RIKEN Center for Advanced Intelligence Project working on trustworthy AI, and a Research Fellow at the Japan Society for the Promotion of Science (JSPS).
\end{IEEEbiography}

\begin{IEEEbiography}[{\includegraphics[width=1in,height=1.25in,clip,keepaspectratio]{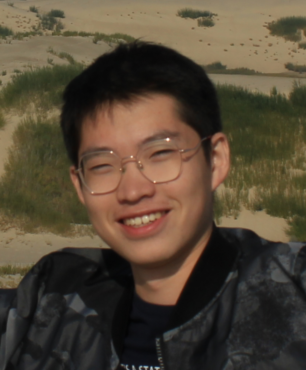}}]{Qingzhe Jin} is a Master course student in the Graduate School of Information Science and Technology at the University of Tokyo. He received his B.E. degree in Computer Science and Technology from Peking University, China in 2020. His research interests include but not limited to federated learning, computer networks and graph neural networks.
\end{IEEEbiography}

\begin{IEEEbiography}[{\includegraphics[width=1in,height=1.25in,clip,keepaspectratio]{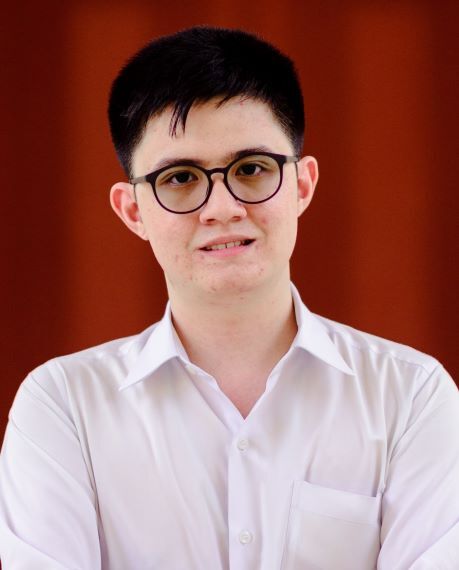}}]{Nattanon Wongwiwatchai} is a Ph.D. candidate in the Graduate School of Information Science and Technology at the University of Tokyo. He received B.Eng. and M.Eng. in Computer Engineering from Chulalongkorn University, Thailand, in 2019 and 2021, respectively. His research interests are in mobile application security and privacy, and decentralized machine learning.
\end{IEEEbiography}

\begin{IEEEbiography}[{\includegraphics[width=1in,height=1.25in,clip,keepaspectratio]{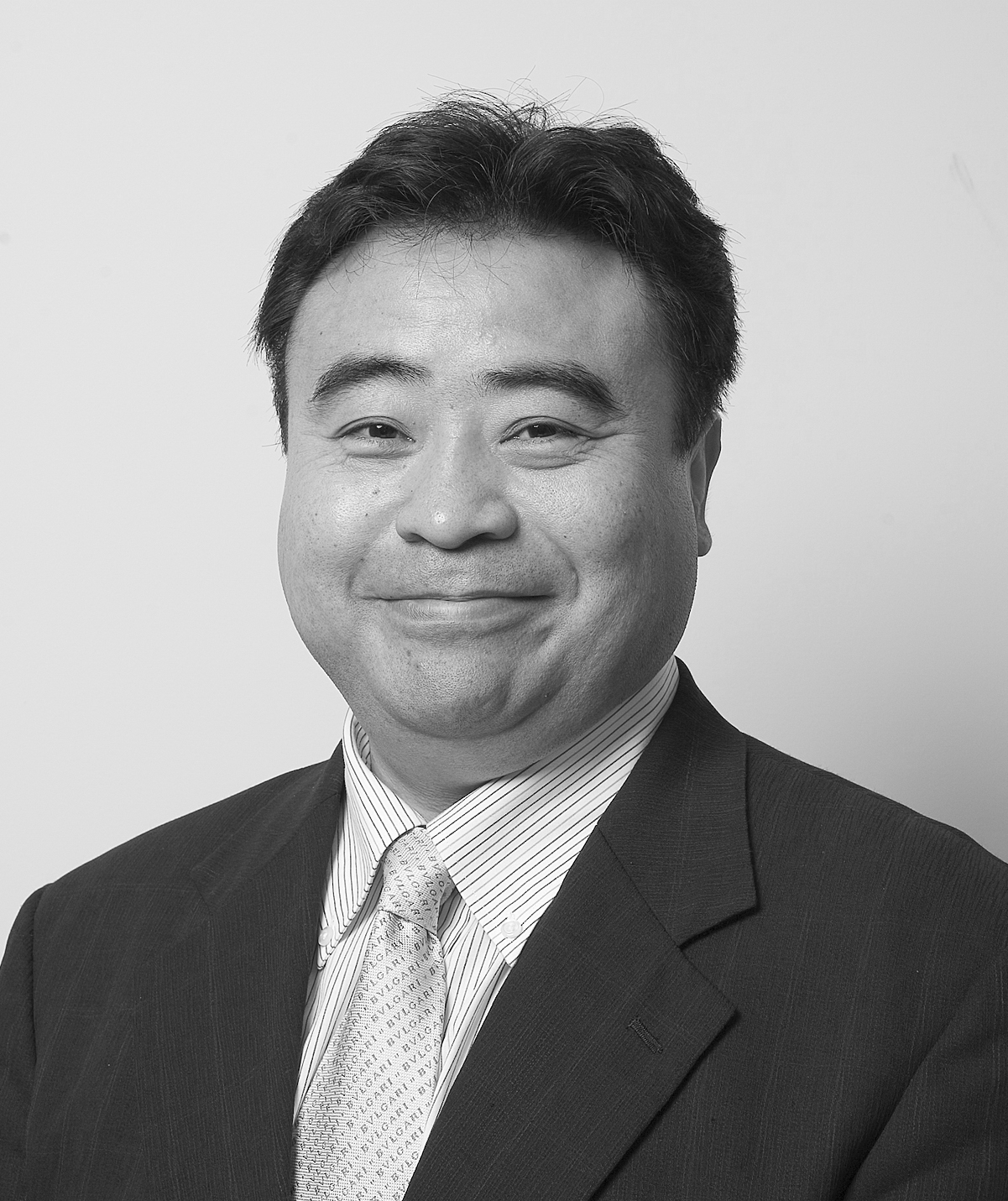}}]{Hiroshi Esaki}{\space}(Member, IEEE) received Ph.D. from the University of Tokyo, Japan, in 1998. In 1987, he joined Research and Development Center, Toshiba Corporation. From 1990 to 1991, he was at Applied Research Laboratory of Bell-core Inc., New Jersey, as a residential researcher. From 1994 to 1996, he was at Center for Telecommunication Research of
Columbia University in New York. From 1998, he has been serving as a professor at the University of Tokyo, and as a board member of WIDE Project. Currently, he is the executive director of IPv6 promotion council, vice president of JPNIC, IPv6 Forum Fellow, and director of
WIDE Project, chief architect of Digital Agency, Japan.
\end{IEEEbiography}

\end{document}